\documentclass[sigconf]{aamas} 

\usepackage{balance} 

\usepackage{amsmath,amsfonts,bm}

\def\eqref#1{equation~\ref{#1}}

\def\1{\bm{1}}

\DeclareMathAlphabet{\mathsfit}{\encodingdefault}{\sfdefault}{m}{sl}
\SetMathAlphabet{\mathsfit}{bold}{\encodingdefault}{\sfdefault}{bx}{n}

\usepackage{float}
\usepackage{hyperref}
\usepackage{url}
\usepackage{graphicx, wrapfig}
\usepackage{caption}
\usepackage{subcaption}
\usepackage{fancyref}
\usepackage{bbm}
\usepackage{bbold}
\usepackage{comment}
\usepackage{multirow}
\usepackage{breqn}
\usepackage{optidef}
\usepackage{algorithm}
\usepackage{algorithmic}
\usepackage{soul}
\usepackage{balance}
\usepackage{amsmath}

\usepackage{tikz}

\usetikzlibrary{arrows,shapes,decorations.markings}

\tikzstyle{vecArrow} = [thick, decoration={markings,mark=at position
   1 with {\arrow[semithick]{open triangle 60}}},
   double distance=1.4pt, shorten >= 5.5pt,
   preaction = {decorate},
   postaction = {draw,line width=1.4pt, white,shorten >= 4.5pt}]
   
  \tikzstyle{block} = [rectangle, draw, 
    text width=8em, text centered, rounded corners, minimum height=4em]
\tikzstyle{line} = [draw, -latex]
   
\usetikzlibrary{shapes,positioning,arrows}

\usepackage{pifont}
\usepackage{xcolor}
\newcommand{\cmark}{\textcolor{green!80!black}{\ding{51}}}
\newcommand{\xmark}{\textcolor{red}{\ding{55}}}

\setcopyright{ifaamas}
\acmConference[AAMAS '23]{Proc.\@ of the 22nd International Conference
on Autonomous Agents and Multiagent Systems (AAMAS 2023)}{May 29 -- June 2, 2023}
{London, United Kingdom}{A.~Ricci, W.~Yeoh, N.~Agmon, B.~An (eds.)}
\copyrightyear{2023}
\acmYear{2023}
\acmDOI{}
\acmPrice{}
\acmISBN{}

\acmSubmissionID{537}

\title{Enhancing Reinforcement Learning Agents with Local Guides}

\author{Paul Daoudi$^*$}
\affiliation{
  \institution{Huawei Noah's Ark Lab}
  \city{Paris}
  \country{France}}
\email{paul.daoudi1@huawei.com}

\author{Bogdan Robu}
\affiliation{
  \institution{GIPSA-Lab}
  \city{Grenoble}
  \country{France}}
\email{bogdan.robu@gipsa-lab.grenoble-inp.fr}

\author{Christophe Prieur}
\affiliation{
  \institution{GIPSA-Lab}
  \city{Grenoble}
  \country{France}}
\email{christophe.prieur@gipsa-lab.fr}

\author{Ludovic Dos Santos$^*$}
\affiliation{
  \institution{Criteo AI Lab}
  \city{Paris}
  \country{France}}
\email{l.dossantos@criteo.com}

\author{Merwan Barlier$^*$}
\affiliation{
  \institution{Huawei Noah's Ark Lab}
  \city{Paris}
  \country{France}}
\email{merwan.barlier@huawei.com}

\begin{abstract}

This paper addresses the problem of integrating local guide policies into a Reinforcement Learning agent. For this, we show how to adapt existing algorithms to this setting before introducing a novel algorithm based on a noisy policy-switching procedure. This approach builds on a proper Approximate Policy Evaluation (APE) scheme to provide a perturbation that carefully leads the local guides towards better actions. We evaluated our method on a set of classical Reinforcement Learning problems, including safety-critical systems where the agent cannot enter some areas at the risk of triggering catastrophic consequences. In all the proposed environments, our agent proved to be efficient at leveraging those policies to improve the performance of any APE-based Reinforcement Learning algorithm, especially in its first learning stages.

\end{abstract}


\newcommand{\BibTeX}{\rm B\kern-.05em{\sc i\kern-.025em b}\kern-.08em\TeX}

\begin{document}


\pagestyle{fancy}
\fancyhead{}


\maketitle 
\def\thefootnote{*}\footnotetext{Core Contributors}


\section{Introduction}

Reinforcement Learning (RL) aims at learning an optimal policy in an unknown environment by interacting with it. This discipline has known many successes on a wide range of simulated systems from recommender systems \citep{rojanavasu2005new, ZhengZZXY0L18} to complex video games \citep{MnihKSGAWR13, fuchs2021super}. Despite some advances in real-world environments such as balloon navigation \citep{bellemare2020autonomous} or plasma control in Tokamaks \citep{degrave2022magnetic}, RL is not ready to be applied to most real-world applications. Too many challenges must first be resolved \citep{dulac2021challenges, wang2020reinforcement}. In particular, the agent requires many interactions with the system to learn a good policy, which can be prohibitive in various cases: running real-world experiments may be more time - or money - consuming by many orders of magnitude compared to simulations, or when failing would eventually damage the system. In fact, these interactions may be completely restricted in the case of safety-critical environments, as they would have catastrophic consequences on themselves or their surroundings if an inadequate action is chosen in the wrong state. Hence, the agent must be able to learn with a small, potentially narrow, amount of samples. 

One of the main reasons for this requirement is the poor quality of the data sampled by the agent at the beginning of learning: an improper initialization of the policy drives the agent to visit meaningless or even catastrophic parts of the environment. In the former case, the agent would need a significant number of interactions with the system before discovering suitable areas of the environment. This results in running a mediocre policy for a long period, which is unacceptable on most real-world systems. For instance, a traditional way of teaching a robot how to walk would result in a substantial amount of falling, which would eventually damage it. Another example is the control of the cooling system of a server room: forcing the room to stay at a high temperature would damage the servers. This problem is all the more pronounced in high-dimensional environments with sparse rewards where relevant information are located in a narrow subspace of the environment. Real-life exploration is similar to what can be encountered when playing Montezuma's Revenge, an arcade game notoriously difficult to solve \citep{bellemare2013arcade, bellemare2016unifying} because of its sparse signals provided only after completing specific series of actions over extended periods. 

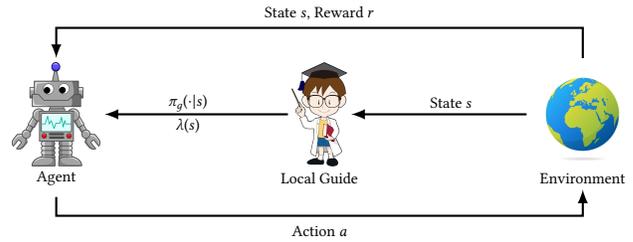
\begin{figure}[t]
\begin{tikzpicture}[node distance = 15em, auto, thick, scale=0.5, every node/.style={transform shape}]


    \node (Agent) at (-2, 0.2) {\includegraphics[height=2.75cm]{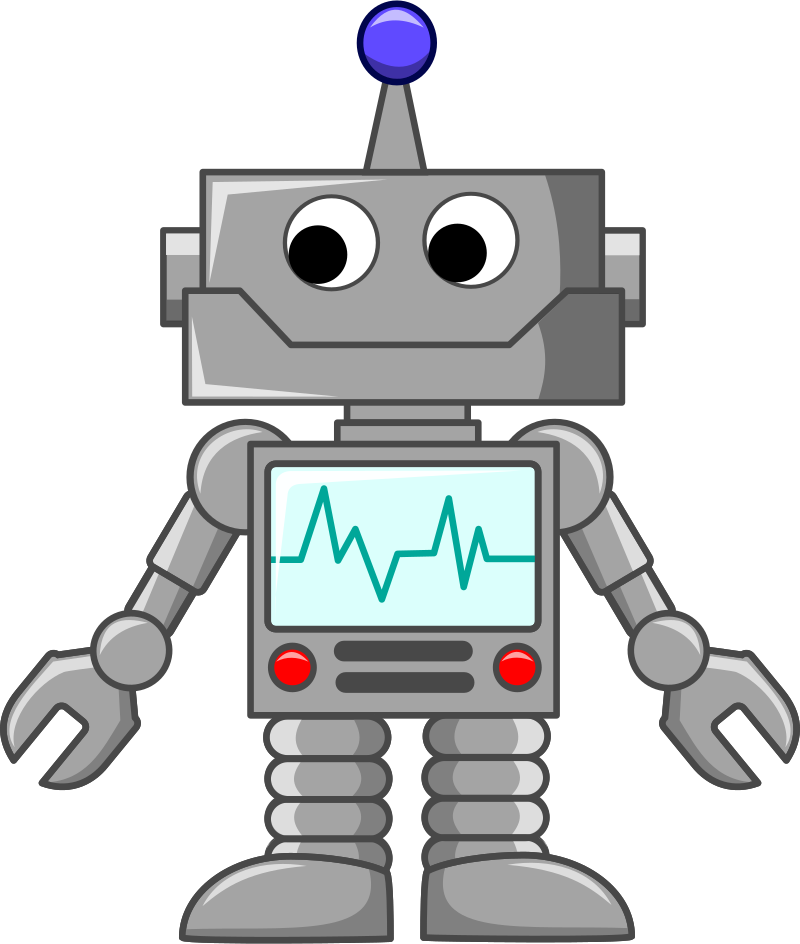}};
        \draw [color=black] (-2,-1.5) node {\LARGE Agent};

    \node (Expert) at (5,0.2) {\includegraphics[height=2.75cm]{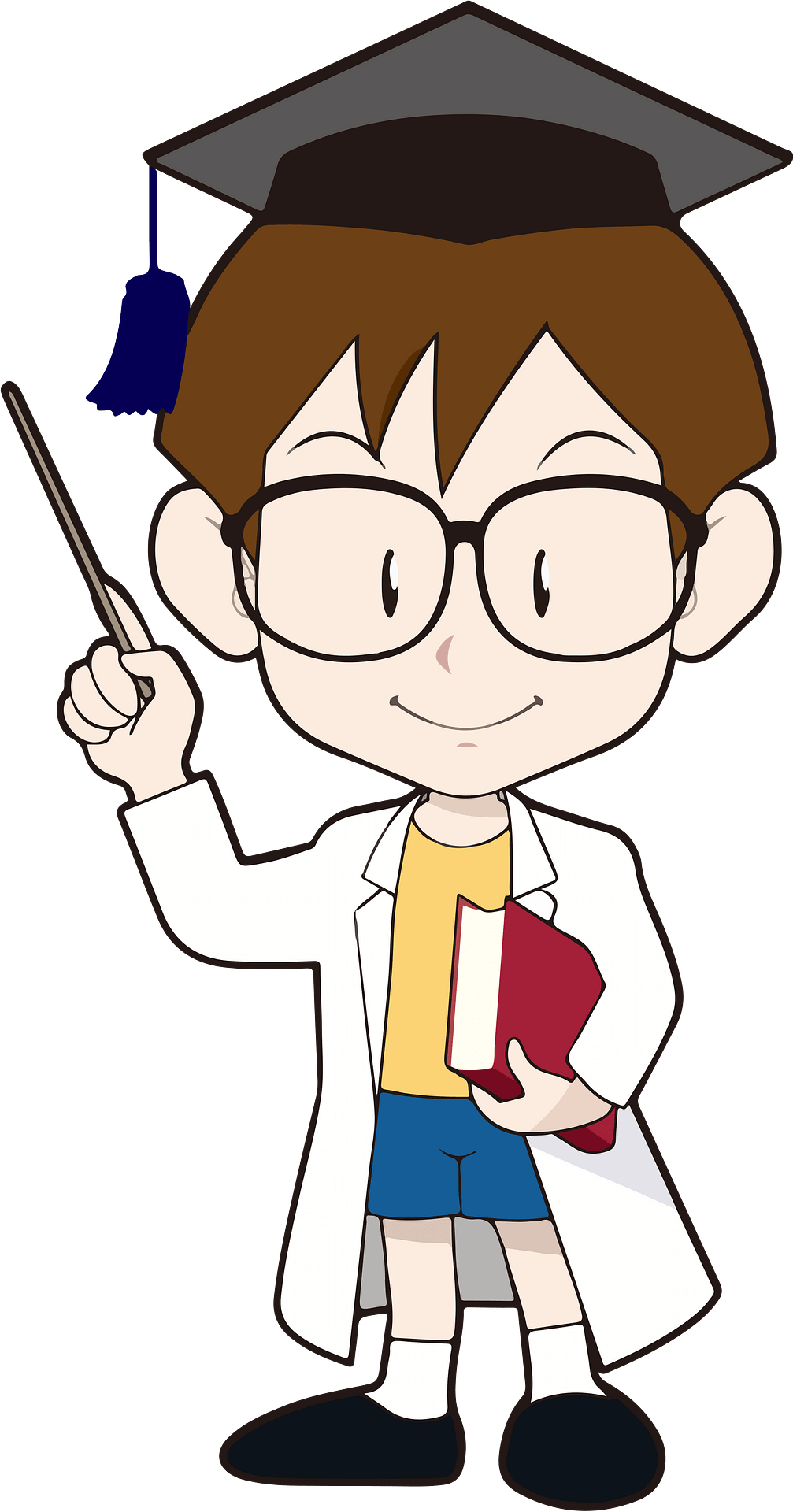}};
        \draw [color=black] (5,-1.5) node {\LARGE Local Guide};
    
    \node  (Environment) at (12,0.2) {\includegraphics[height=2.75cm]{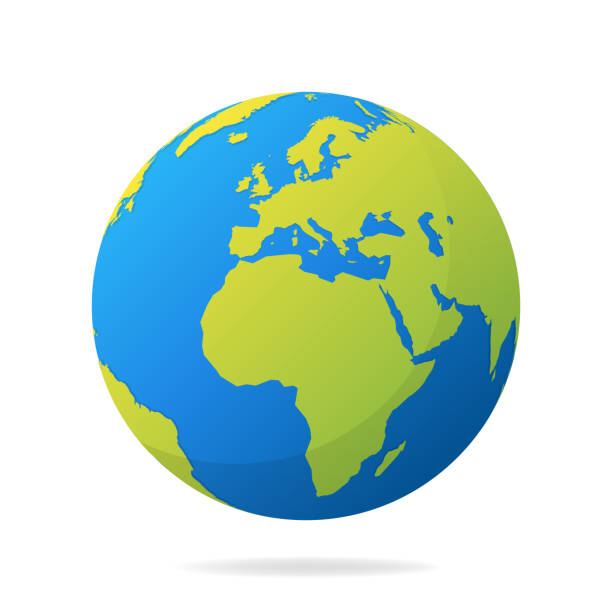}};
    \draw [color=black] (12,-1.5) node {\LARGE Environment};

    \path [line] (-2, -2) -- (-2, -2.5) -- (12, -2.5) -- (12, -1.8);

    \path [line] (Environment.90) -- (12, 2.5) -- (-2, 2.5) -- (Agent.90);
    
    \node () at (5, 2.9) {\LARGE State $s$, Reward $r$};

    \path [line] (Environment.180) -- (Expert.0);

    \node () at (8.5, 0.5) {\LARGE State $s$};
    
    \node () at (1.5, 0.5) {\LARGE $\pi_g(\cdot\vert s)$};

    \node () at (1.5, -0.1) {\LARGE $\lambda(s)$};

    \node () at (5, -2.9) {\LARGE Action $a$};

    \path [line] (Expert.180) -- (Agent.0);

\end{tikzpicture}
\caption{Reinforcement Learning with Local Guides}{At each step, the agent receives a state $s$, the reward $r$ of the previous state-action pair, as well as the guide policy $\pi_g(\cdot\vert s)$ with its associated confidence function $\lambda(s)$.} \label{rllg}
\vskip -0.15in
\end{figure}

Consequently, satisfactory RL agents must be able to quickly discover meaningful information about the environment with carefully chosen interactions with the system, that are both informative and safe. In the literature, a common way to go towards this purpose is to consider information provided by a guide policy (potentially sub-optimal). One proposition is to rely on external data that has been gathered with this guide to properly initialize any RL agent \citep{nair2020awac, lee2022offline, lyu2022mildly}, and/or to lead the subsequent optimization process \citep{schaal1996learning, ravichandar2020recent}. This has been successful in solving complex tasks, including the first levels of Montezuma's Revenge \citep{aytar2018playing, pohlen2018observe}. However, the suggested algorithms require a substantial amount of data to be efficient \citep{nair2018overcoming} which can be difficult to get in some real-world environments where sampling is expensive. In fact, data can be nearly impossible to gather from safety-critical environments since states leading to disastrous repercussions cannot be visited. Another line of work directly considers having access to the global sub-optimal guide which avoids the need for sampling from a costly environment. It acts as a teacher for the RL agent \citep{zimmer2014teacher}, which can either be \textit{attractive} \citep{schmitt2018kickstarting, uchendu2022jump, agarwal2022beyond} to lead the agent towards meaningful areas, or \textit{repulsive} \citep{turchetta2020safe, wagener2021safe} to prevent it from entering unsafe states. Although presenting considerable improvements, almost none of the methods introducing a guide tackles the policy initialization with the exception of \citep{uchendu2022jump}, or \citep{agarwal2022beyond} at the cost of initial sampling.

More importantly, both of these families assume access to a guide that would be efficient in all the state space of the environment, named in this paper as \textit{global guide}. Considering the increasing complexity of real-world systems, such global information becomes too difficult to build using classical tools \citep{lasserre2008nonlinear, taylor2021towards}. Nonetheless, information can often still be extracted, regardless of the complexity of the environment. For instance, experts may build heuristics such as catch-up policies to avoid going into unsafe states, or, when the dynamics can be inferred on some parts of the system, apply some of the many tools from Optimal Control (OC) theory \citep{Khalilbook, todorov2006optimal} for building attractive policies to guide the agent towards relevant information. In particular, Gain Scheduling \citep{rugh2000research, leith2000survey} is thoroughly used in the industry \citep{alcala2018gain, gallego2019gain, coutinho2021dynamic} to provide controllers that are relevant only at the vicinity of chosen operating points.

Inspired by these real-world considerations, we propose to address a novel framework called \textit{Reinforcement Learning with Local Guides} (RLLG), that seeks to reduce the number of interactions between the RL agent and the environment. Within this setting, the agent must find a suitable global policy in the entire state-action space with the help of a controller that would be relevant only in a known region of the state space, named \textit{local guide}. This extension is representative of real-world demands and includes any kind of local controller - whether attractive or repulsive - that may be present in real-world systems. This setting generalizes to any external policy as it also encompasses global ones.

In this paper, we first present how to adapt the Approximate Policy Iteration scheme to take an external policy into account to the RLLG setting, and analyze their advantages and drawbacks. Then, we introduce a novel algorithm to cope with the presented weaknesses and validate our approach on a variety of complex environments in two different use cases. The various algorithms are first compared to guide the exploration process to improve the quality of the gathered data during the first episodes. They are further tested to prevent the agent from entering catastrophic states. In both cases, the relevance of introducing local controllers as well as the advantages of our method are outlined in the proposed environments.

\section{Related work}

Before diving into the RLLG setting, we present how expertise has previously been integrated into the RL framework. This expert information is often global and can take different forms: data, reward, or policy. We cover them below and include the few existing approaches leveraging local information in the last paragraph.

\paragraph{Imitation Learning with Demonstrations} Many works focused on a setting where the agent has access to a large amount of demonstration trajectories from a guide of the task. A first proposition to recover the policy of the guide is Behavioral Cloning (BC) \citep{SuttonB98} which uses the traditional Supervised Learning scheme. When the dataset is complete and comes from an expert policy, this technique was successful in special cases of autonomous driving \citep{pomerleau1988alvinn} and flying \citep{SammutHKM92}. However, when the agent finds itself in a situation not described by the dataset, it may choose catastrophic actions and its performance can quickly degrade \citep{CodevillaSLG19}. In fact, \citet{ross2010efficient} show that this distribution shift induces a quadratic error in the length of the episode for standard Supervised Learning algorithms, so \citet{ross2011reduction} proposed DAGGER, an approach mixing the expert and the learnt policy, to get a linear error. Among the many different Imitation Learning frameworks lies Inverse RL \citep{abbeel2004apprenticeship, HusseinGEJ17, arora2021survey} which attempts at correcting the weaknesses of BC by encoding the expert policy into a reward function that could be optimized. This approach has shown impressive results in a wide variety of environments, including high-dimensional ones \citep{ho2016generative, xiao2019wasserstein, cai2019imitation}. Nonetheless, this set of techniques seeks to reproduce and generalize the guide policy, not improve it. It can be problematic when the guide is sub-optimal.

\paragraph{Reinforcement Learning from Demonstrations (RLfD)} Instead of copying and generalizing the guide policy from the provided dataset, RLfD \citep{schaal1996learning, ravichandar2020recent} uses RL tools to find a better policy. In practice, the additional data is mostly used to overcome unnecessary exploration and lead the agent towards interesting parts of the MDP \citep{nair2018overcoming}. They were originally employed to initialize the policy with BC \citep{peters2008reinforcement, kober2008policy}, and then throughout the entire optimization process. For example, DQNfD \citep{hester2018deep} and DDPGfD \citep{vecerik2017leveraging} include the demonstration data in the replay buffer and add new regularizations to force the optimization process to better consider the demonstrations. In a similar fashion, POfD \citep{kang2018policy} and LOGO \citep{rengarajan2022reinforcement} force the policy of the agent to stay close to the guide policy by penalizing or constraining the RL objective. However, these methods are only possible when the expert policy is attractive as no data could be retrieved near states having catastrophic consequences. Besides, they might require substantial data in order to guide the RL agent \citep{hester2018deep, nair2018overcoming, rengarajan2022reinforcement}, which can be difficult to obtain on systems that are costly to sample.

\paragraph{Reward Shaping} An additional intuitive idea to include domain knowledge in the RL process is to modify the reward function \citep{mataric1994reward, dorigo1994robot}, but an arbitrary reward shaping is dangerous as it might deviate the agent from its original goal \citep{randlov1998learning}. Thus, \citet{ng1999policy} proposed Potential Based Reward Shaping, a framework that constrains the added rewards to be in a specific potential form to guarantee for the original goal to be unchanged, though finding such potential from any external knowledge is difficult \citep{HarutyunyanDVN15}. A clever alternative strategy is to add the heuristic in the $Q$-function instead \citep{knox2010combining, knox2012reinforcement} which has fewer chances of deviating from the original goal, and/or to decrease the impact of the heuristic over time \citep{cheng2021heuristic}.

\paragraph{Reinforcement Learning with a Global Guide} Closely related to our work is to transfer the knowledge of a known teacher \citep{zimmer2014teacher} - or guide policy - to the RL agent \citep{schmitt2018kickstarting, uchendu2022jump}. Despite the need of building a global guide, this circumvents the problem of acquiring a large amount of data. Similarly than in RLfD, the guide policy leads the RL agent in its first learning stages \citep{knox2012reinforcement, schmitt2018kickstarting}. More recently, \citet{uchendu2022jump} and \citet{agarwal2022beyond} took advantage of the access to a global guide to properly initialize an RL policy, even though \citep{agarwal2022beyond} introduced QDagger, an algorithm that is pre-trained on the data sampled by the teacher that could be unavailable in practice. Rather than finding a global policy that would be better than the global guide, \citet{jacq2022lazy} proposed the Lazy-MDPs framework to learn to take over the guide only when it is noticeably sub-optimal. While it provides interesting insights into the properties of the environment, it does not prioritize sample efficiency.

\paragraph{Local Expertise} All the above methods rely on a global guide or heuristic that can be difficult to acquire. Very few works focused on local ones. One approach is COG \citep{singh2020cog} which leverages data from a sub-task to guide the agent in the RL optimization process. However, it suffers from the drawbacks of relying on data discussed in the previous paragraph and is only applied in the sparse reward setting. Most related to our work is to directly consider having access to local guides. It has been studied in two cases. First, when the local controller is optimal, as the one for those built with Optimal Control (OC) theory \citep{Khalilbook, mayne2000constrained, lewis2012optimal}, some works proposed a switching mechanism between the RL and the local controller \citep{zoboli2021reinforcement, gillen2020combining, lee2018composing, byun2021training}. Such switching cannot overcome the sub-optimality of the local policy. Another line of research considered emergency procedures that are activated when the agent enters a dangerous part of the environment to prevent the agent from entering catastrophic states \citep{turchetta2020safe, wagener2021safe}. In addition to the local controller, \citet{turchetta2020safe} assume a curriculum is available to maintain safety, and \citet{wagener2021safe} rely on an advantage estimate of the local policy which might not be available in practice. Contrary to these works, our method does not require any additional heavy-to-build knowledge.

\section{Preliminaries and Problem Setting}

\subsection{Preliminaries}\label{preliminaries}

Let $\Delta(\mathcal{X})$ be the set of all probability measures on $\mathcal{X}$. The agent-environment framework is modeled as a Markov Decision Process (MDP) $\left( \mathcal{S}, \mathcal{A}, r, P, \rho, \gamma \right)$. It is composed of a state space $\mathcal{S}$, an action space $\mathcal{A}$, a transition kernel $P : \mathcal{S} \times \mathcal{A} \to \Delta(\mathcal{S})$, a reward function $r : \mathcal{S} \times \mathcal{A} \to [R_{min}, R_{max}]$, an initial state distribution $\rho$ and a discount factor $\gamma\in \left[0,1\right)$. We focus on the general setting where $\mathcal{A}$ is continuous and propose an extension to handle a discrete action space in Appendix~\ref{app:discrete}.

A policy $\pi : \mathcal{S} \to \Delta(\mathcal{A})$ is a decision rule mapping a state over a distribution of actions. The RL objective is to find a policy maximizing the expected discounted cumulative reward $G_t = \sum_{i=0}^\infty \gamma^t r(s_{t+i}, a_{t+i})$ over the distribution induced by the policy $\pi$ and the transition kernel $P$. The value of a policy $\pi$ is measured through the value function $V^\pi(s) = \mathbb{E}_P \left[ G_t \ | \ s_t=s, \ a_{t+i} \sim\pi(\cdot | s_{t+1}) \forall i \geq 0) \right]$ and its associated $Q$-value function $Q^\pi(s,a) = \mathbb{E}_P \left[ G_t \ | \ s_t=s, \ a_t=a, \right. \\ \left. a_{t+i} \sim\pi(\cdot | s_{t+i}) \ \forall i \geq 1 \right]$. Let $V^*$ and $Q^*$ be the optimal value functions associated with the highest expected cumulative rewards.

Let the Bellman operator for the $Q$-value function $\mathcal{B}^\pi\left[Q\right](s,a) = r(s,a) + \gamma \mathbb{E}_{s'\sim P(\cdot|s,a), a'\sim \pi(\cdot|s')} \left[ Q(s',a') \right]$, and the Bellman optimality operator $\mathcal{B}^{*}\left[Q\right](s,a) = r(s,a) + \gamma \mathbb{E}_{s'\sim P(\cdot|s,a)} \left[ \max_{a'} Q(s',a') \right]$. Being $\gamma$-contractions, iteratively applying these operators to any initial $Q^0$ converges to their associated fixed points $Q^\pi$ or $Q^{*}$.

\subsection{Approximate Policy Iteration}

In view of the usual high dimensions of the state-action space and the lack of access to the environment and its transition kernel, these operators cannot be computed. Instead, the RL approach uses data to approximate them. We denote $\hat{\mathcal{B}}^\pi$ and $\hat{\mathcal{B}}^{*}$ the associated empirical Bellman operators that use samples to estimate the expectation under $P(\cdot|s, a)$. A general framework to find a good policy is Approximate Policy Iteration, which is at the core of many state-of-the-art algorithms \citep{lillicrap2015continuous, mnih2016asynchronous, schulman2017proximal, fujimoto2018addressing, haarnoja2018soft} that are efficient on environments with a continuous action space. As we will see in the following section, such scheme is particularly relevant to the RRLG setting.


At each epoch $k$, the agent collects data with the current policy $\pi^k$, evaluates its associated $Q$-function via Approximate Policy Evaluation ~(\ref{eq:policy_eval}) and improves its policy through Approximate Policy Improvement~(\ref{eq:policy_improvement}). Given a dataset $\mathcal{D} = \{ (s_i, a_i, r_i, s_{i+1})_{i=1}^N \}$, $\hat{\mathbb{E}}$ the empirical expectation of the state-action pair $(s, a)$ induced by $\mathcal{D}$, $\omega\in\Omega$ and $\theta\in\Theta$ the respective parameters of $Q$ and $\pi$, $\bar{\omega}_k \in\Omega$ the frozen weights associated to the $Q$-target, it is formalized as:

\begin{align}
    & Q_\omega^{{k+1}} \leftarrow \arg\min_{\omega\in\Omega} \,\hat{\mathbb{E}} \left[ \left( Q_\omega - \hat{\mathcal{B}}^{\pi_{\theta}^k}\left[ Q_{\bar{\omega}_k}^k \right] \right)^2 \right] \tag{APE},\label{eq:policy_eval} \\
    & \pi_\theta^{k+1} \leftarrow \arg\max_{\theta\in\Theta} \, J_{\pi_\theta}^{\mathcal{D}}\left(Q_{\omega}^{k+1}\right) \tag{API}\label{eq:policy_improvement}.
\end{align}

The objective $J_{\pi_\theta}^{\mathcal{D}}\left(Q_\omega^{k+1}\right)$ in the Approximate Policy Improvement step \ref{eq:policy_improvement} may vary depending on the RL algorithm, but is always a function of the estimated $Q$-values.


Using this process greedily could lead to poor policies \citep{SuttonB98, jin2018q}, especially at the beginning of learning when the estimates are inaccurate. The agents thus need to explore the environment to gather relevant data that would improve these estimations. Since little information about the environment is known, this exploration often relies on random noise added to the policy. It allows the discovery of interesting parts of the environment, but can also very well guide the agent towards meaningless regions. Even worse, this random exploration might lead the agent to catastrophic states that would be unacceptable in real-world systems.

\begin{table*}[t!]
\centering
\begin{tabular}{|l|c|c|c|c|}
  \hline
  Method & Good initialization & Hyper-parameter tuning & Overcome bad guide & Safe agent \\ \hline
  Strict Action Guided (SAG) & \cmark & \cmark & \xmark & \cmark \\ \hline
  Reward Guided (RG) & \xmark & \xmark & \cmark & \xmark \\ \hline
  Policy Improvement Guided (PIG) & \xmark & \xmark & \cmark & \xmark \\ \hline
  Parameterized Action Guided (PAG) & \cmark & \cmark & \cmark & \cmark \\ \hline
\end{tabular} 
\caption{Advantages and Drawbacks of the different agents in the Reinforcement Learning with Local Guides (RLLG) setting}%
\vskip -0.15in
\end{table*}

\subsection{Problem setting}


The goal is to find a good policy on the entire state space with as few interactions as possible with the environment while avoiding catastrophic states in a setting where the system has been previously studied by experts. The idea is to integrate local state space expertise from, \textit{e.g.} any available source, to lead and possibly constrain the RL optimization process. This information is formalized as a local guide $\pi_{\text{g}} : \mathcal{S} \to \Delta\left(\mathcal{A}\right)$ that is relevant only in a potentially small region of the state space $\mathcal{S}_{\text{g}}\in\mathcal{S}$. Note this controller may be the concatenation of $N$ different local guides that are relevant in non-overlapping regions of the state space ($S_{\text{g}}^{i}$ for expert $i$), in which case $\mathcal{S}_{\text{g}} = \bigcup_{i=1}^N S_{\text{g}}^{i}$.

Along with the local heuristic, we consider having access to a function $\lambda : \mathcal{S} \to \left[ 0,1 \right]$ reporting the confidence of the local guide in state $s$. This could be a binary function when perfect knowledge about the local policy and the environment is known, where $\lambda(s) = 1$ if $s\in\mathcal{S}_{\text{g}}$ else $\lambda(s) = 0$. While this parameter may not be known with precision, it can be estimated by practitioners.

We do not assume that the local guide is optimal as this is not necessarily the case in real-world applications. This relaxes the need to know with precision the confidence function $\lambda$. Additionally, we do not assume a combination of the local guides can solve the task, hence the necessity of introducing RL.

In addition, this work is agnostic to the type of the guide policy, \textit{e.g.} stochastic or deterministic, but for the rest of the paper, we consider the realistic setting where the guide policy is deterministic that may be prevalent in the real world. It is unlikely that experts design randomized policies that are then applied in real-world systems: actions of the tail distribution of the controller may harm the system. We thus denote $a_{\text{g}}^s$ the only action the guide policy outputs in the state $s$. The extension to stochastic policies is straightforward as $a_{\text{g}}^s$ could be replaced with an expectation over $\pi_{\text{g}}(\cdot|s)$.

\section{Reinforcement Learning with Local Guides}

In this section, we first introduce three straightforward ways to adapt APE RL algorithms to our setting and analyze their advantages and disadvantages. We then propose a new simple yet efficient approach to retain the identified advantages while avoiding the presented drawbacks.

\subsection{Classical integration of the local controller}

\paragraph{Strict Action Guided (SAG)} To include the local controller in this setting, one can simply use it when available. Thanks to the provided indicator function $\lambda(\cdot)$, the agent will switch between the different policies when the guide is judged sufficiently relevant, that is when $\lambda(s) \geq \lambda^-$. The global policy at iteration $k$, $\pi^k_{\text{SAG}}$, can be written:

\begin{equation}
\pi^k_{\text{SAG}}(\cdot | s) = \left\{
    \begin{array}{ll}
        a_{\text{g}}^s & \mbox{if } \lambda(s) \geq \lambda^-, \\
        \pi^k_\theta(\cdot | s) & \mbox{otherwise.}
    \end{array}
\right.
\end{equation}

This method is most appropriate when the local policy is optimal, or when the sub-optimality of the local controller is deemed sufficient for the problem at hand. Applying this switching mechanism seems intuitive and has been successfully used in prior works \citep{gillen2020combining, zoboli2021reinforcement}. However, the use of Deep Neural Networks combined with a bootstrapped target loss in the (\ref{eq:policy_eval}) step might lead to a distributional shift previously studied in the offline setting \citep{fujimoto2018addressing, kumar2020conservative, levine2020offline}. Indeed, when evaluating the $Q$-function of a policy $\pi$ with a dataset that has been gathered with a policy far from the evaluated one, using $\pi$ to bootstrap its target would wrongly and continually back-propagate bad estimates. To cope with this issue, we propose to estimate the $Q$-values of the switched policy $\pi_{\text{SAG}}^k$ instead of $\pi^k_\theta$, simply by building the target with $\hat{\mathcal{B}}^{\pi_{\text{SAG}}^k}$ instead of $\hat{\mathcal{B}}^{\pi^k_\theta}$. This step would be impossible in the Approximate Value Iteration scheme, comforting our choice of focusing on API-based algorithms.

\paragraph{Reward Guided (RG)} While the previous policy definitely provides a boost in learning, especially for initializing the policy and making sure it respects the potential constraints of the environment, the agent is stuck with the potential sub-optimality of the guide. A first way to cope with this is to shape the reward to include the local information \citep{knox2010combining, knox2012reinforcement}. Let $\mathcal{M}^{\pi_{\text{g}}}_{\pi^k_\theta}(s)$ be a behavioral cloning (BC) metric that reports how close the policy of the agent $\pi^k_\theta$ is to the guide policy $\pi_{\text{g}}$. Different BC metrics can be used depending on the considered setting,  \textit{e.g.} $- \| a - a_{\text{g}}^s\|^2$ when $\pi^k_\theta$ is deterministic, or $ \log \pi^k_\theta(a_{\text{g}}^s|s)$ when the density of the policy is available. Given these notations, considering a hand-crafted scheduler $\beta^k_{\text{RG}}$, the shaped reward can be written as:

\begin{align}
    \tilde{r}^{k}(s, a, \pi_{\text{g}}, \pi^k_\theta, \lambda) = r(s,a) + \beta^k_{\text{RG}} \ \lambda(s) \  \mathcal{M}^{\pi_{\text{g}}}_{\pi^k_\theta}(s) .
\end{align}

However, as discussed in the related works section, playing with the rewards is hazardous as it might lead to a completely modified goal \citep{NgHR99}. Besides, the impact of the metric $\mathcal{M}^{\pi_{\text{g}}}_{\pi^k_\theta}(s)$ is difficult to control when it is added to the reward function. This phenomenon was observed in our experiments where we did not get good results, see Appendix~\ref{app:reward_guided_agents}.

\begin{figure*}[ht!]
\centering
\begin{subfigure}{.16\textwidth}
  \centering
  \includegraphics[height=2.5cm, width=2.5cm]{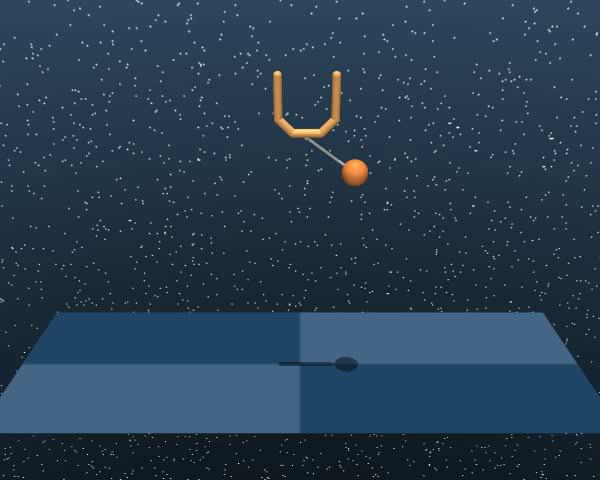}
  \caption{Ball-in-Cup}
  \label{fig:ball_in_cup}
\end{subfigure}%
\begin{subfigure}{.16\textwidth}
  \centering
  \includegraphics[height=2.5cm, width=2.5cm]{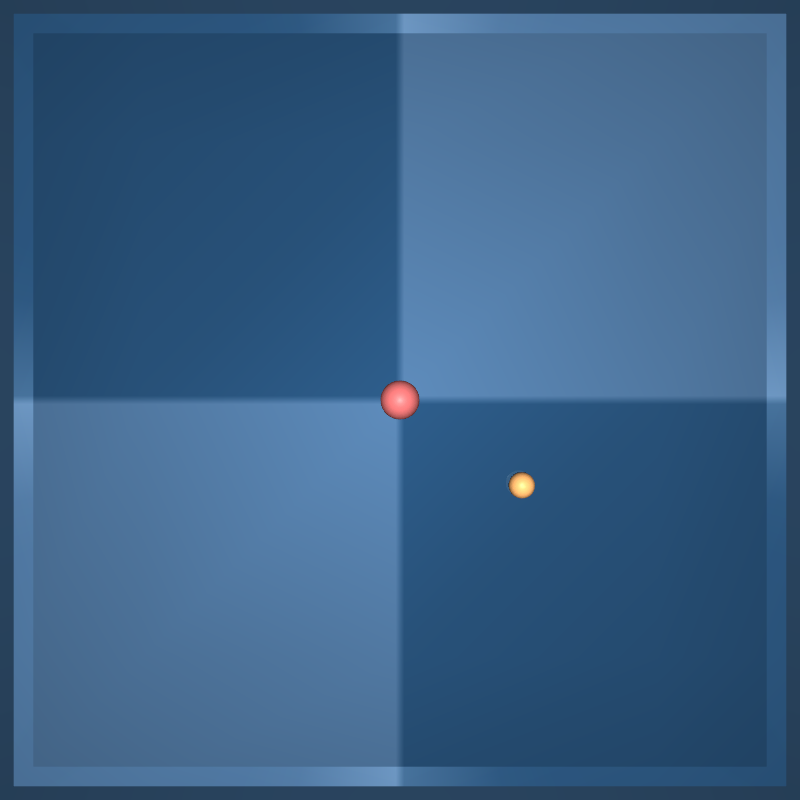}
  \caption{Point-Mass}
  \label{fig:point_mass}
\end{subfigure}%
\begin{subfigure}{.16\textwidth}
  \centering
  \includegraphics[height=2.5cm, width=2cm]{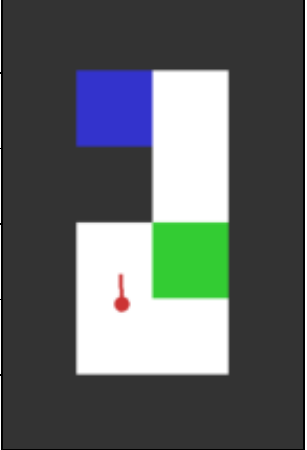}
  \caption{Point-Fall}
  \label{fig:hirl_point_fall_cropped}
\end{subfigure}%
\begin{subfigure}{.16\textwidth}
  \centering
  \includegraphics[height=2.5cm, width=2.5cm]{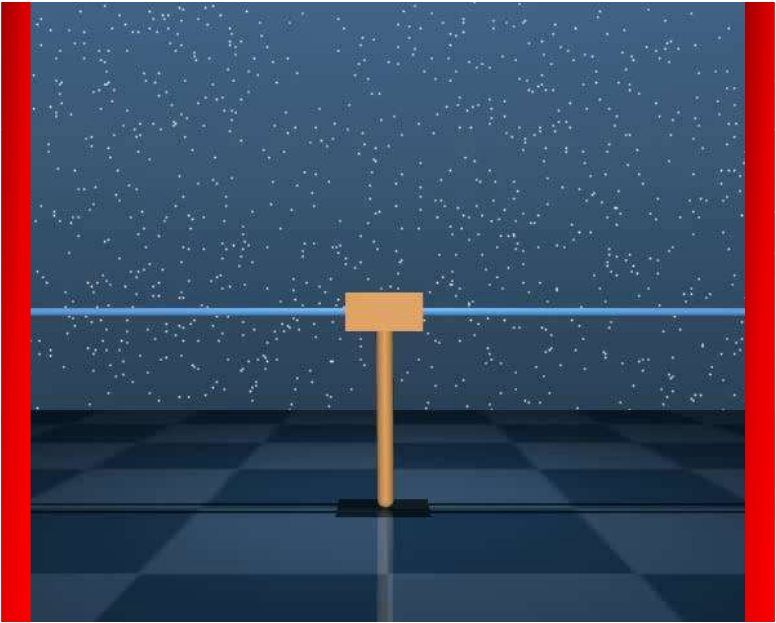}
  \caption{Safe Cartpole}
  \label{fig:safe_cartpole}
\end{subfigure}%
\begin{subfigure}{.16\textwidth}
  \centering
  \includegraphics[height=2.5cm, width=2.5cm]{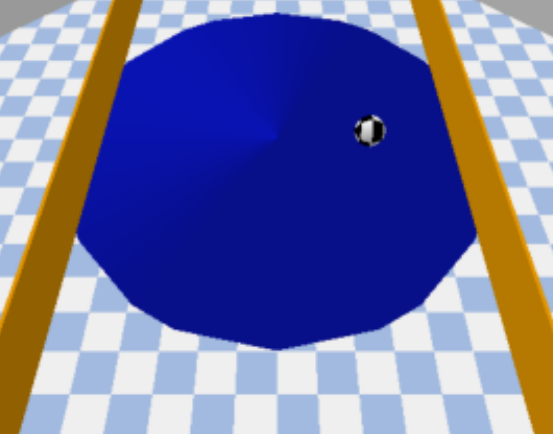}
  \caption{Point-Circle}
  \label{fig:bullet_circle_no_red_new}
\end{subfigure}%
\begin{subfigure}{.16\textwidth}
  \centering
  \includegraphics[height=2.5cm, width=2.5cm]{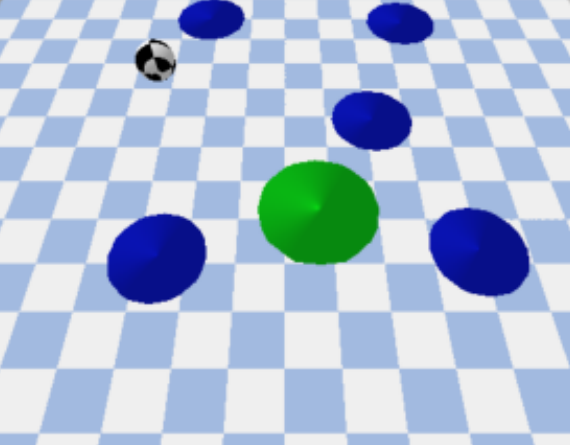}
  \caption{Point-Reach}
  \label{fig:bullet_reach_no_red_new}
\end{subfigure}%
\caption{Environment visualizations.}
  \label{fig:env_visualizations}
\vskip -0.15in
\end{figure*}

\paragraph{Policy Improvement Guided (PIG)} Another approach to guide the RL agent with the local controller is to constrain the optimization process, in particular in the Approximate Policy Improvement step \citep{knox2012reinforcement, wu2019behavior, kumar2019stabilizing}. It is formalized as a Trust Region approach to make sure the agent explores at the right location:

\begin{argmaxi} 
    {\theta\in\Theta}{ J_{\pi_\theta}^{\mathcal{D}}\left(Q_\omega^{k+1}\right)}
{\hspace{-2.8 cm}}{}
\label{opt:control_sharing_eleguant}
\addConstraint{\hat{\mathbb{E}}_{s\sim\mathcal{D}} \left[ \lambda(s) \mathcal{M}^{\pi_{\text{g}}}_{\pi_\theta}(s) \right]}{\leq M_k},
\end{argmaxi}

where $M_k$ is an unknown constant that would have to be determined by the practitioner. However, when learning policies are parameterized by Deep Neural Networks, this optimization problem is difficultly solved analytically. Following the regularized RL framework \citep{geist2019theory}, practitioners resort to solving the relaxed Lagrangian optimization problem:

\begin{equation}
\begin{split}
    \pi_\theta^{k+1} \leftarrow \arg\max_{\theta\in\Theta} \, J_{\pi_\theta}^{\mathcal{D}}\left(Q_\omega^{k+1}\right) + \\ \beta^k_{\text{PIG}} \ 
    \hat{\mathbb{E}}_{s\sim\mathcal{D}} \left[ \lambda(s) \mathcal{M}^{\pi_{\text{g}}}_{\pi_\theta}(s) \right]. 
\end{split}
\end{equation}

In the literature $\beta^k_{\text{PIG}}$ is either kept constant \citep{kumar2019stabilizing, wu2019behavior} or slightly decayed over the episodes \citep{schmitt2018kickstarting, kang2018policy,  agarwal2022beyond}. This controller integration is actually a state-the-art method to guide the RL policy when a global sub-optimal guide is available and was first proposed in \citep{schmitt2018kickstarting}. Other approaches to constrain a policy to remain close to another exist \citep{kostrikov2021offline, rengarajan2022reinforcement}, though they all rely on solving a Lagrangian problem, which would suffer from the same drawbacks as the presented method.

\subsection{Relaxing the guide action strict focus}

In the previous section, we discussed the advantages and drawbacks of the three introduced methods to integrate a local controller: SAG would benefit from a jump start but would not be able to outperform a sub-optimal guide, whereas RG and PIG could eventually improve it. However, in RG, shaping the reward is hard and presents the risk of deviating the agent from its original goal. The exact Trust Region procedure of Equation~(\ref{opt:control_sharing_eleguant}) would be a good solution to answer this problem. Indeed, setting a small $M_k$ at the early stages of learning makes sure the learnt policy begins near the local guide, and gradually increasing it would allow a clever and safe exploration of the environment. However, the relaxed Lagrangian approach is a relaxation of the constrained optimization problem so the agent is very likely to go beyond the trust region. When it is crucial to stay close to the local controller, this is not acceptable.

In this section, we propose a novel method to take the best of all approaches: it has a good policy initialization, sets a well-defined constraint on the closeness between the learnt and guide policies while still being able to overcome the sub-optimality of the guide. Note that our agent takes advantage of the continuous structure of the action space, but we propose an extension to the discrete setting in Appendix~\ref{app:discrete}.

\paragraph{Perturbed Action Guided (PAG)} We propose to keep the Action Guided approach from SAG that integrates the local controller action into a global policy to enjoy a good initialization. Although, in a similar fashion to \cite{fujimoto2019off}, we introduce a parameterized perturbation $\xi_\phi^k$ to gradually improve the local controller and overcome its limitations. The agent would be able to visit interesting regions of the state space while in practice respecting safety constraints encoded in the guide actions $a_{\text{g}}^s$. Formally, the parameterized perturbation $\xi_\phi(\cdot|s, a^s_{\text{g}}, \Phi) \in \left[ -\Phi, \Phi \right]$, with $\phi\in\Xi$ takes as arguments the state, the guide action and a bound $\Phi$ over the action space. This perturbation slightly transforms the guide action allowing close exploration and eventually improving the guide policy. The global policy at iteration $k$, $\pi^k_{\text{PAG}}$, can be written:

\begin{equation} \label{eq:guided_switch}
\pi^k_{\text{PAG}}(\cdot | s) = \left\{
    \begin{array}{ll}
        a_{\text{g}}^s + \beta^k_{\text{PAG}} \ \xi_\phi^k(\cdot|s, a_{\text{g}}^s, \Phi)  & \mbox{if } \lambda(s) \geq \lambda^-, \\
        \pi^k_\theta(\cdot | s) & \mbox{otherwise.}
    \end{array}
\right.
\end{equation}

Where $\beta^k_{\text{PAG}}$ is a scheduler introduced to further control the weight of the parameterized perturbation $\xi_\phi^k$. Intuitively, $\beta^k_{\text{PAG}}$ should be set close to $0$ in the early stages of learning to enjoy a good policy initialization thanks to the local controller, and gradually increase to $1$ as the perturbation $\xi_\phi^k$ gets more relevant.

The bound $\Phi$ on the perturbation $\xi_\phi^k$ can be chosen by the practitioner depending on the nature of the environment. For instance, when safety is at stake, practitioners might choose a small $\Phi$ to remain close to the guide actions. When it is not, $\Phi$ could be increased to have a wider exploration and eventually improve the guide policy.

Thanks to the Approximate Policy Iteration structure~(\ref{eq:policy_eval}-\ref{eq:policy_improvement}), $\xi_\phi^k$ can directly be trained to maximize $Q_\omega^{k+1}$, that is a global estimate of the $Q$-values of the global policy $\pi^k_{\text{PAG}}$:

\begin{equation} \label{eq:improved_switch}
    \xi_\phi^{k+1} \leftarrow \arg\max_{\phi\in\Xi} \, J_{\xi_\phi}^{\mathcal{D}}\left(Q^{k+1}_\omega\right),
\end{equation}

with $J_{\xi_\phi}^{\mathcal{D}}\left(Q^{k+1}_\omega\right) = \hat{\mathbb{E}}_{s\sim\mathcal{D}, a'\sim\xi_\phi(\cdot|s, a^s_{\text{g}}, \Phi)} \left[ \lambda(s) \ Q_\omega^{k+1}(s,a_{\text{g}}^s + a') \right]$. See Algorithm~\ref{alg:improved_switch} for a pseudo-code description of our proposed method PAG.


\begin{figure*}
\centering
\begin{subfigure}{.28\textwidth}
  \centering
  \includegraphics[width=.99\linewidth]{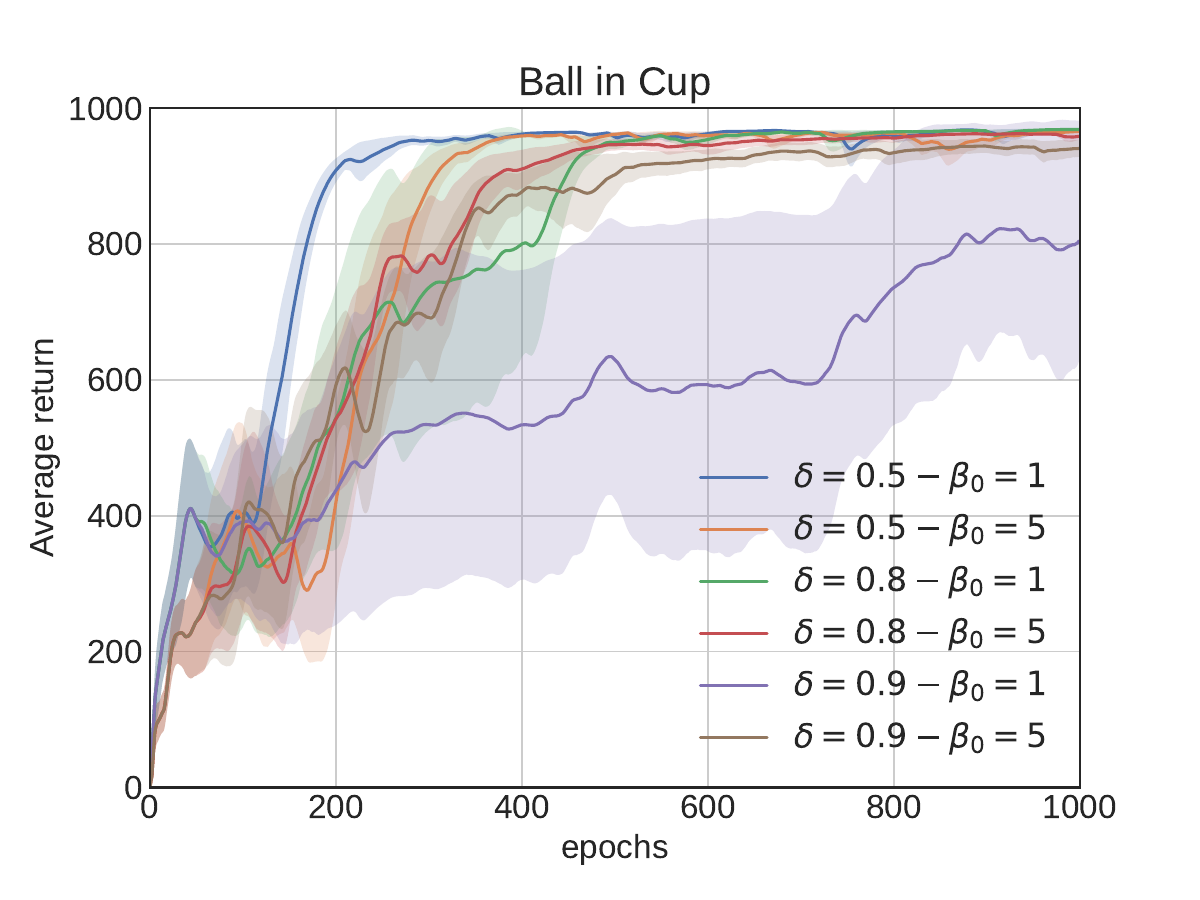}
\end{subfigure}
\vspace{-0.5\baselineskip}
\begin{subfigure}{.28\textwidth}
  \centering
  \includegraphics[width=.99\linewidth]{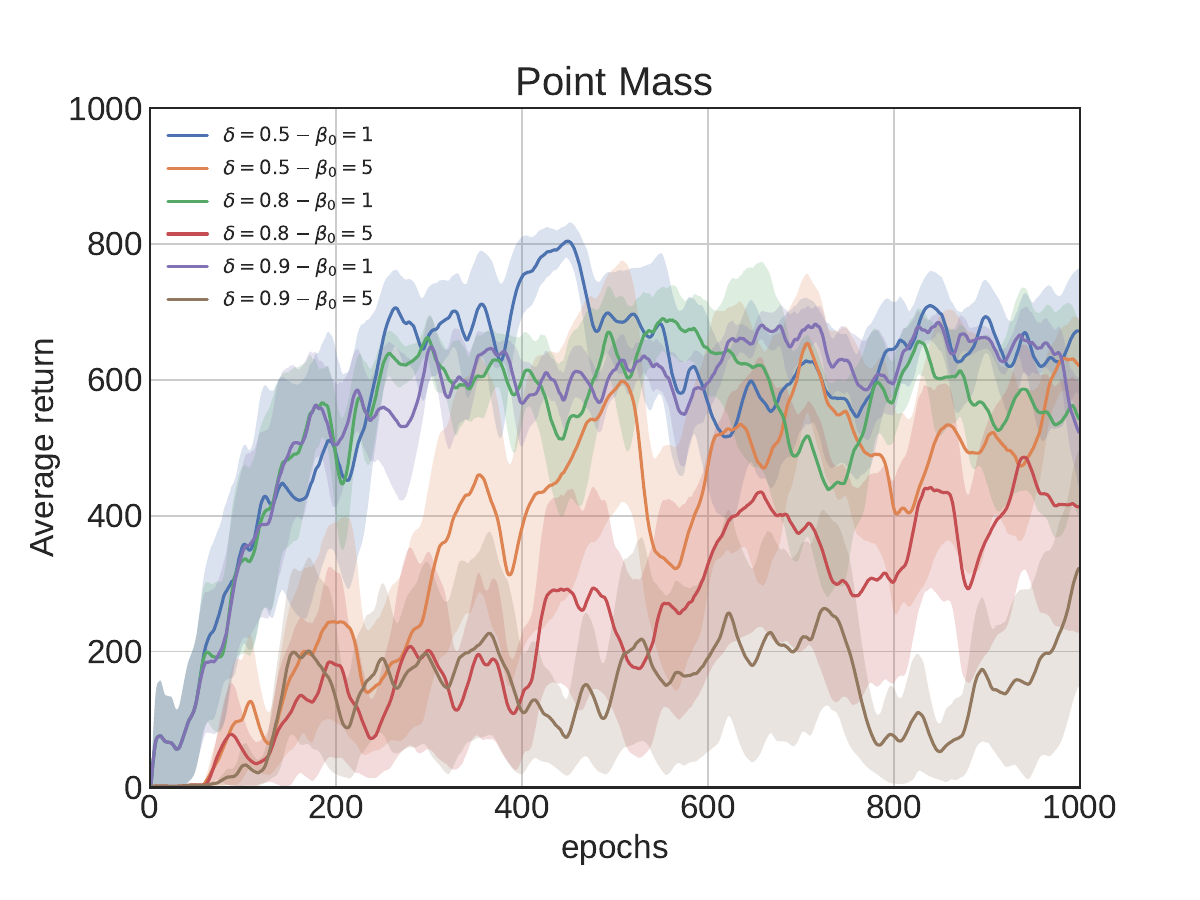}
\end{subfigure}%
\vspace{-0.5\baselineskip}
\begin{subfigure}{.28\textwidth}
  \centering
  \includegraphics[width=.99\linewidth]{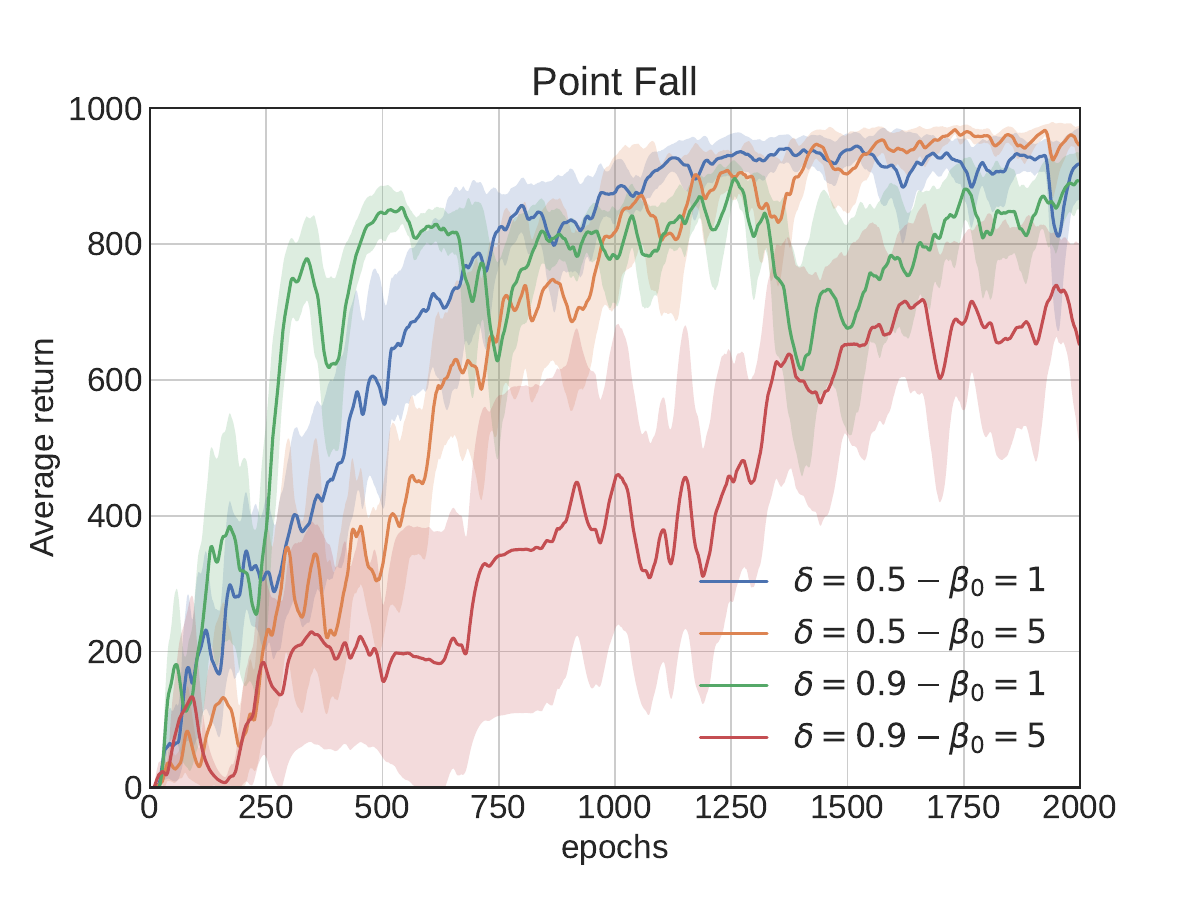}
\end{subfigure} \hspace{\fill} 
\vspace{-0.5\baselineskip}
\begin{subfigure}{.28\textwidth}
  \centering
  \includegraphics[width=.99\linewidth]{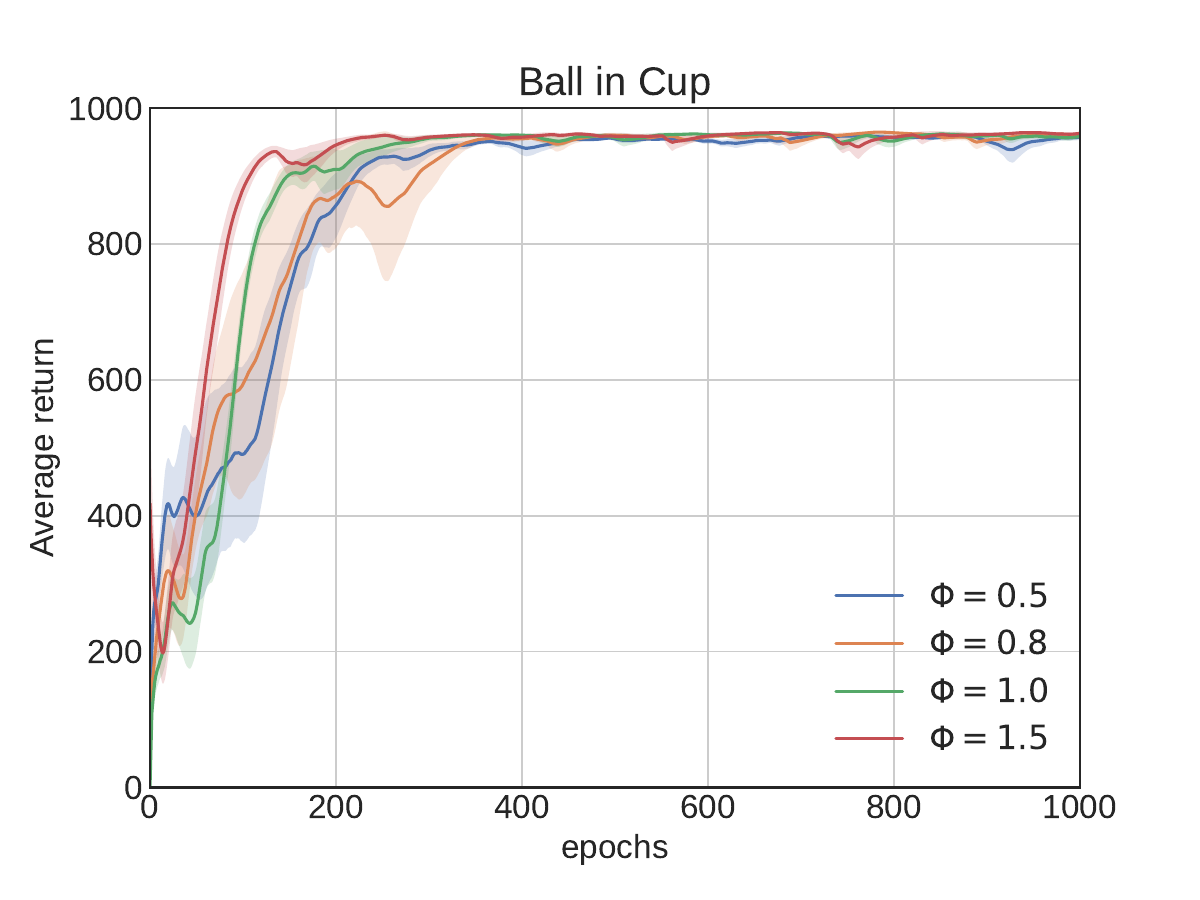}
  \label{fig:ball_in_cup_ablation_policy_decay}
\end{subfigure}%
\vspace{-0.3\baselineskip}
\begin{subfigure}{.28\textwidth}
  \centering
  \includegraphics[width=.99\linewidth]{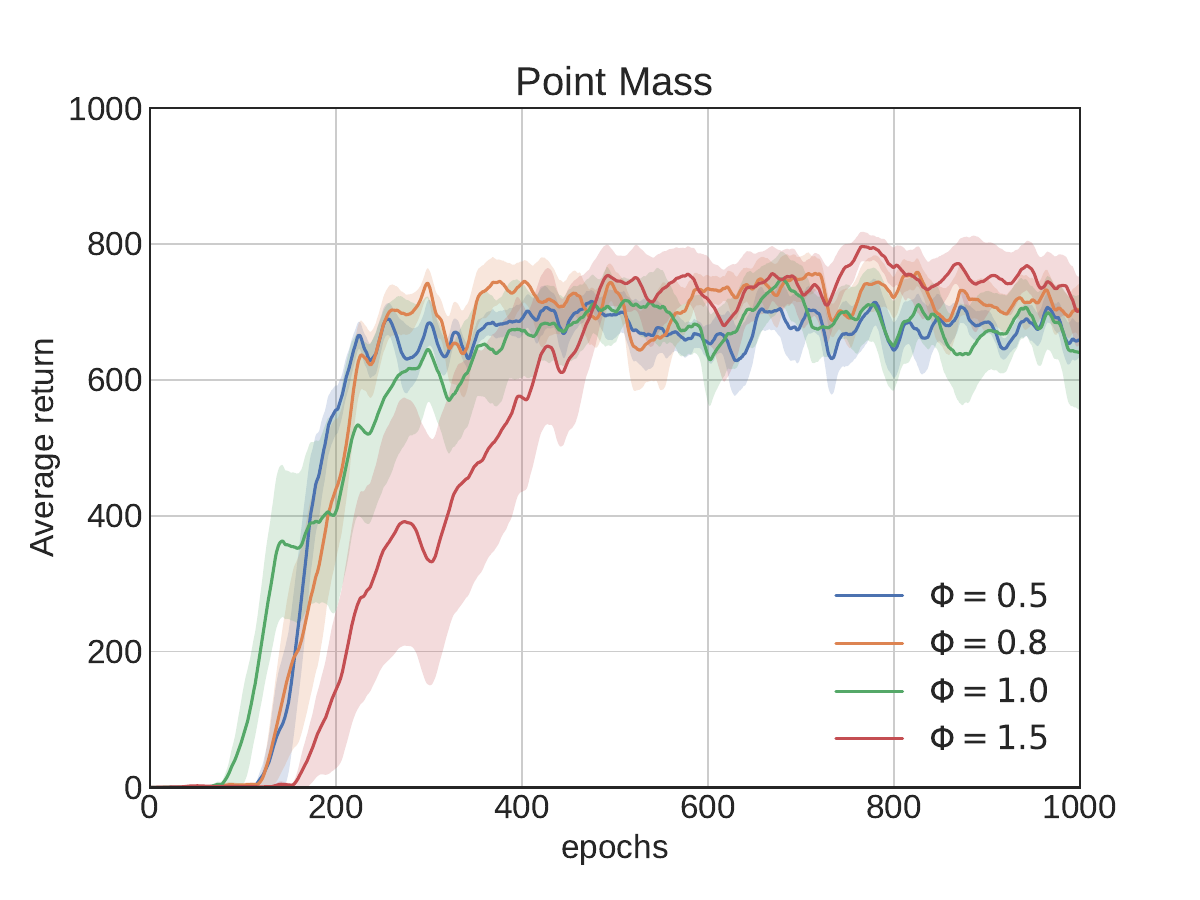}
  \label{fig:point_mass_ablation_policy_decay}
\end{subfigure}%
\vspace{-0.3\baselineskip}
\begin{subfigure}{.28\textwidth}
  \centering
  \includegraphics[width=.99\linewidth]{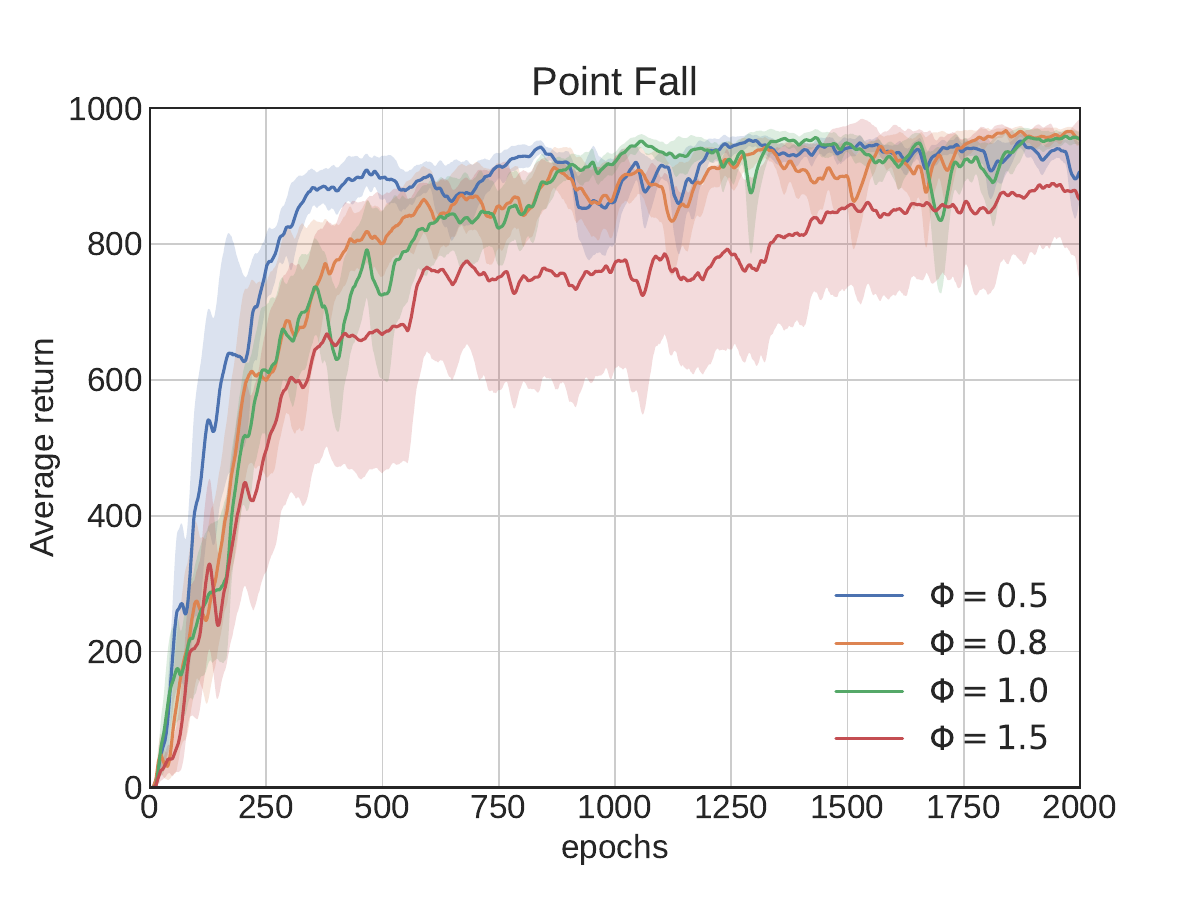}
  \label{fig:point_fall_ablation_policy_decay}
\end{subfigure}%
\vspace{-0.3\baselineskip}
\caption{Hyper-parameter analysis of PIG (top) and PAG (bottom) on environments with attractive policies.}
  \label{fig:ablation_guided_agents}
\end{figure*}

\begin{figure*}
\vspace{-0.5\baselineskip}
\centering
\begin{subfigure}{.28\textwidth}
  \centering
  \includegraphics[width=.99\linewidth]{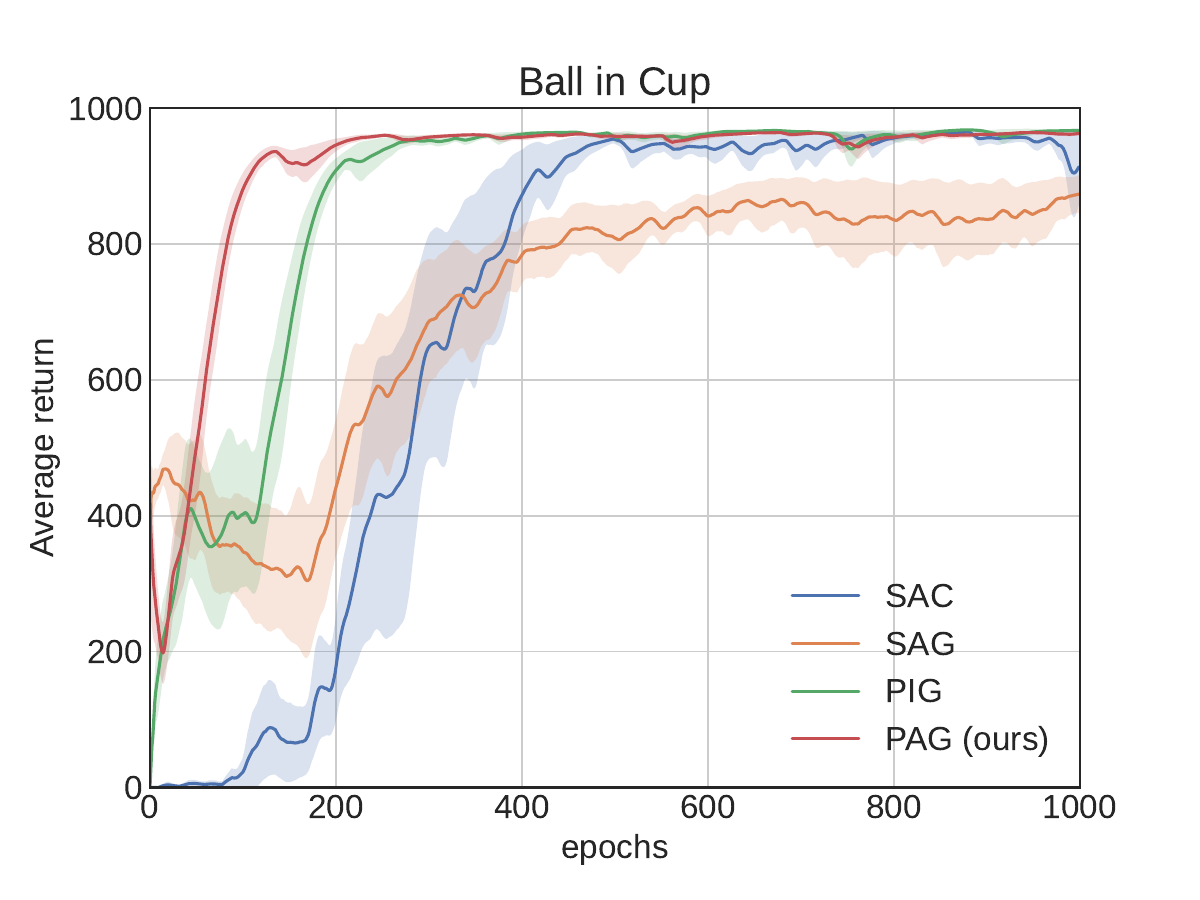}
  \label{fig:ball_in_cup_global}
\end{subfigure}%
\vspace{-0.3\baselineskip}
\begin{subfigure}{.28\textwidth}
  \centering
  \includegraphics[width=.99\linewidth]{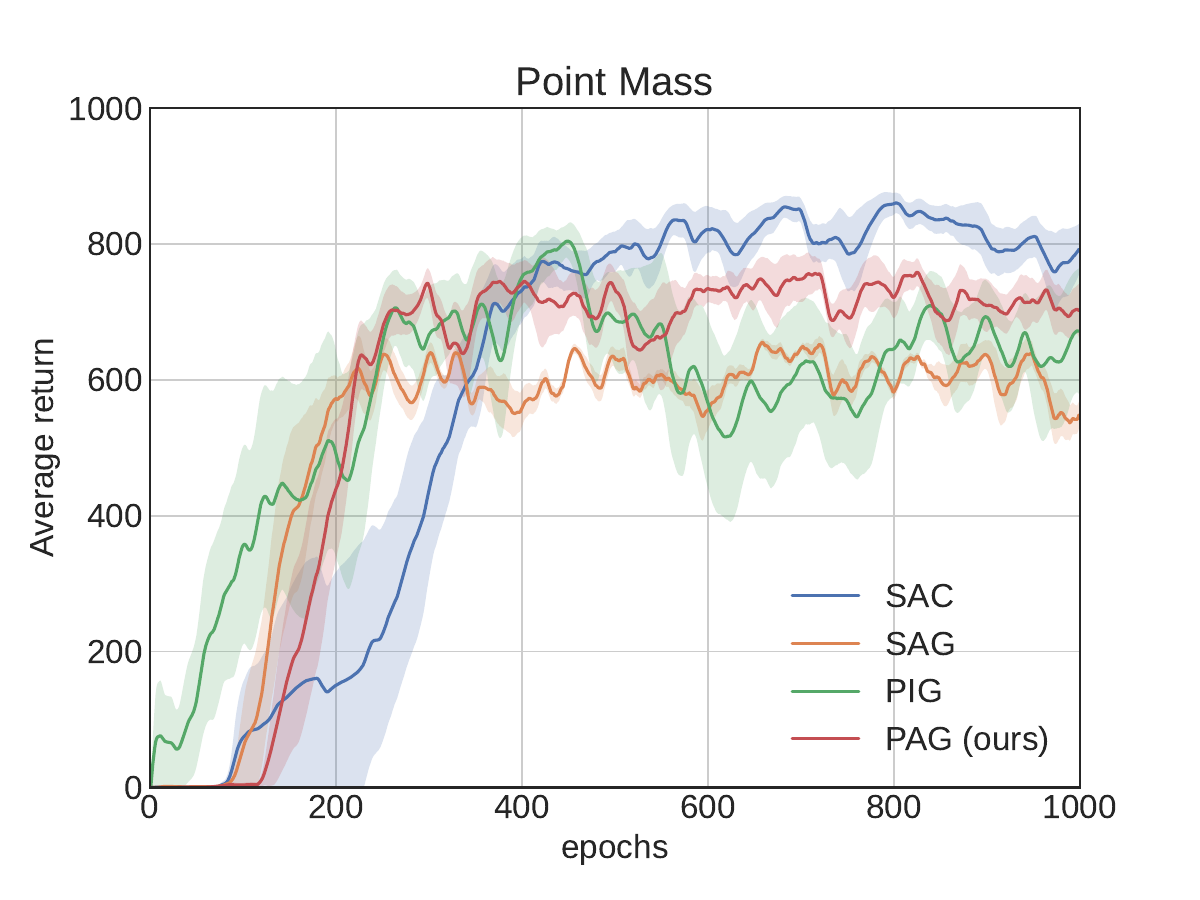}
  \label{fig:point_mass_global}
\end{subfigure}%
\vspace{-0.3\baselineskip}
\begin{subfigure}{.28\textwidth}
  \centering
  \includegraphics[width=.99\linewidth]{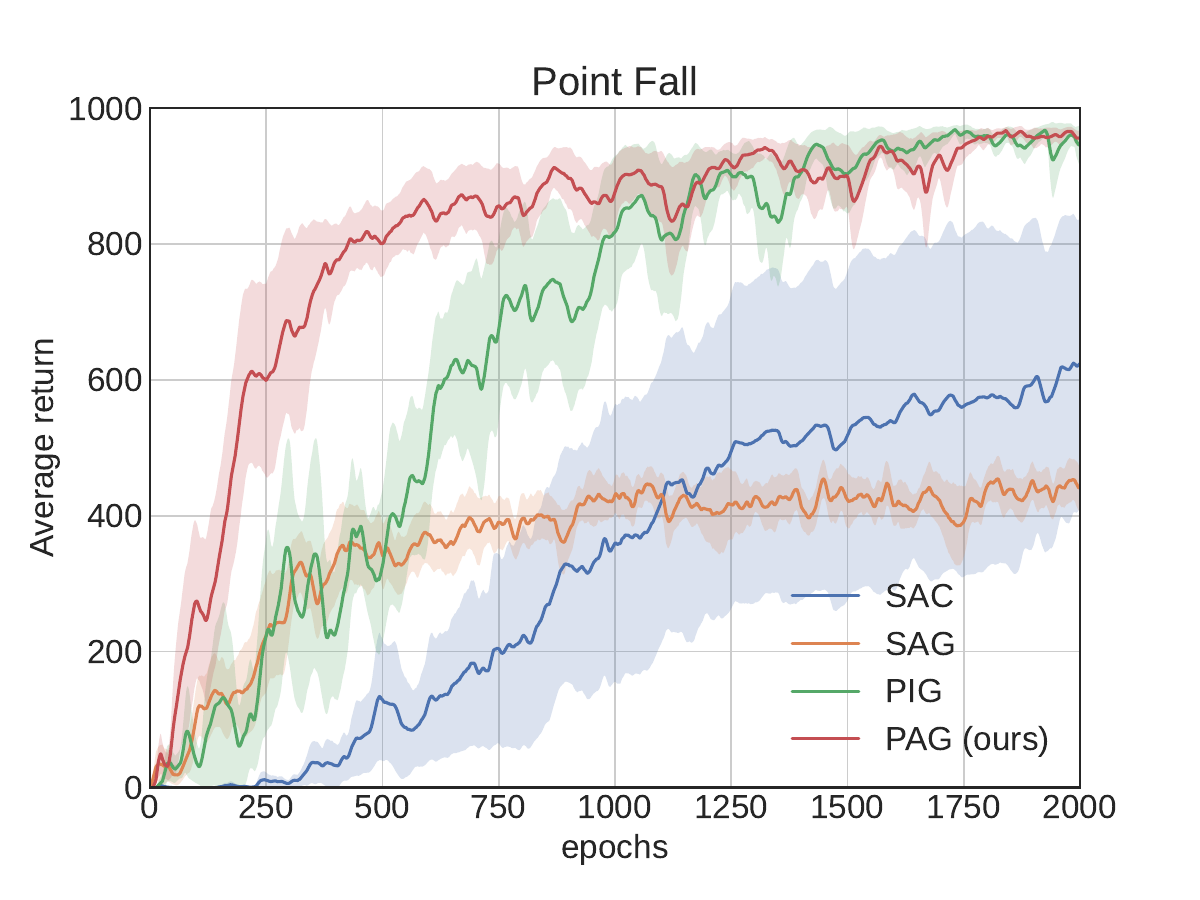}
  \label{fig:manipulator_glob}
\end{subfigure}%
\vspace{-0.3\baselineskip}
\caption{Overall performances comparing PAG with SAC, SAG and PIG on 3 different environments with attractive policies.}
  \label{fig:global_results_guided}
\end{figure*}

\begin{algorithm}[H]
\begin{algorithmic}
\STATE Initialize $Q^0_\omega$,  $\pi^0_\theta$, and $\xi^0_\phi$
\FOR {$k \in (1, \dots, K)$} 
    \STATE Gather data with $\pi^k_{\text{PAG}}$ from Eq.~(\ref{eq:guided_switch})
    \STATE Add data to the replay buffer $\mathcal{D}$
    \STATE Sample a batch from $\mathcal{D}$
    \STATE Update $Q^{k+1}_\omega$ with gradient descent on Eq.~(\ref{eq:policy_eval}) with $\hat{\mathcal{B}}^{\pi^k_{PAG}}$
    \STATE Update $\pi^{k+1}_\theta$ with gradient ascent on Eq.~(\ref{eq:policy_improvement})
    \STATE Update $\xi^{k+1}_\phi$ with gradient ascent on Eq.~(\ref{eq:improved_switch})
\ENDFOR
\end{algorithmic}
\caption{Perturbed Action Guided (PAG)}
\label{alg:improved_switch}
\end{algorithm}

\section{Experiments}

In this section, we evaluate the performances of previously introduced methods that integrate local controllers in the RL framework in two different settings. In the first one, the guide leads the decision-making process to maximize the performance of the agent. We denote those local controllers \textit{attractive policies}. In the second one, the guide should prevent the agent from entering dangerous zones of the environment using conservative repulsive guide policies. We denote those local controllers \textit{repulsive policies}.

\paragraph{Figure description} In all figures, the x-axis represents the number of epochs from $0$ to $K$, which corresponds to $1000$ interactions with the system as well as $1000$ gradient updates of the different networks. The y-axis represents the averaged cumulative return over $5$ evaluation trials. All plots represent an exponential smoothed average of $5$ runs, and the shaded areas correspond to half of their standard deviation.

\paragraph{Agents} We conduct our experiments using Soft Actor Critic (SAC) \citep{haarnoja2018soft}, although our method can be used on any Approximate Policy Evaluation-based RL algorithm. Reward-Guided agents' performances are deferred in Appendix~\ref{app:reward_guided_agents} as they did not provide interesting results.

\paragraph{Chosen hyper-parameters} All hyper-parameters are based on the default hyper-parameters of SAC except for the size of the hidden layers and the activation functions of the networks. Both the policy and the $Q$-functions are Feed-Forward Neural Networks with $2$ layers of $64$ neurons for all the guided environments and $2$ layers of $32$ neurons for the safety-critical environments. Besides, all networks have a ReLU activation with the exception of the $Q$-Network used in \textit{Safe Cartpole Swingup} that has a TanH activation which produced better results. All additional hyperparameters were optimized using a grid search, and are detailed in the next sections. The Pytorch code of this work can be found in \url{https://github.com/huawei-noah/HEBO/tree/master/RLLG}.

\begin{figure*}
\centering
\begin{subfigure}{.28\textwidth}
  \centering
  \includegraphics[width=.99\linewidth]{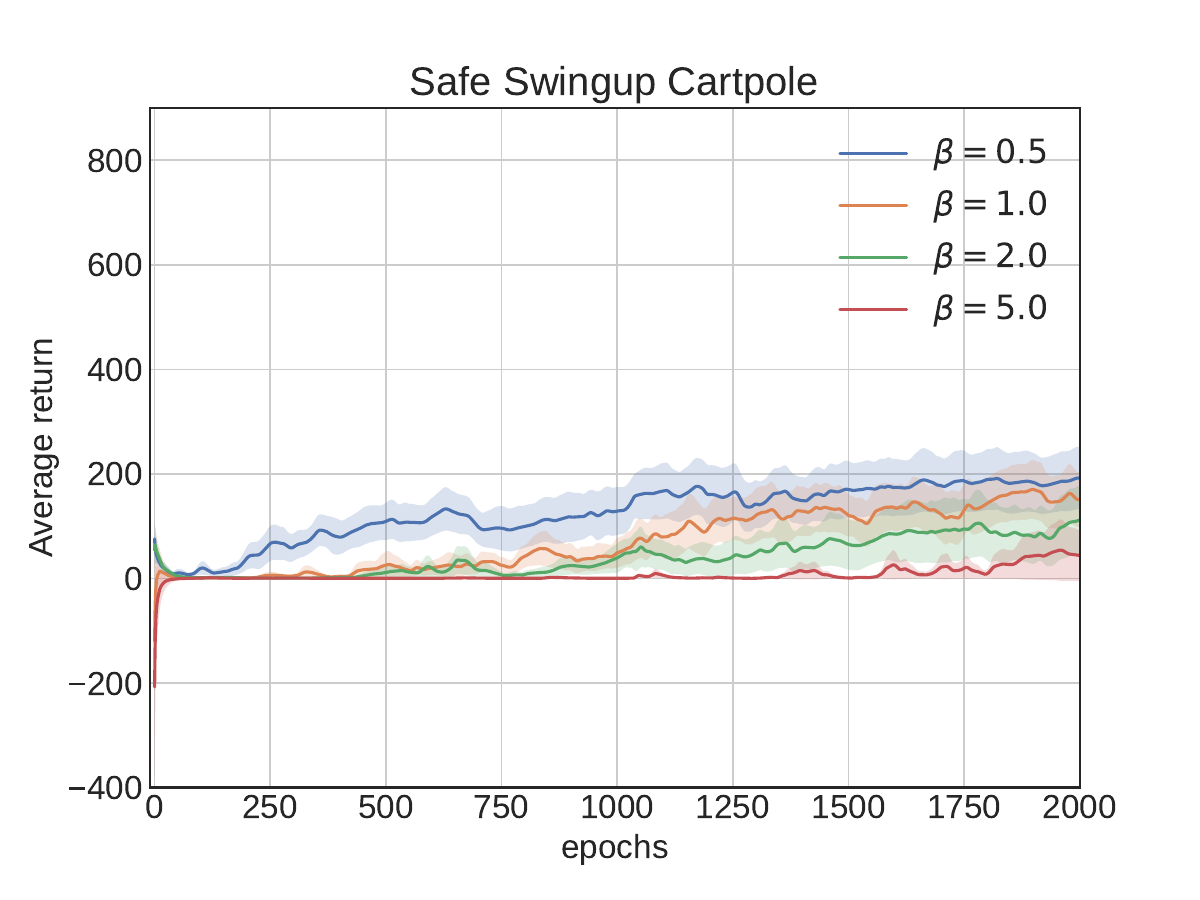}
\end{subfigure}%
\vspace{-0.5\baselineskip}
\begin{subfigure}{.28\textwidth}
  \centering
  \includegraphics[width=.99\linewidth]{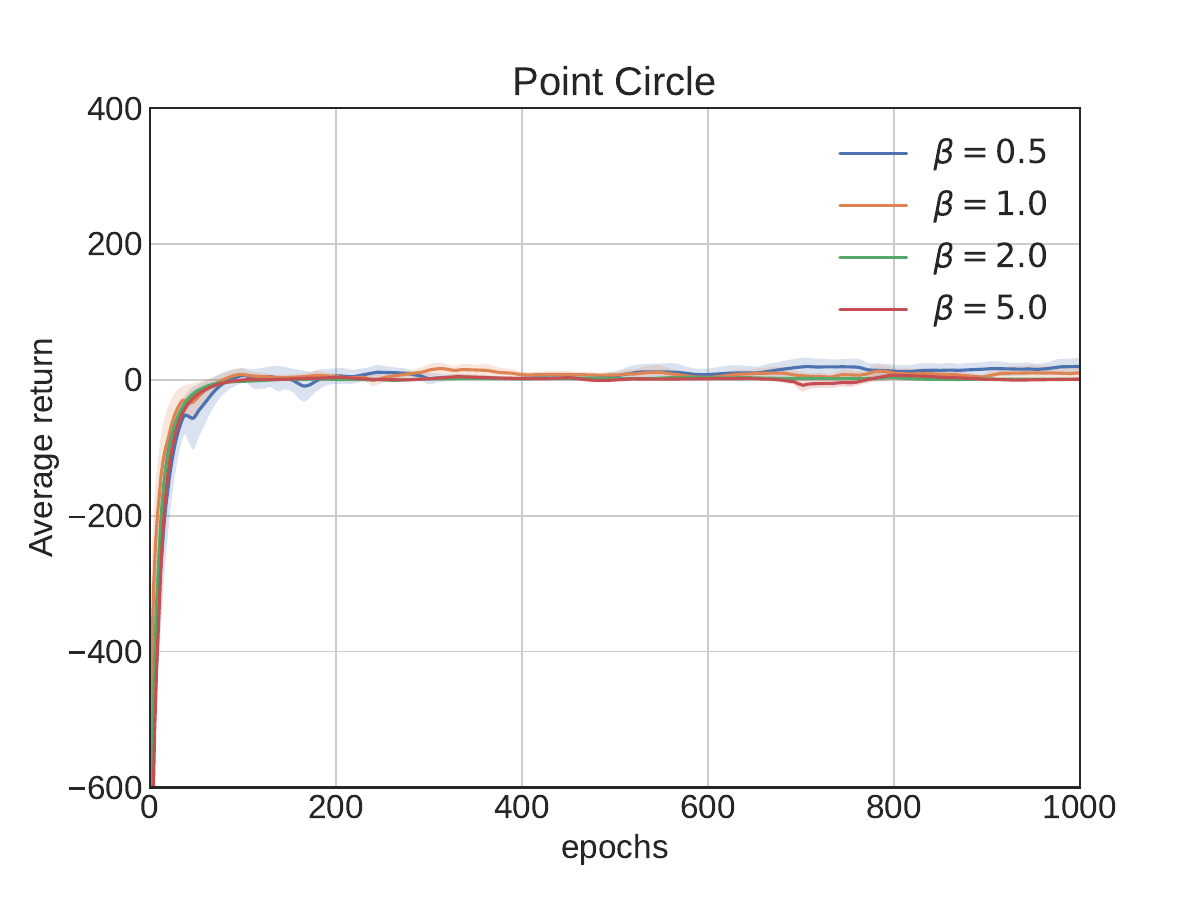}
\end{subfigure}%
\vspace{-0.5\baselineskip}
\begin{subfigure}{.28\textwidth}
  \centering
  \includegraphics[width=.99\linewidth]{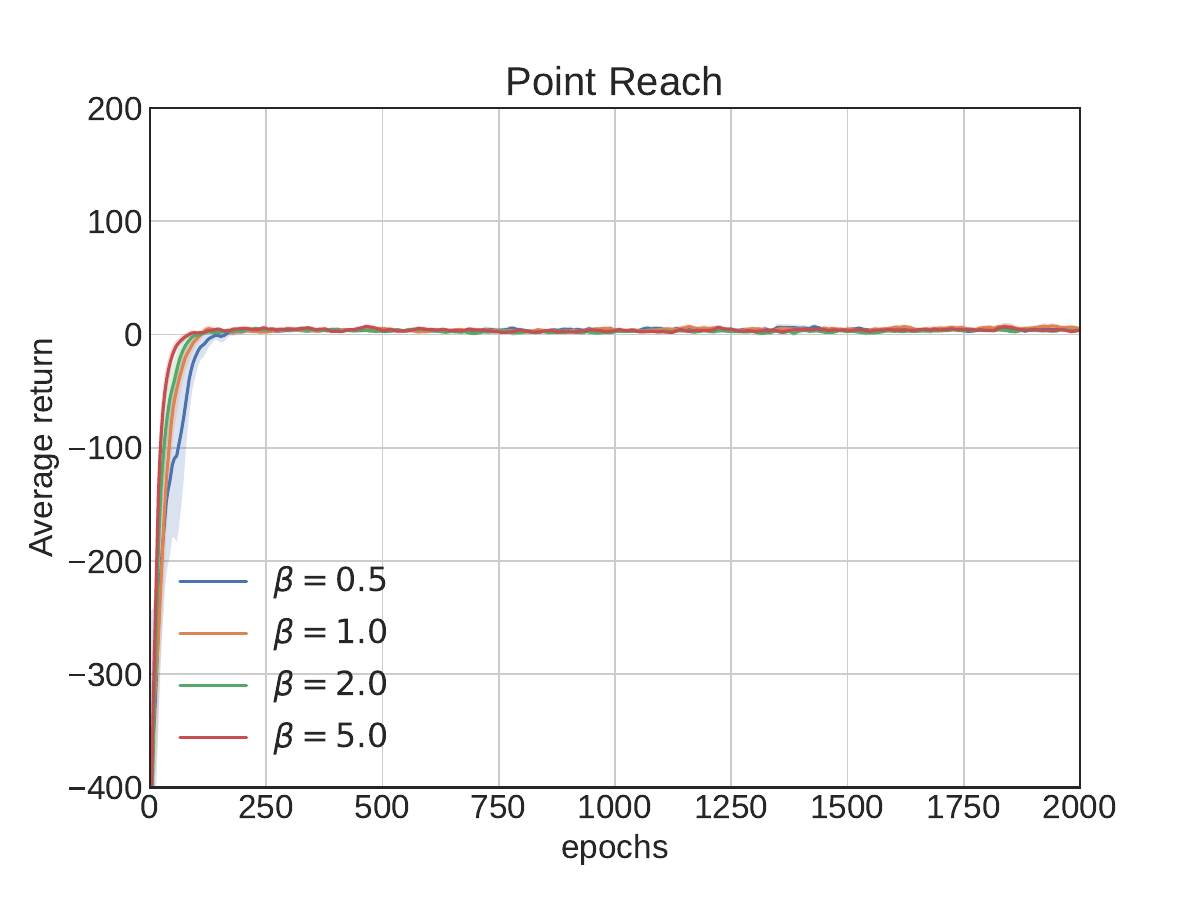}
\end{subfigure} \hspace{\fill}
\begin{subfigure}{.28\textwidth}
  \centering
  \includegraphics[width=.99\linewidth]{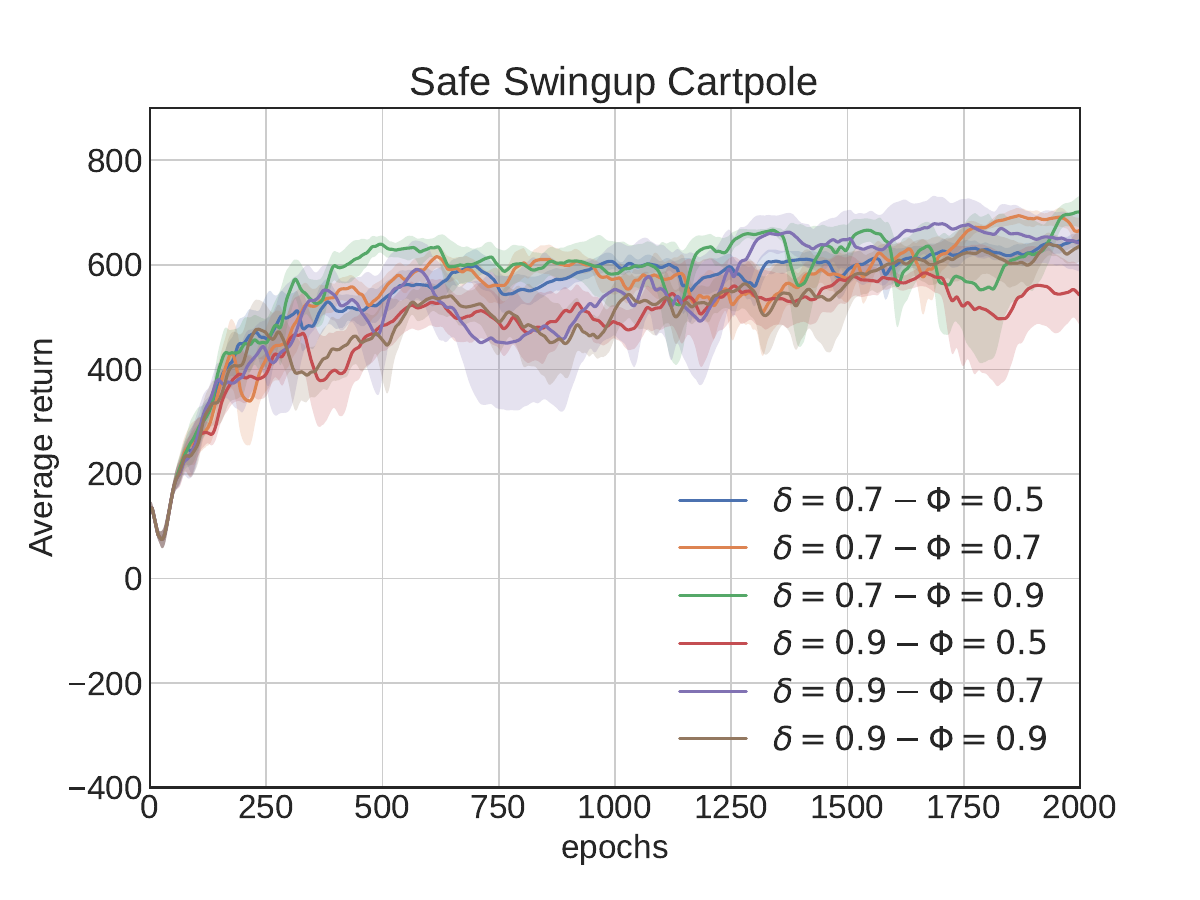}
  \label{fig:safe_cartpole_swingup_improved_switched}
\end{subfigure}%
\vspace{-0.3\baselineskip}
\begin{subfigure}{.28\textwidth}
  \centering
  \includegraphics[width=.99\linewidth]{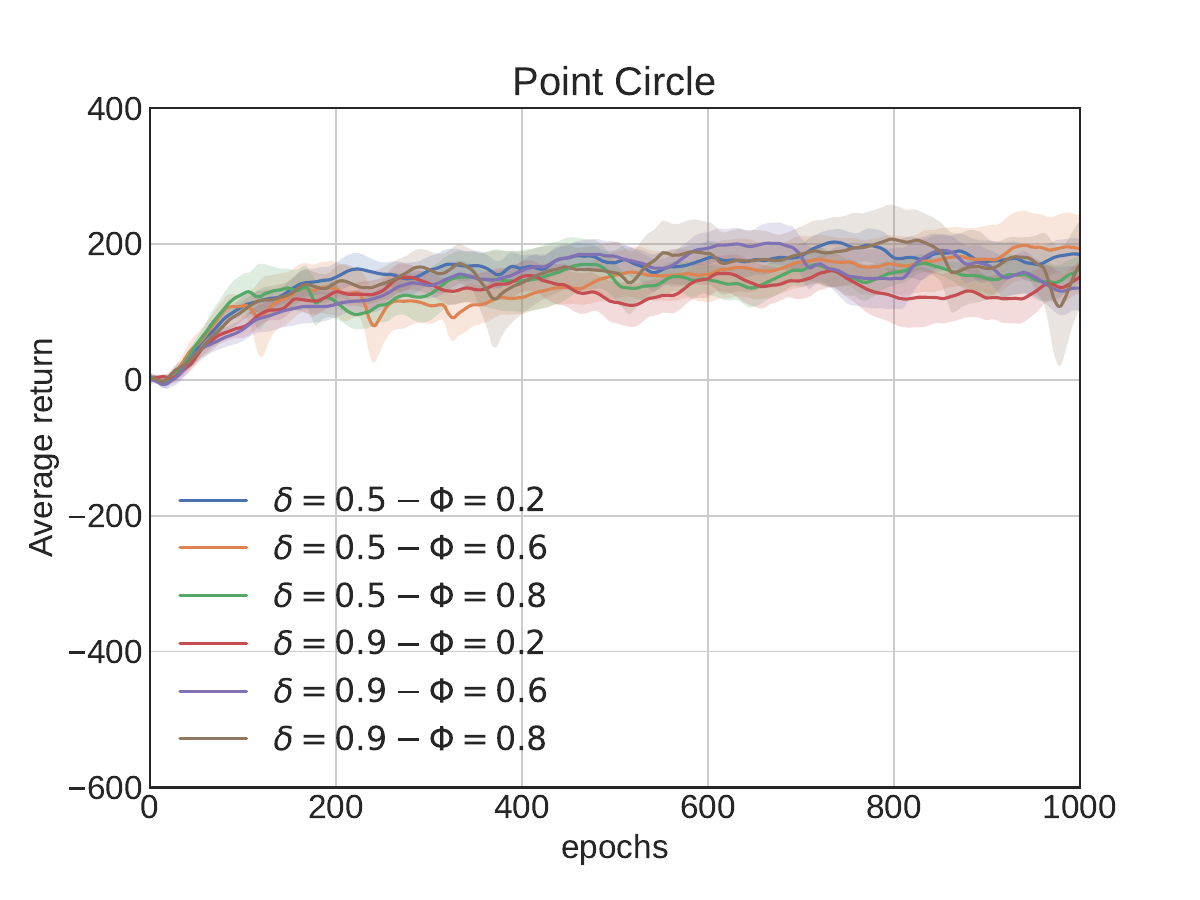}
  \label{fig:point_circle_ablation_improved_switched}
\end{subfigure}%
\vspace{-0.3\baselineskip}
\begin{subfigure}{.28\textwidth}
  \centering
  \includegraphics[width=.99\linewidth]{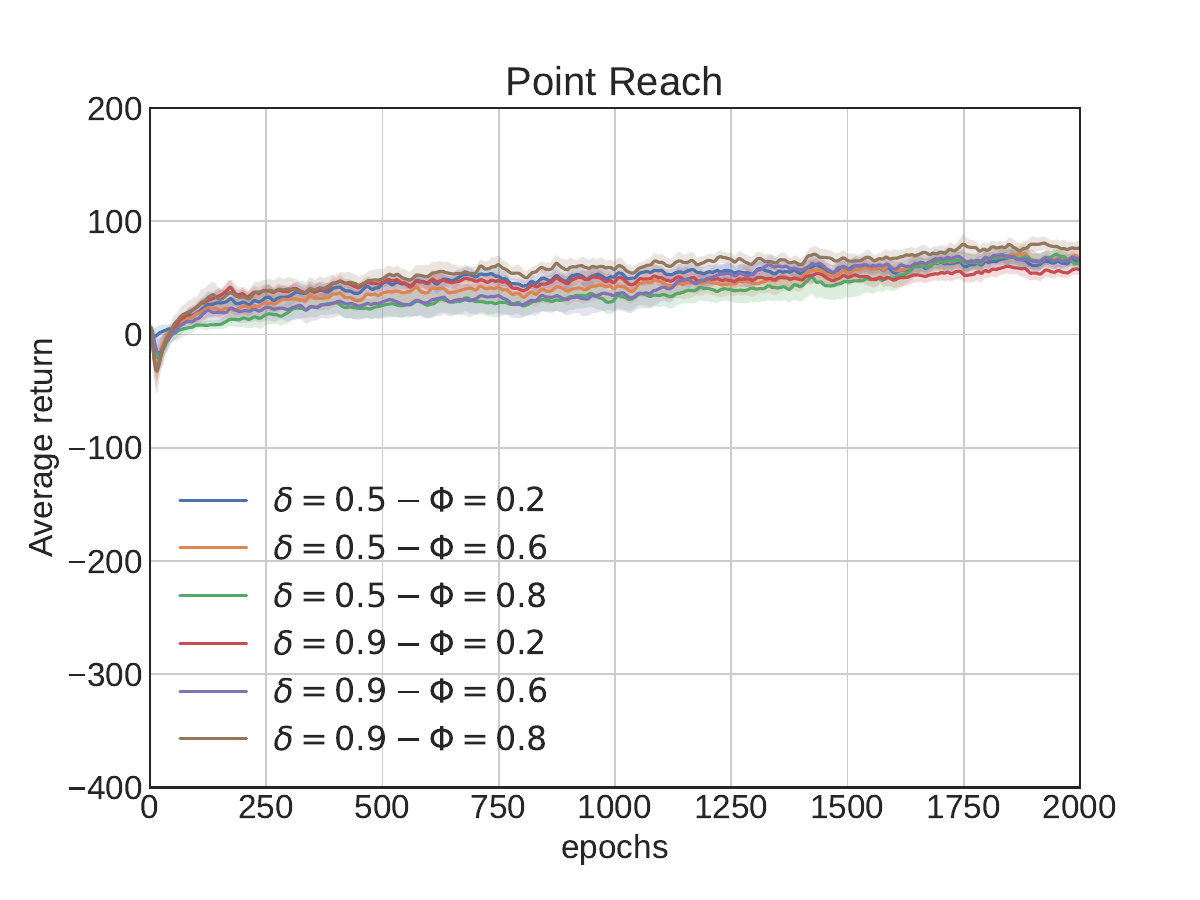}
  \label{fig:point_reach_ablation_improved_switched}
\end{subfigure}%
\vspace{-0.3\baselineskip}
\caption{Hyper-parameter analysis of PIG (top) and PAG (bottom) on environments with repulsive policies.}
  \label{fig:ablation_safe_agents}
\end{figure*}

\begin{figure*}
\vspace{-0.5\baselineskip}
\centering
\begin{subfigure}{.28\textwidth}
  \centering
  \includegraphics[width=.99\linewidth]{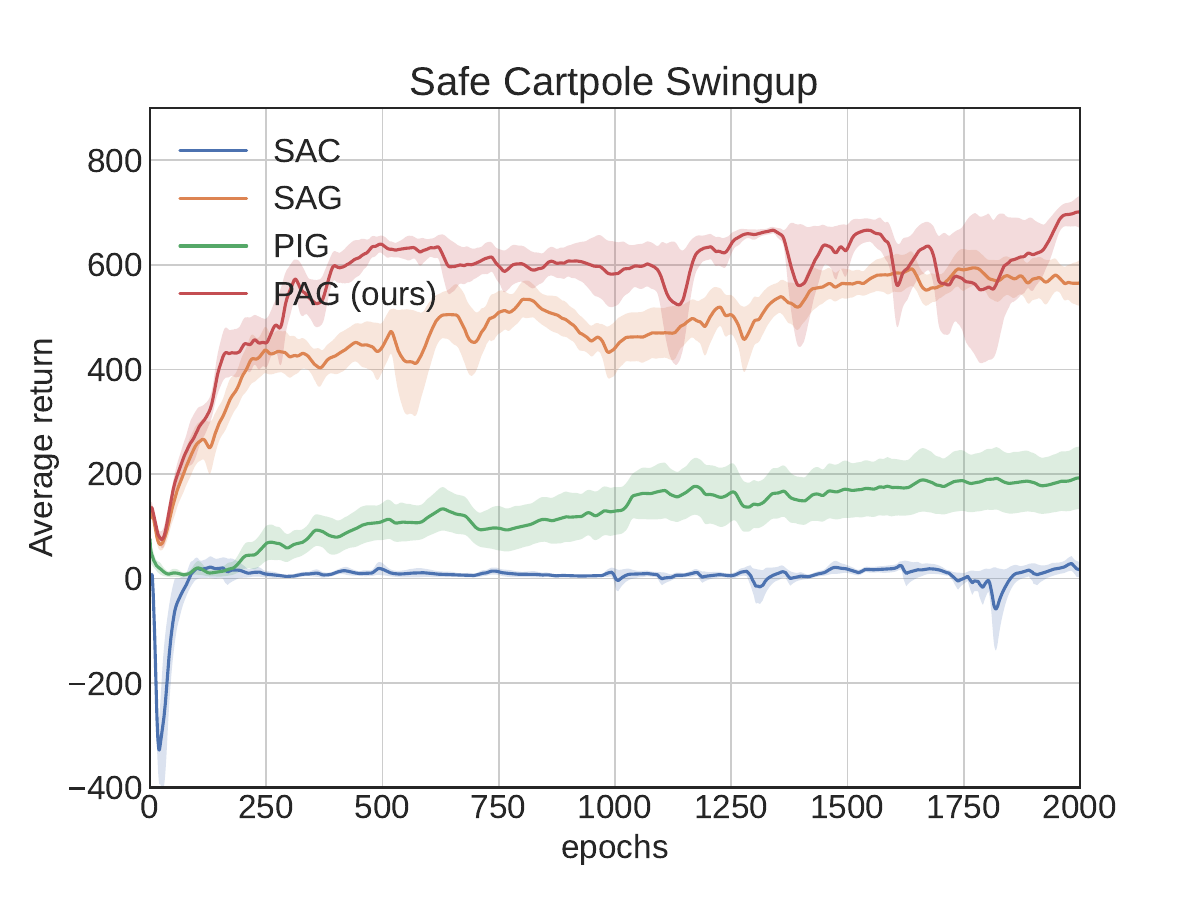}
  \label{fig:swingupcartpole_viz}
\end{subfigure}%
\vspace{-0.3\baselineskip}
\begin{subfigure}{.28\textwidth}
  \centering
  \includegraphics[width=.99\linewidth]{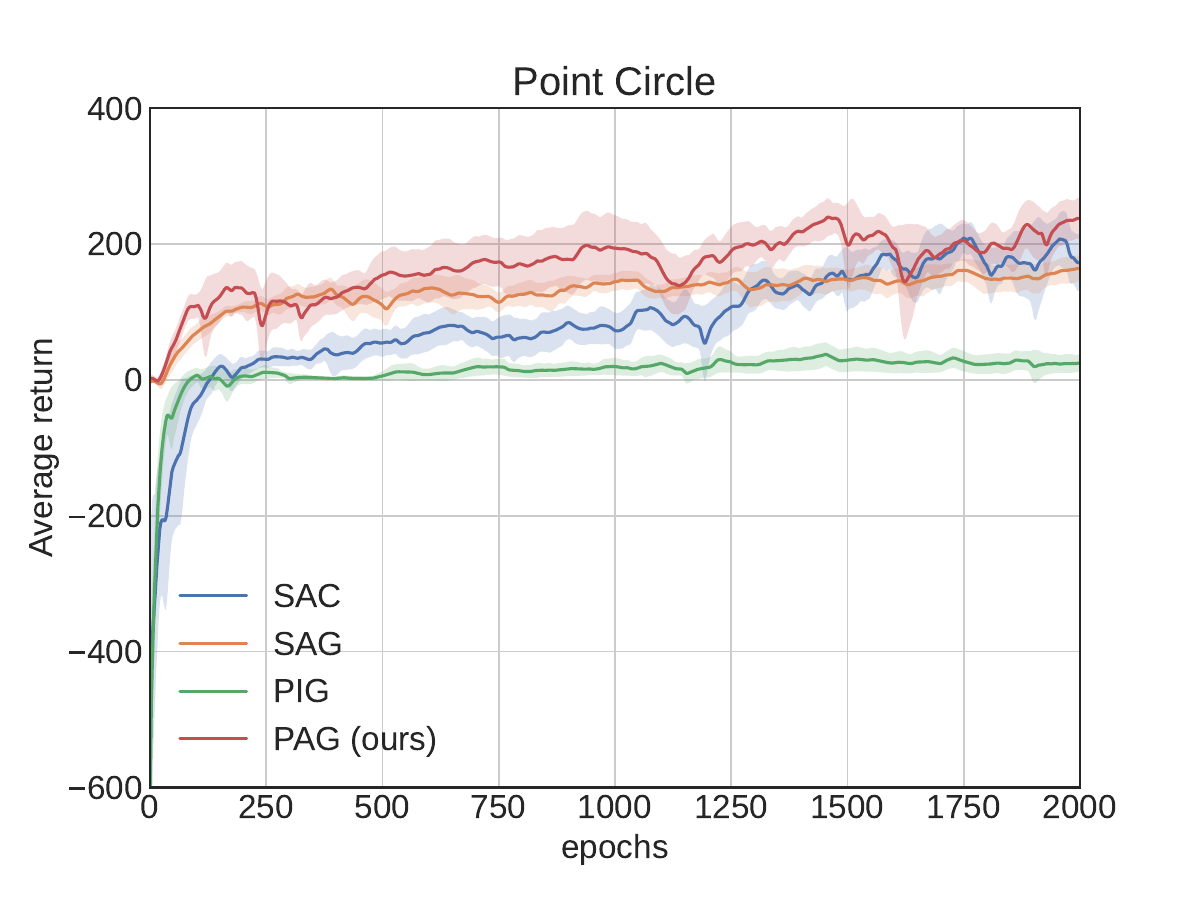}
  \label{fig:point_circle_viz}
\end{subfigure}
\vspace{-0.3\baselineskip}
\begin{subfigure}{.28\textwidth}
  \centering
  \includegraphics[width=.99\linewidth]{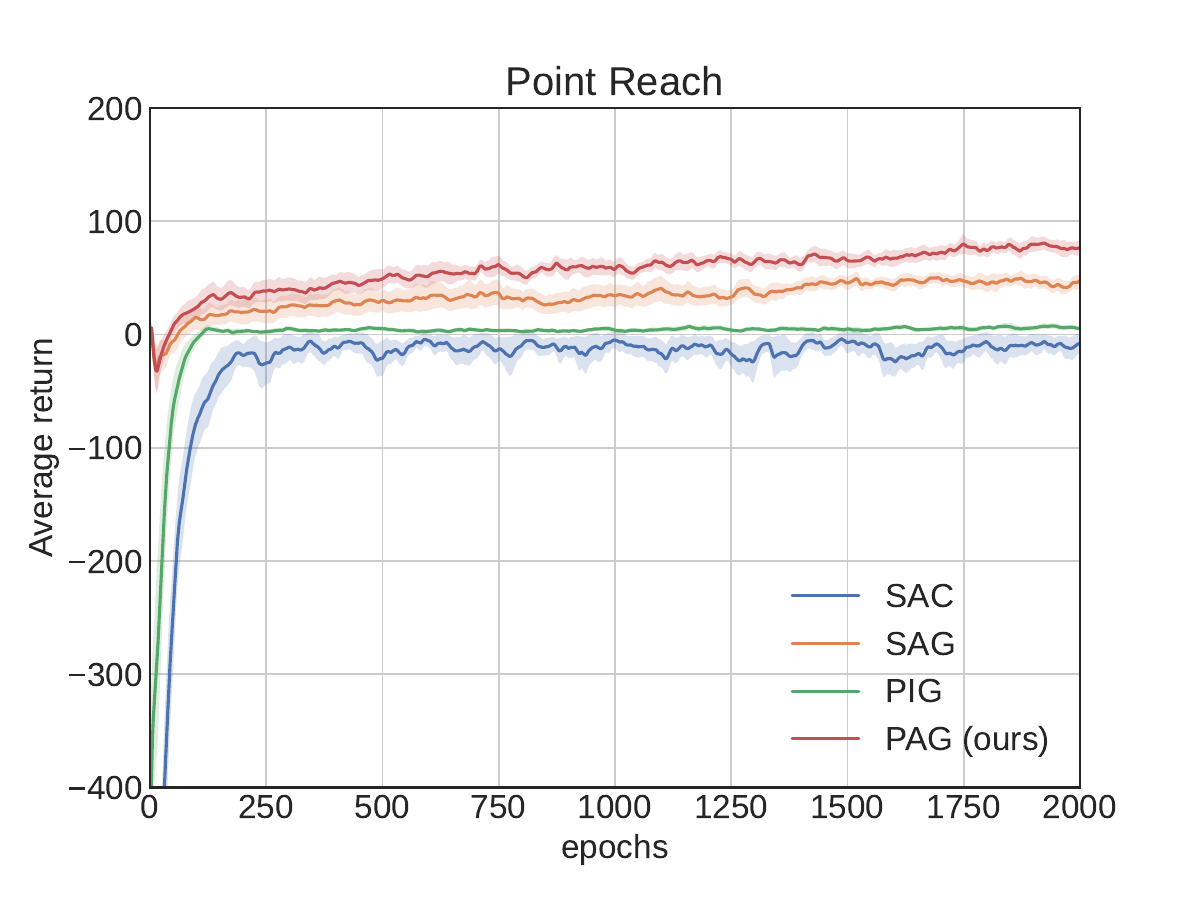}
  \label{fig:point_reach_viz}
\end{subfigure}
\vspace{-0.3\baselineskip}
\caption{Overall performances comparing PAG with SAC, SAG and PIG on 3 different environments with repulsive policies.}
  \label{fig:global_results_safe}
\end{figure*}

\paragraph{Metric} Along with the final performance of the agent and the respect of the safety constraints, we attentively pay attention to its initial performance. Thus, we compare the different agents using the normalized \textit{Area-Under-the-Curve} (AUC) as in \citep{teh2017distral, hessel2018rainbow} of the averaged cumulative performance of the agent during the total number of epochs. The AUC is normalized by the AUC of the perfect agent, that is the agent constantly having the optimal cumulative reward, chosen in this work as the best reward observed by the best agent. With an agent $A$, $CR(k)$ the cumulative return of the agent at epoch $k$ and $CR^*$ being the best $CR$ of the best agent, it is formalized as:

\begin{equation}
    AUC(A) = \frac{ \int_{k=0}^{K-1} CR(k) \,dk }{ K *  CR^* }.
\end{equation}

Those results are summarized in Table~\ref{tab:classic_control_results}.

We perform the analysis of the different agents on different environments from MuJoCo \citep{todorov2012mujoco} in the Deepmind Control Suite \citep{tassa2018deepmind} and PyBullet \citep{benelot2018} in Bullet-Safety-Gym \citep{gronauer2022bullet}. Additional details regarding the experimental protocol can be found in Appendix~\ref{app:details_env}.

\subsection{Local exploration with attractive policies}

We first consider local controllers to guide the agent on some parts of the environment, so the agent can quickly access relevant information and focus on the most difficult parts of the environment using standard RL. The objective is to learn a good policy with as few interactions as possible. 

In this setting, safety is not at stake, there is no need to gradually increase the impact of the parameterized perturbation $\xi^k_\phi$. Hence, the scheduler $\{\beta^k_{\text{PAG}}\}_{k=0}^K$ is set to $1$ in all experiments and, the scheduler $\{\beta^k_{\text{PIG}}\}_{k=0}^K$ starts at $1$ and is gradually decreased with a multiplicative factor $\delta^\kappa$, with $\kappa$ an integer starting at $0$ and increased by $1$ every $50$ epochs. The added heuristic has little to no impact at the end of learning (i.e $\kappa=20$) for the PIG agent.

\subsubsection{Environments}

In this part, we focused on \textit{Ball in Cup}, \textit{Point-Mass}, and \textit{Point-Maze}. The local guide is always a SAC agent stopped during mid-training. A complete description of these environments and guides is available in Appendix~\ref{app:details_env}, and can be visualized in Figure~\ref{fig:env_visualizations}. 

\subsubsection{Instability of standard approaches} \label{section:robustness}

First, we perform an extensive analysis of the current state-of-the-art PIG agent on different environments. This empirical study highlights the difficulty of applying the PIG agent in a real-world setting. Figure~\ref{fig:ablation_guided_agents} clearly emphasizes that PIG can perform well with a proper scheduler $\{\beta^k_{\text{PIG}}\}_{k=0}^K$. It is able to quickly find a near-optimal policy with a limited amount of samples. However, this method depends heavily on the choice of $\{\beta^k_{\text{PIG}}\}_{k=0}^K$: it requires the right initial $\beta^0_{\text{PIG}}$ with the right decay rate $\delta$. This phenomenon is notably visible on \textit{Point-Mass} in Figure~\ref{fig:ablation_guided_agents} where the performance of the PIG agent is unstable. In fact, the guidance is even detrimental with some schedulers on these environments and might prevent the agent from finding a near-optimal policy. To the best of our knowledge, there currently exists no way of knowing an appropriate scheduler in advance. This motivated our work to introduce a more stable and efficient algorithm.

\subsubsection{Robustness of our PAG agent to the hyper-parameter choices}

As opposed to existing approaches, our proposed algorithm PAG is more robust on the choice of its hyper-parameter $\Phi$. On all environments, Figure~\ref{fig:ablation_guided_agents} attests that the performance of the agent scarcely depends on the choice of $\Phi$. Even setting $\Phi = 1.5$, where the role of the local expert is reduced as the resulting controller would cover all of the action space, is useful to guide the agent in its first learning stages. The worst case scenario is to have the same performance as the standard RL agent, notably seen on \textit{Point-Mass}. Lower choices of $\Phi$ all lead to better results as they fully take advantage of the expertise provided by the local policy.

\subsubsection{Comparison of the different agents}  We include the standard RL agent (SAC) which does not use the local information and compare the best versions of the different guided agents in Figure~\ref{fig:global_results_safe}. In the proposed environments, all guided agents profit from a better initialization that the unguided one. However, SAG is not able to find a near-optimal policy because of its direct dependency on the local expert. Both the other guided agents - PIG and PAG - find a near-optimal policy much faster thanks to a clever integration of the local guide. In addition, our agent PAG turns out to be the most efficient approach in all the environments.

\subsection{Safe exploration with repulsive policies}

Another crucial application including a local policy in the process is learning without any safety violations, that is learning without exploring states that may have catastrophic consequences. While many successful approaches using the CMDP framework have been proposed \citep{garcia2015comprehensive, achiam2017constrained, chow2021safe}, we discard them as they need to see the constraint signal to build a safe policy. In a setting where this signal only provides information when these constraints are violated, it should never be seen. Hence, we investigate if it is possible to never see any constraint violation during learning with the help of an overly conservative emergency procedure that prevents the agent from entering such states. 

In this case, the agent should always stay close to the emergency procedure, so we set the scheduler $\{\beta^k_{\text{PIG}}\}_{k=0}^K$ to be fixed over learning for the PIG agent. Regarding our PAG agent, we set $\beta^0_{\text{PAG}}$ at $0$, and gradually increase it to $1$ following $(1 - \delta^\kappa)$, with $\kappa$ an integer starting at $0$ and increased by $1$ every $50$ epochs.

\subsubsection{Environments} We focused on \textit{Safe Cartpole Swingup}, \textit{Point-Circle} and \textit{Point-Reach}, and the local guides are scripted emergency procedures. Similarly, the description of the environments can be found in Appendix~\ref{app:details_env} and visualized in Figure~\ref{fig:env_visualizations}.

\subsubsection{Inefficiency of existing approaches}

Similar to the previous section, we investigate how the state-of-the-art method PIG performs in this setting. We observe in Figure~\ref{fig:ablation_safe_agents} that it is difficult to force the RL agent to stay within the safe zone only with a penalty on the \ref{eq:policy_improvement} step. In fact, most of the chosen hyper-parameters $\beta$ in most environments lead to safety violations at the beginning of learning in most environments. All the more surprising is that the choice of a relevant $\beta$ is complicated. For instance, in the \textit{Safe Cartpole Swingup} environment, the only agent that respects the constraints is when $\beta_{PIG} = 0.5$. Higher choices of $\beta_{\text{PIG}}$ ($\beta_{\text{PIG}} \in [1, 2, 5]$) visit catastrophic regions of the environment. Once again, this analysis empirically underlines the necessity of introducing a new method.

\subsubsection{Robustness of our PAG agent}

Our proposed PAG agent possesses an important hyper-parameter that allows a strict closeness with the emergency procedure if needed: $\Phi$. Hence, with a properly chosen $\Phi$, the agent never enters dangerous parts of the environment but is still able to overcome the conservativeness of the emergency procedure if needed. In addition, Figure~\ref{fig:ablation_safe_agents} attests that the performance PAG is robust to the choice of $\Phi$, as long as this one does not exceed a certain threshold.

\subsubsection{Comparison of the different agents}

In \textit{Safe Cartpole Swingup} and \textit{Point-Reach} environments, the classical RL agent SAC only learns not to go into catastrophic states, it does not solve the task. In \textit{Point-Circle}, it is able to do so but at the cost of visiting catastrophic states. The PIG agent violates the constraints less often than the unguided agent, but still consistently breaks them at the beginning of learning. In the \textit{Safe Cartpole Swingup} environment, it may be able to avoid breaking these constraints, but it is rather out of luck than from a proper choice of hyper-parameter. On the other hand, the agents that utilize the local policy at a higher level, \textit{e.g.} SAG and PAG agents, are successful at keeping the agent safe throughout learning. In addition, our PAG agent is also able to improve on the SAG agent that suffers from an overly conservative emergency procedure and find a better policy than the other methods.


\begin{table}
\caption{Normalized-AUC on the different environments.} \vskip -0.10in
\label{tab:classic_control_results} 
\begin{center}
\begin{small}
\begin{sc}
\begin{tabular}{l||cccc}
Environment & SAC & SAG & PIG & PAG (ours) \\[0.1cm] 
\hline
\texttt{ball-in-cup}  & $0.73 \pm 0.06$ & $0.74 \pm 0.08$ &  $0.9 \pm 0.01$ & $\bm{0.95 \pm 0.01}$ \\[0.1cm] 
   \hline
\texttt{point-mass} & $\bm{0.72 \pm 0.06}$ & $0.63 \pm 0.03$ & $0.71 \pm 0.1$ & $\bm{0.72 \pm 0.02}$ \\[0.1cm] 
 \hline
\texttt{point-fall} & $0.34 \pm 0.27$ & $0.38 \pm 0.01$ & $0.70 \pm 0.08$ & $\bm{0.85 \pm 0.06}$  \\[0.1cm] 
\hline
\texttt{cartpole} & $0.01 \pm 0.01$ & $0.69 \pm 0.04$ & $0.18 \pm 0.09$ & $\bm{0.82 \pm 0.06}$ \\[0.1cm] 
\hline
\texttt{point-circle} & $0.39 \pm 0.13$ & $0.55 \pm 0.1$ & $0.06 \pm 0.06$ & $\bm{0.71 \pm 0.19}$ \\[0.1cm] 
\hline
\texttt{point-reach} & $-0.29 \pm 0.1$ & $0.43 \pm 0.12$ & $0.01 \pm 0.01$ & $\bm{0.71 \pm 0.09}$ \\[0.1cm] 
\hline
\texttt{Total Mean} & $0.32$ & $0.57$ & $0.43$ & $\bm{0.8}$  \\[0.1cm] 
\end{tabular}
\end{sc}
\end{small}
\end{center}
\vskip -0.22in
\end{table}

\section{Conclusion}

We formalized a novel setting that generalizes access to any kind of controller that may be available in real-world systems. We especially study the introduction of local controllers to enhance any Approximate Policy Iteration-based RL agent thanks to a novel algorithm. We studied two important use cases: guiding the agent towards meaningful regions of the environment and preventing it from entering dangerous state spaces. Our method was stable, robust, and more efficient than standard approaches applied in this setting. In addition, we allow a good initialization and a strict constraint on the closeness with respect to the local controller that may be important on safety critical systems to learn a good policy with zero safety violations.
However, in all the tested environments, we relied on a known confidence function that details the relevance of the local controller in simple environments. This may be a strong assumption in more complex systems that would require a finer analysis, which will be the focus of our future works.






\bibliographystyle{ACM-Reference-Format} 
\balance
\bibliography{aamas}


\begin{thebibliography}{93}


\ifx \showCODEN    \undefined \def \showCODEN     #1{\unskip}     \fi
\ifx \showDOI      \undefined \def \showDOI       #1{#1}\fi
\ifx \showISBNx    \undefined \def \showISBNx     #1{\unskip}     \fi
\ifx \showISBNxiii \undefined \def \showISBNxiii  #1{\unskip}     \fi
\ifx \showISSN     \undefined \def \showISSN      #1{\unskip}     \fi
\ifx \showLCCN     \undefined \def \showLCCN      #1{\unskip}     \fi
\ifx \shownote     \undefined \def \shownote      #1{#1}          \fi
\ifx \showarticletitle \undefined \def \showarticletitle #1{#1}   \fi
\ifx \showURL      \undefined \def \showURL       {\relax}        \fi
\providecommand\bibfield[2]{#2}
\providecommand\bibinfo[2]{#2}
\providecommand\natexlab[1]{#1}
\providecommand\showeprint[2][]{arXiv:#2}

\bibitem[\protect\citeauthoryear{Abbeel and Ng}{Abbeel and Ng}{2004}]%
        {abbeel2004apprenticeship}
\bibfield{author}{\bibinfo{person}{Pieter Abbeel} {and} \bibinfo{person}{Andrew~Y Ng}.} \bibinfo{year}{2004}\natexlab{}.
\newblock \showarticletitle{Apprenticeship learning via inverse reinforcement learning}. In \bibinfo{booktitle}{\emph{International conference on machine learning}}. \bibinfo{pages}{1}.
\newblock


\bibitem[\protect\citeauthoryear{Achiam, Held, Tamar, and Abbeel}{Achiam et~al\mbox{.}}{2017}]%
        {achiam2017constrained}
\bibfield{author}{\bibinfo{person}{Joshua Achiam}, \bibinfo{person}{David Held}, \bibinfo{person}{Aviv Tamar}, {and} \bibinfo{person}{Pieter Abbeel}.} \bibinfo{year}{2017}\natexlab{}.
\newblock \showarticletitle{Constrained policy optimization}. In \bibinfo{booktitle}{\emph{International conference on machine learning}}. PMLR, \bibinfo{pages}{22--31}.
\newblock


\bibitem[\protect\citeauthoryear{Agarwal, Schwarzer, Castro, Courville, and Bellemare}{Agarwal et~al\mbox{.}}{2022}]%
        {agarwal2022beyond}
\bibfield{author}{\bibinfo{person}{Rishabh Agarwal}, \bibinfo{person}{Max Schwarzer}, \bibinfo{person}{Pablo~Samuel Castro}, \bibinfo{person}{Aaron Courville}, {and} \bibinfo{person}{Marc~G Bellemare}.} \bibinfo{year}{2022}\natexlab{}.
\newblock \showarticletitle{Beyond Tabula Rasa: Reincarnating Reinforcement Learning}.
\newblock \bibinfo{journal}{\emph{arXiv preprint arXiv:2206.01626}} (\bibinfo{year}{2022}).
\newblock


\bibitem[\protect\citeauthoryear{Alcala, Puig, Quevedo, and Escobet}{Alcala et~al\mbox{.}}{2018}]%
        {alcala2018gain}
\bibfield{author}{\bibinfo{person}{Eugenio Alcala}, \bibinfo{person}{Vicen{\c{c}} Puig}, \bibinfo{person}{Joseba Quevedo}, {and} \bibinfo{person}{Teresa Escobet}.} \bibinfo{year}{2018}\natexlab{}.
\newblock \showarticletitle{Gain-scheduling LPV control for autonomous vehicles including friction force estimation and compensation mechanism}.
\newblock \bibinfo{journal}{\emph{IET Control Theory \& Applications}} \bibinfo{volume}{12}, \bibinfo{number}{12} (\bibinfo{year}{2018}), \bibinfo{pages}{1683--1693}.
\newblock


\bibitem[\protect\citeauthoryear{Arora and Doshi}{Arora and Doshi}{2021}]%
        {arora2021survey}
\bibfield{author}{\bibinfo{person}{Saurabh Arora} {and} \bibinfo{person}{Prashant Doshi}.} \bibinfo{year}{2021}\natexlab{}.
\newblock \showarticletitle{A survey of inverse reinforcement learning: Challenges, methods and progress}.
\newblock \bibinfo{journal}{\emph{Artificial Intelligence}}  \bibinfo{volume}{297} (\bibinfo{year}{2021}), \bibinfo{pages}{103500}.
\newblock


\bibitem[\protect\citeauthoryear{Aytar, Pfaff, Budden, Paine, Wang, and De~Freitas}{Aytar et~al\mbox{.}}{2018}]%
        {aytar2018playing}
\bibfield{author}{\bibinfo{person}{Yusuf Aytar}, \bibinfo{person}{Tobias Pfaff}, \bibinfo{person}{David Budden}, \bibinfo{person}{Thomas Paine}, \bibinfo{person}{Ziyu Wang}, {and} \bibinfo{person}{Nando De~Freitas}.} \bibinfo{year}{2018}\natexlab{}.
\newblock \showarticletitle{Playing hard exploration games by watching youtube}.
\newblock \bibinfo{journal}{\emph{Advances in neural information processing systems}}  \bibinfo{volume}{31} (\bibinfo{year}{2018}).
\newblock


\bibitem[\protect\citeauthoryear{Bellemare, Srinivasan, Ostrovski, Schaul, Saxton, and Munos}{Bellemare et~al\mbox{.}}{2016}]%
        {bellemare2016unifying}
\bibfield{author}{\bibinfo{person}{Marc Bellemare}, \bibinfo{person}{Sriram Srinivasan}, \bibinfo{person}{Georg Ostrovski}, \bibinfo{person}{Tom Schaul}, \bibinfo{person}{David Saxton}, {and} \bibinfo{person}{Remi Munos}.} \bibinfo{year}{2016}\natexlab{}.
\newblock \showarticletitle{Unifying count-based exploration and intrinsic motivation}.
\newblock \bibinfo{journal}{\emph{Advances in neural information processing systems}}  \bibinfo{volume}{29} (\bibinfo{year}{2016}).
\newblock


\bibitem[\protect\citeauthoryear{Bellemare, Candido, Castro, Gong, Machado, Moitra, Ponda, and Wang}{Bellemare et~al\mbox{.}}{2020}]%
        {bellemare2020autonomous}
\bibfield{author}{\bibinfo{person}{Marc~G Bellemare}, \bibinfo{person}{Salvatore Candido}, \bibinfo{person}{Pablo~Samuel Castro}, \bibinfo{person}{Jun Gong}, \bibinfo{person}{Marlos~C Machado}, \bibinfo{person}{Subhodeep Moitra}, \bibinfo{person}{Sameera~S Ponda}, {and} \bibinfo{person}{Ziyu Wang}.} \bibinfo{year}{2020}\natexlab{}.
\newblock \showarticletitle{Autonomous navigation of stratospheric balloons using reinforcement learning}.
\newblock \bibinfo{journal}{\emph{Nature}} \bibinfo{volume}{588}, \bibinfo{number}{7836} (\bibinfo{year}{2020}), \bibinfo{pages}{77--82}.
\newblock


\bibitem[\protect\citeauthoryear{Bellemare, Naddaf, Veness, and Bowling}{Bellemare et~al\mbox{.}}{2013}]%
        {bellemare2013arcade}
\bibfield{author}{\bibinfo{person}{Marc~G Bellemare}, \bibinfo{person}{Yavar Naddaf}, \bibinfo{person}{Joel Veness}, {and} \bibinfo{person}{Michael Bowling}.} \bibinfo{year}{2013}\natexlab{}.
\newblock \showarticletitle{The arcade learning environment: An evaluation platform for general agents}.
\newblock \bibinfo{journal}{\emph{Journal of Artificial Intelligence Research}}  \bibinfo{volume}{47} (\bibinfo{year}{2013}), \bibinfo{pages}{253--279}.
\newblock


\bibitem[\protect\citeauthoryear{Byun and Perrault}{Byun and Perrault}{2021}]%
        {byun2021training}
\bibfield{author}{\bibinfo{person}{Ju-Seung Byun} {and} \bibinfo{person}{Andrew Perrault}.} \bibinfo{year}{2021}\natexlab{}.
\newblock \showarticletitle{Training Transition Policies via Distribution Matching for Complex Tasks}.
\newblock \bibinfo{journal}{\emph{arXiv preprint arXiv:2110.04357}} (\bibinfo{year}{2021}).
\newblock


\bibitem[\protect\citeauthoryear{Cai, Ding, Jiang, and Zhou}{Cai et~al\mbox{.}}{2019}]%
        {cai2019imitation}
\bibfield{author}{\bibinfo{person}{Xin-Qiang Cai}, \bibinfo{person}{Yao-Xiang Ding}, \bibinfo{person}{Yuan Jiang}, {and} \bibinfo{person}{Zhi-Hua Zhou}.} \bibinfo{year}{2019}\natexlab{}.
\newblock \showarticletitle{Imitation learning from pixel-level demonstrations by hashreward}.
\newblock \bibinfo{journal}{\emph{arXiv preprint arXiv:1909.03773}} (\bibinfo{year}{2019}).
\newblock


\bibitem[\protect\citeauthoryear{Cheng, Kolobov, and Swaminathan}{Cheng et~al\mbox{.}}{2021}]%
        {cheng2021heuristic}
\bibfield{author}{\bibinfo{person}{Ching-An Cheng}, \bibinfo{person}{Andrey Kolobov}, {and} \bibinfo{person}{Adith Swaminathan}.} \bibinfo{year}{2021}\natexlab{}.
\newblock \showarticletitle{Heuristic-guided reinforcement learning}.
\newblock \bibinfo{journal}{\emph{Advances in Neural Information Processing Systems}}  \bibinfo{volume}{34} (\bibinfo{year}{2021}), \bibinfo{pages}{13550--13563}.
\newblock


\bibitem[\protect\citeauthoryear{Chow, Nachum, Faust, Due{\~n}ez-Guzman, and Ghavamzadeh}{Chow et~al\mbox{.}}{2021}]%
        {chow2021safe}
\bibfield{author}{\bibinfo{person}{Yinlam Chow}, \bibinfo{person}{Ofir Nachum}, \bibinfo{person}{Aleksandra Faust}, \bibinfo{person}{Edgar Due{\~n}ez-Guzman}, {and} \bibinfo{person}{Mohammad Ghavamzadeh}.} \bibinfo{year}{2021}\natexlab{}.
\newblock \showarticletitle{Safe Policy Learning for Continuous Control}. In \bibinfo{booktitle}{\emph{Conference on Robot Learning}}. PMLR, \bibinfo{pages}{801--821}.
\newblock


\bibitem[\protect\citeauthoryear{Codevilla, Santana, L{\'{o}}pez, and Gaidon}{Codevilla et~al\mbox{.}}{2019}]%
        {CodevillaSLG19}
\bibfield{author}{\bibinfo{person}{Felipe Codevilla}, \bibinfo{person}{Eder Santana}, \bibinfo{person}{Antonio~M. L{\'{o}}pez}, {and} \bibinfo{person}{Adrien Gaidon}.} \bibinfo{year}{2019}\natexlab{}.
\newblock \showarticletitle{Exploring the Limitations of Behavior Cloning for Autonomous Driving}. In \bibinfo{booktitle}{\emph{{ICCV}}}. \bibinfo{publisher}{{IEEE}}, \bibinfo{pages}{9328--9337}.
\newblock


\bibitem[\protect\citeauthoryear{Coutinho and Palhares}{Coutinho and Palhares}{2021}]%
        {coutinho2021dynamic}
\bibfield{author}{\bibinfo{person}{Pedro Henrique~Silva Coutinho} {and} \bibinfo{person}{Reinaldo~Mart{\'\i}nez Palhares}.} \bibinfo{year}{2021}\natexlab{}.
\newblock \showarticletitle{Dynamic periodic event-triggered gain-scheduling control co-design for quasi-LPV systems}.
\newblock \bibinfo{journal}{\emph{Nonlinear Analysis: Hybrid Systems}}  \bibinfo{volume}{41} (\bibinfo{year}{2021}), \bibinfo{pages}{101044}.
\newblock


\bibitem[\protect\citeauthoryear{Degrave, Felici, Buchli, Neunert, Tracey, Carpanese, Ewalds, Hafner, Abdolmaleki, de~Las~Casas, et~al\mbox{.}}{Degrave et~al\mbox{.}}{2022}]%
        {degrave2022magnetic}
\bibfield{author}{\bibinfo{person}{Jonas Degrave}, \bibinfo{person}{Federico Felici}, \bibinfo{person}{Jonas Buchli}, \bibinfo{person}{Michael Neunert}, \bibinfo{person}{Brendan Tracey}, \bibinfo{person}{Francesco Carpanese}, \bibinfo{person}{Timo Ewalds}, \bibinfo{person}{Roland Hafner}, \bibinfo{person}{Abbas Abdolmaleki}, \bibinfo{person}{Diego de Las~Casas}, {et~al\mbox{.}}} \bibinfo{year}{2022}\natexlab{}.
\newblock \showarticletitle{Magnetic control of tokamak plasmas through deep reinforcement learning}.
\newblock \bibinfo{journal}{\emph{Nature}} \bibinfo{volume}{602}, \bibinfo{number}{7897} (\bibinfo{year}{2022}), \bibinfo{pages}{414--419}.
\newblock


\bibitem[\protect\citeauthoryear{Dorigo and Colombetti}{Dorigo and Colombetti}{1994}]%
        {dorigo1994robot}
\bibfield{author}{\bibinfo{person}{Marco Dorigo} {and} \bibinfo{person}{Marco Colombetti}.} \bibinfo{year}{1994}\natexlab{}.
\newblock \showarticletitle{Robot shaping: Developing autonomous agents through learning}.
\newblock \bibinfo{journal}{\emph{Artificial intelligence}} \bibinfo{volume}{71}, \bibinfo{number}{2} (\bibinfo{year}{1994}), \bibinfo{pages}{321--370}.
\newblock


\bibitem[\protect\citeauthoryear{Dulac-Arnold, Levine, Mankowitz, Li, Paduraru, Gowal, and Hester}{Dulac-Arnold et~al\mbox{.}}{2021}]%
        {dulac2021challenges}
\bibfield{author}{\bibinfo{person}{Gabriel Dulac-Arnold}, \bibinfo{person}{Nir Levine}, \bibinfo{person}{Daniel~J Mankowitz}, \bibinfo{person}{Jerry Li}, \bibinfo{person}{Cosmin Paduraru}, \bibinfo{person}{Sven Gowal}, {and} \bibinfo{person}{Todd Hester}.} \bibinfo{year}{2021}\natexlab{}.
\newblock \showarticletitle{Challenges of real-world reinforcement learning: definitions, benchmarks and analysis}.
\newblock \bibinfo{journal}{\emph{Machine Learning}} \bibinfo{volume}{110}, \bibinfo{number}{9} (\bibinfo{year}{2021}), \bibinfo{pages}{2419--2468}.
\newblock


\bibitem[\protect\citeauthoryear{Ellenberger}{Ellenberger}{2019}]%
        {benelot2018}
\bibfield{author}{\bibinfo{person}{Benjamin Ellenberger}.} \bibinfo{year}{2018--2019}\natexlab{}.
\newblock \bibinfo{title}{PyBullet Gymperium}.
\newblock \bibinfo{howpublished}{\url{https://github.com/benelot/pybullet-gym}}.
\newblock


\bibitem[\protect\citeauthoryear{Fuchs, Song, Kaufmann, Scaramuzza, and D{\"u}rr}{Fuchs et~al\mbox{.}}{2021}]%
        {fuchs2021super}
\bibfield{author}{\bibinfo{person}{Florian Fuchs}, \bibinfo{person}{Yunlong Song}, \bibinfo{person}{Elia Kaufmann}, \bibinfo{person}{Davide Scaramuzza}, {and} \bibinfo{person}{Peter D{\"u}rr}.} \bibinfo{year}{2021}\natexlab{}.
\newblock \showarticletitle{Super-human performance in gran turismo sport using deep reinforcement learning}.
\newblock \bibinfo{journal}{\emph{IEEE Robotics and Automation Letters}} \bibinfo{volume}{6}, \bibinfo{number}{3} (\bibinfo{year}{2021}), \bibinfo{pages}{4257--4264}.
\newblock


\bibitem[\protect\citeauthoryear{Fujimoto, Hoof, and Meger}{Fujimoto et~al\mbox{.}}{2018}]%
        {fujimoto2018addressing}
\bibfield{author}{\bibinfo{person}{Scott Fujimoto}, \bibinfo{person}{Herke Hoof}, {and} \bibinfo{person}{David Meger}.} \bibinfo{year}{2018}\natexlab{}.
\newblock \showarticletitle{Addressing function approximation error in actor-critic methods}. In \bibinfo{booktitle}{\emph{International conference on machine learning}}. PMLR, \bibinfo{pages}{1587--1596}.
\newblock


\bibitem[\protect\citeauthoryear{Fujimoto, Meger, and Precup}{Fujimoto et~al\mbox{.}}{2019}]%
        {fujimoto2019off}
\bibfield{author}{\bibinfo{person}{Scott Fujimoto}, \bibinfo{person}{David Meger}, {and} \bibinfo{person}{Doina Precup}.} \bibinfo{year}{2019}\natexlab{}.
\newblock \showarticletitle{Off-policy deep reinforcement learning without exploration}. In \bibinfo{booktitle}{\emph{International conference on machine learning}}. PMLR, \bibinfo{pages}{2052--2062}.
\newblock


\bibitem[\protect\citeauthoryear{Gallego, Merello, Berenguel, and Camacho}{Gallego et~al\mbox{.}}{2019}]%
        {gallego2019gain}
\bibfield{author}{\bibinfo{person}{Antonio~J Gallego}, \bibinfo{person}{Gonzalo~M Merello}, \bibinfo{person}{Manuel Berenguel}, {and} \bibinfo{person}{Eduardo~F Camacho}.} \bibinfo{year}{2019}\natexlab{}.
\newblock \showarticletitle{Gain-scheduling model predictive control of a Fresnel collector field}.
\newblock \bibinfo{journal}{\emph{Control Engineering Practice}}  \bibinfo{volume}{82} (\bibinfo{year}{2019}), \bibinfo{pages}{1--13}.
\newblock


\bibitem[\protect\citeauthoryear{Garc{\i}a and Fern{\'a}ndez}{Garc{\i}a and Fern{\'a}ndez}{2015}]%
        {garcia2015comprehensive}
\bibfield{author}{\bibinfo{person}{Javier Garc{\i}a} {and} \bibinfo{person}{Fernando Fern{\'a}ndez}.} \bibinfo{year}{2015}\natexlab{}.
\newblock \showarticletitle{A comprehensive survey on safe reinforcement learning}.
\newblock \bibinfo{journal}{\emph{Journal of Machine Learning Research}} \bibinfo{volume}{16}, \bibinfo{number}{1} (\bibinfo{year}{2015}), \bibinfo{pages}{1437--1480}.
\newblock


\bibitem[\protect\citeauthoryear{Geist, Scherrer, and Pietquin}{Geist et~al\mbox{.}}{2019}]%
        {geist2019theory}
\bibfield{author}{\bibinfo{person}{Matthieu Geist}, \bibinfo{person}{Bruno Scherrer}, {and} \bibinfo{person}{Olivier Pietquin}.} \bibinfo{year}{2019}\natexlab{}.
\newblock \showarticletitle{A theory of regularized markov decision processes}. In \bibinfo{booktitle}{\emph{International conference on machine learning}}. PMLR, \bibinfo{pages}{2160--2169}.
\newblock


\bibitem[\protect\citeauthoryear{Gillen, Molnar, and Byl}{Gillen et~al\mbox{.}}{2020}]%
        {gillen2020combining}
\bibfield{author}{\bibinfo{person}{Sean Gillen}, \bibinfo{person}{Marco Molnar}, {and} \bibinfo{person}{Katie Byl}.} \bibinfo{year}{2020}\natexlab{}.
\newblock \showarticletitle{Combining deep reinforcement learning and local control for the acrobot swing-up and balance task}. In \bibinfo{booktitle}{\emph{2020 59th IEEE Conference on Decision and Control (CDC)}}. IEEE, \bibinfo{pages}{4129--4134}.
\newblock


\bibitem[\protect\citeauthoryear{Gronauer}{Gronauer}{2022}]%
        {gronauer2022bullet}
\bibfield{author}{\bibinfo{person}{Sven Gronauer}.} \bibinfo{year}{2022}\natexlab{}.
\newblock \showarticletitle{Bullet-Safety-Gym: a framework for constrained Reinforcement Learning}.
\newblock  (\bibinfo{year}{2022}).
\newblock


\bibitem[\protect\citeauthoryear{Haarnoja, Zhou, Hartikainen, Tucker, Ha, Tan, Kumar, Zhu, Gupta, Abbeel, et~al\mbox{.}}{Haarnoja et~al\mbox{.}}{2018}]%
        {haarnoja2018soft}
\bibfield{author}{\bibinfo{person}{Tuomas Haarnoja}, \bibinfo{person}{Aurick Zhou}, \bibinfo{person}{Kristian Hartikainen}, \bibinfo{person}{George Tucker}, \bibinfo{person}{Sehoon Ha}, \bibinfo{person}{Jie Tan}, \bibinfo{person}{Vikash Kumar}, \bibinfo{person}{Henry Zhu}, \bibinfo{person}{Abhishek Gupta}, \bibinfo{person}{Pieter Abbeel}, {et~al\mbox{.}}} \bibinfo{year}{2018}\natexlab{}.
\newblock \showarticletitle{Soft actor-critic algorithms and applications}.
\newblock \bibinfo{journal}{\emph{arXiv preprint arXiv:1812.05905}} (\bibinfo{year}{2018}).
\newblock


\bibitem[\protect\citeauthoryear{Harutyunyan, Devlin, Vrancx, and Now{\'{e}}}{Harutyunyan et~al\mbox{.}}{2015}]%
        {HarutyunyanDVN15}
\bibfield{author}{\bibinfo{person}{Anna Harutyunyan}, \bibinfo{person}{Sam Devlin}, \bibinfo{person}{Peter Vrancx}, {and} \bibinfo{person}{Ann Now{\'{e}}}.} \bibinfo{year}{2015}\natexlab{}.
\newblock \showarticletitle{Expressing Arbitrary Reward Functions as Potential-Based Advice}. In \bibinfo{booktitle}{\emph{Proceedings of the AAAI Conference on Artificial Intelligence}}. \bibinfo{publisher}{{AAAI} Press}, \bibinfo{pages}{2652--2658}.
\newblock


\bibitem[\protect\citeauthoryear{Hessel, Modayil, Van~Hasselt, Schaul, Ostrovski, Dabney, Horgan, Piot, Azar, and Silver}{Hessel et~al\mbox{.}}{2018}]%
        {hessel2018rainbow}
\bibfield{author}{\bibinfo{person}{Matteo Hessel}, \bibinfo{person}{Joseph Modayil}, \bibinfo{person}{Hado Van~Hasselt}, \bibinfo{person}{Tom Schaul}, \bibinfo{person}{Georg Ostrovski}, \bibinfo{person}{Will Dabney}, \bibinfo{person}{Dan Horgan}, \bibinfo{person}{Bilal Piot}, \bibinfo{person}{Mohammad Azar}, {and} \bibinfo{person}{David Silver}.} \bibinfo{year}{2018}\natexlab{}.
\newblock \showarticletitle{Rainbow: Combining improvements in deep reinforcement learning}. In \bibinfo{booktitle}{\emph{Proceedings of the AAAI Conference on Artificial Intelligence}}.
\newblock


\bibitem[\protect\citeauthoryear{Hester, Vecerik, Pietquin, Lanctot, Schaul, Piot, Horgan, Quan, Sendonaris, Osband, et~al\mbox{.}}{Hester et~al\mbox{.}}{2018}]%
        {hester2018deep}
\bibfield{author}{\bibinfo{person}{Todd Hester}, \bibinfo{person}{Matej Vecerik}, \bibinfo{person}{Olivier Pietquin}, \bibinfo{person}{Marc Lanctot}, \bibinfo{person}{Tom Schaul}, \bibinfo{person}{Bilal Piot}, \bibinfo{person}{Dan Horgan}, \bibinfo{person}{John Quan}, \bibinfo{person}{Andrew Sendonaris}, \bibinfo{person}{Ian Osband}, {et~al\mbox{.}}} \bibinfo{year}{2018}\natexlab{}.
\newblock \showarticletitle{Deep q-learning from demonstrations}. In \bibinfo{booktitle}{\emph{Proceedings of the AAAI Conference on Artificial Intelligence}}, Vol.~\bibinfo{volume}{32}.
\newblock


\bibitem[\protect\citeauthoryear{Ho and Ermon}{Ho and Ermon}{2016}]%
        {ho2016generative}
\bibfield{author}{\bibinfo{person}{Jonathan Ho} {and} \bibinfo{person}{Stefano Ermon}.} \bibinfo{year}{2016}\natexlab{}.
\newblock \showarticletitle{Generative adversarial imitation learning}.
\newblock \bibinfo{journal}{\emph{Advances in neural information processing systems}}  \bibinfo{volume}{29} (\bibinfo{year}{2016}).
\newblock


\bibitem[\protect\citeauthoryear{Hussein, Gaber, Elyan, and Jayne}{Hussein et~al\mbox{.}}{2017}]%
        {HusseinGEJ17}
\bibfield{author}{\bibinfo{person}{Ahmed Hussein}, \bibinfo{person}{Mohamed~Medhat Gaber}, \bibinfo{person}{Eyad Elyan}, {and} \bibinfo{person}{Chrisina Jayne}.} \bibinfo{year}{2017}\natexlab{}.
\newblock \showarticletitle{Imitation Learning: {A} Survey of Learning Methods}.
\newblock \bibinfo{journal}{\emph{{ACM} Comput. Surv.}} \bibinfo{volume}{50}, \bibinfo{number}{2} (\bibinfo{year}{2017}), \bibinfo{pages}{21:1--21:35}.
\newblock


\bibitem[\protect\citeauthoryear{Jacq, Ferret, Pietquin, and Geist}{Jacq et~al\mbox{.}}{2022}]%
        {jacq2022lazy}
\bibfield{author}{\bibinfo{person}{Alexis Jacq}, \bibinfo{person}{Johan Ferret}, \bibinfo{person}{Olivier Pietquin}, {and} \bibinfo{person}{Matthieu Geist}.} \bibinfo{year}{2022}\natexlab{}.
\newblock \showarticletitle{Lazy-MDPs: Towards Interpretable RL by Learning When to Act}. In \bibinfo{booktitle}{\emph{International Conference on Autonomous Agents and Multiagent Systems}}. \bibinfo{pages}{669--677}.
\newblock


\bibitem[\protect\citeauthoryear{Jin, Allen-Zhu, Bubeck, and Jordan}{Jin et~al\mbox{.}}{2018}]%
        {jin2018q}
\bibfield{author}{\bibinfo{person}{Chi Jin}, \bibinfo{person}{Zeyuan Allen-Zhu}, \bibinfo{person}{Sebastien Bubeck}, {and} \bibinfo{person}{Michael~I Jordan}.} \bibinfo{year}{2018}\natexlab{}.
\newblock \showarticletitle{Is Q-learning provably efficient?}
\newblock \bibinfo{journal}{\emph{Advances in neural information processing systems}}  \bibinfo{volume}{31} (\bibinfo{year}{2018}).
\newblock


\bibitem[\protect\citeauthoryear{Kang, Jie, and Feng}{Kang et~al\mbox{.}}{2018}]%
        {kang2018policy}
\bibfield{author}{\bibinfo{person}{Bingyi Kang}, \bibinfo{person}{Zequn Jie}, {and} \bibinfo{person}{Jiashi Feng}.} \bibinfo{year}{2018}\natexlab{}.
\newblock \showarticletitle{Policy optimization with demonstrations}. In \bibinfo{booktitle}{\emph{International conference on machine learning}}. PMLR, \bibinfo{pages}{2469--2478}.
\newblock


\bibitem[\protect\citeauthoryear{Khachaturov}{Khachaturov}{2021}]%
        {khachaturov2021susceptibility}
\bibfield{author}{\bibinfo{person}{David~Grigorovich Khachaturov}.} \bibinfo{year}{2021}\natexlab{}.
\newblock \showarticletitle{Susceptibility of Hierarchical Reinforcement Learning to Adversarial Examples Computer Science Tripos--Part II Churchill College}.
\newblock  (\bibinfo{year}{2021}).
\newblock


\bibitem[\protect\citeauthoryear{Khalil}{Khalil}{2002}]%
        {Khalilbook}
\bibfield{author}{\bibinfo{person}{Hassan~K Khalil}.} \bibinfo{year}{2002}\natexlab{}.
\newblock \bibinfo{booktitle}{\emph{{Nonlinear systems; 3rd ed.}}}
\newblock \bibinfo{publisher}{Prentice-Hall}, \bibinfo{address}{Upper Saddle River, NJ}.
\newblock
\urldef\tempurl%
\url{https://cds.cern.ch/record/1173048}
\showURL{%
\tempurl}
\newblock
\shownote{The book can be consulted by contacting: PH-AID: Wallet, Lionel.}


\bibitem[\protect\citeauthoryear{Knox and Stone}{Knox and Stone}{2010}]%
        {knox2010combining}
\bibfield{author}{\bibinfo{person}{W~Bradley Knox} {and} \bibinfo{person}{Peter Stone}.} \bibinfo{year}{2010}\natexlab{}.
\newblock \showarticletitle{Combining manual feedback with subsequent MDP reward signals for reinforcement learning.}. In \bibinfo{booktitle}{\emph{International Conference on Autonomous Agents and Multiagent Systems}}. \bibinfo{pages}{5--12}.
\newblock


\bibitem[\protect\citeauthoryear{Knox and Stone}{Knox and Stone}{2011}]%
        {knox2011augmenting}
\bibfield{author}{\bibinfo{person}{W~Bradley Knox} {and} \bibinfo{person}{Peter Stone}.} \bibinfo{year}{2011}\natexlab{}.
\newblock \showarticletitle{Augmenting reinforcement learning with human feedback}. In \bibinfo{booktitle}{\emph{ICML 2011 Workshop on New Developments in Imitation Learning (July 2011)}}, Vol.~\bibinfo{volume}{855}. \bibinfo{pages}{3}.
\newblock


\bibitem[\protect\citeauthoryear{Knox and Stone}{Knox and Stone}{2012}]%
        {knox2012reinforcement}
\bibfield{author}{\bibinfo{person}{W~Bradley Knox} {and} \bibinfo{person}{Peter Stone}.} \bibinfo{year}{2012}\natexlab{}.
\newblock \showarticletitle{Reinforcement learning from simultaneous human and MDP reward.}. In \bibinfo{booktitle}{\emph{International Conference on Autonomous Agents and Multiagent Systems}}. \bibinfo{pages}{475--482}.
\newblock


\bibitem[\protect\citeauthoryear{Kober and Peters}{Kober and Peters}{2008}]%
        {kober2008policy}
\bibfield{author}{\bibinfo{person}{Jens Kober} {and} \bibinfo{person}{Jan Peters}.} \bibinfo{year}{2008}\natexlab{}.
\newblock \showarticletitle{Policy search for motor primitives in robotics}.
\newblock \bibinfo{journal}{\emph{Advances in neural information processing systems}}  \bibinfo{volume}{21} (\bibinfo{year}{2008}).
\newblock


\bibitem[\protect\citeauthoryear{Kostrikov, Fergus, Tompson, and Nachum}{Kostrikov et~al\mbox{.}}{2021}]%
        {kostrikov2021offline}
\bibfield{author}{\bibinfo{person}{Ilya Kostrikov}, \bibinfo{person}{Rob Fergus}, \bibinfo{person}{Jonathan Tompson}, {and} \bibinfo{person}{Ofir Nachum}.} \bibinfo{year}{2021}\natexlab{}.
\newblock \showarticletitle{Offline reinforcement learning with fisher divergence critic regularization}. In \bibinfo{booktitle}{\emph{International conference on machine learning}}. PMLR, \bibinfo{pages}{5774--5783}.
\newblock


\bibitem[\protect\citeauthoryear{Kumar, Fu, Soh, Tucker, and Levine}{Kumar et~al\mbox{.}}{2019}]%
        {kumar2019stabilizing}
\bibfield{author}{\bibinfo{person}{Aviral Kumar}, \bibinfo{person}{Justin Fu}, \bibinfo{person}{Matthew Soh}, \bibinfo{person}{George Tucker}, {and} \bibinfo{person}{Sergey Levine}.} \bibinfo{year}{2019}\natexlab{}.
\newblock \showarticletitle{Stabilizing off-policy q-learning via bootstrapping error reduction}.
\newblock \bibinfo{journal}{\emph{Advances in Neural Information Processing Systems}}  \bibinfo{volume}{32} (\bibinfo{year}{2019}).
\newblock


\bibitem[\protect\citeauthoryear{Kumar, Zhou, Tucker, and Levine}{Kumar et~al\mbox{.}}{2020}]%
        {kumar2020conservative}
\bibfield{author}{\bibinfo{person}{Aviral Kumar}, \bibinfo{person}{Aurick Zhou}, \bibinfo{person}{George Tucker}, {and} \bibinfo{person}{Sergey Levine}.} \bibinfo{year}{2020}\natexlab{}.
\newblock \showarticletitle{Conservative q-learning for offline reinforcement learning}.
\newblock \bibinfo{journal}{\emph{Advances in Neural Information Processing Systems}}  \bibinfo{volume}{33} (\bibinfo{year}{2020}), \bibinfo{pages}{1179--1191}.
\newblock


\bibitem[\protect\citeauthoryear{Lasserre, Henrion, Prieur, and Tr{\'e}lat}{Lasserre et~al\mbox{.}}{2008}]%
        {lasserre2008nonlinear}
\bibfield{author}{\bibinfo{person}{Jean~B Lasserre}, \bibinfo{person}{Didier Henrion}, \bibinfo{person}{Christophe Prieur}, {and} \bibinfo{person}{Emmanuel Tr{\'e}lat}.} \bibinfo{year}{2008}\natexlab{}.
\newblock \showarticletitle{Nonlinear optimal control via occupation measures and LMI-relaxations}.
\newblock \bibinfo{journal}{\emph{SIAM journal on control and optimization}} \bibinfo{volume}{47}, \bibinfo{number}{4} (\bibinfo{year}{2008}), \bibinfo{pages}{1643--1666}.
\newblock


\bibitem[\protect\citeauthoryear{Lee, Seo, Lee, Abbeel, and Shin}{Lee et~al\mbox{.}}{2022}]%
        {lee2022offline}
\bibfield{author}{\bibinfo{person}{Seunghyun Lee}, \bibinfo{person}{Younggyo Seo}, \bibinfo{person}{Kimin Lee}, \bibinfo{person}{Pieter Abbeel}, {and} \bibinfo{person}{Jinwoo Shin}.} \bibinfo{year}{2022}\natexlab{}.
\newblock \showarticletitle{Offline-to-online reinforcement learning via balanced replay and pessimistic q-ensemble}. In \bibinfo{booktitle}{\emph{Conference on Robot Learning}}. PMLR, \bibinfo{pages}{1702--1712}.
\newblock


\bibitem[\protect\citeauthoryear{Lee, Sun, Somasundaram, Hu, and Lim}{Lee et~al\mbox{.}}{2018}]%
        {lee2018composing}
\bibfield{author}{\bibinfo{person}{Youngwoon Lee}, \bibinfo{person}{Shao-Hua Sun}, \bibinfo{person}{Sriram Somasundaram}, \bibinfo{person}{Edward~S Hu}, {and} \bibinfo{person}{Joseph~J Lim}.} \bibinfo{year}{2018}\natexlab{}.
\newblock \showarticletitle{Composing complex skills by learning transition policies}. In \bibinfo{booktitle}{\emph{International Conference on Learning Representations}}.
\newblock


\bibitem[\protect\citeauthoryear{Leith and Leithead}{Leith and Leithead}{2000}]%
        {leith2000survey}
\bibfield{author}{\bibinfo{person}{Douglas~J Leith} {and} \bibinfo{person}{William~E Leithead}.} \bibinfo{year}{2000}\natexlab{}.
\newblock \showarticletitle{Survey of gain-scheduling analysis and design}.
\newblock \bibinfo{journal}{\emph{International journal of control}} \bibinfo{volume}{73}, \bibinfo{number}{11} (\bibinfo{year}{2000}), \bibinfo{pages}{1001--1025}.
\newblock


\bibitem[\protect\citeauthoryear{Levine, Kumar, Tucker, and Fu}{Levine et~al\mbox{.}}{2020}]%
        {levine2020offline}
\bibfield{author}{\bibinfo{person}{Sergey Levine}, \bibinfo{person}{Aviral Kumar}, \bibinfo{person}{George Tucker}, {and} \bibinfo{person}{Justin Fu}.} \bibinfo{year}{2020}\natexlab{}.
\newblock \showarticletitle{Offline reinforcement learning: Tutorial, review, and perspectives on open problems}.
\newblock \bibinfo{journal}{\emph{arXiv preprint arXiv:2005.01643}} (\bibinfo{year}{2020}).
\newblock


\bibitem[\protect\citeauthoryear{Lewis, Vrabie, and Syrmos}{Lewis et~al\mbox{.}}{2012}]%
        {lewis2012optimal}
\bibfield{author}{\bibinfo{person}{Frank~L Lewis}, \bibinfo{person}{Draguna Vrabie}, {and} \bibinfo{person}{Vassilis~L Syrmos}.} \bibinfo{year}{2012}\natexlab{}.
\newblock \bibinfo{booktitle}{\emph{Optimal control}}.
\newblock \bibinfo{publisher}{John Wiley \& Sons}.
\newblock


\bibitem[\protect\citeauthoryear{Lillicrap, Hunt, Pritzel, Heess, Erez, Tassa, Silver, and Wierstra}{Lillicrap et~al\mbox{.}}{2015}]%
        {lillicrap2015continuous}
\bibfield{author}{\bibinfo{person}{Timothy~P Lillicrap}, \bibinfo{person}{Jonathan~J Hunt}, \bibinfo{person}{Alexander Pritzel}, \bibinfo{person}{Nicolas Heess}, \bibinfo{person}{Tom Erez}, \bibinfo{person}{Yuval Tassa}, \bibinfo{person}{David Silver}, {and} \bibinfo{person}{Daan Wierstra}.} \bibinfo{year}{2015}\natexlab{}.
\newblock \showarticletitle{Continuous control with deep reinforcement learning}.
\newblock \bibinfo{journal}{\emph{arXiv preprint arXiv:1509.02971}} (\bibinfo{year}{2015}).
\newblock


\bibitem[\protect\citeauthoryear{Lyu, Ma, Li, and Lu}{Lyu et~al\mbox{.}}{2022}]%
        {lyu2022mildly}
\bibfield{author}{\bibinfo{person}{Jiafei Lyu}, \bibinfo{person}{Xiaoteng Ma}, \bibinfo{person}{Xiu Li}, {and} \bibinfo{person}{Zongqing Lu}.} \bibinfo{year}{2022}\natexlab{}.
\newblock \showarticletitle{Mildly conservative Q-learning for offline reinforcement learning}.
\newblock \bibinfo{journal}{\emph{arXiv preprint arXiv:2206.04745}} (\bibinfo{year}{2022}).
\newblock


\bibitem[\protect\citeauthoryear{Mataric}{Mataric}{1994}]%
        {mataric1994reward}
\bibfield{author}{\bibinfo{person}{Maja~J Mataric}.} \bibinfo{year}{1994}\natexlab{}.
\newblock \showarticletitle{Reward functions for accelerated learning}.
\newblock In \bibinfo{booktitle}{\emph{Machine learning proceedings 1994}}. \bibinfo{publisher}{Elsevier}, \bibinfo{pages}{181--189}.
\newblock


\bibitem[\protect\citeauthoryear{Mayne, Rawlings, Rao, and Scokaert}{Mayne et~al\mbox{.}}{2000}]%
        {mayne2000constrained}
\bibfield{author}{\bibinfo{person}{David~Q Mayne}, \bibinfo{person}{James~B Rawlings}, \bibinfo{person}{Christopher~V Rao}, {and} \bibinfo{person}{Pierre~OM Scokaert}.} \bibinfo{year}{2000}\natexlab{}.
\newblock \showarticletitle{Constrained model predictive control: Stability and optimality}.
\newblock \bibinfo{journal}{\emph{Automatica}} \bibinfo{volume}{36}, \bibinfo{number}{6} (\bibinfo{year}{2000}), \bibinfo{pages}{789--814}.
\newblock


\bibitem[\protect\citeauthoryear{Mnih, Badia, Mirza, Graves, Lillicrap, Harley, Silver, and Kavukcuoglu}{Mnih et~al\mbox{.}}{2016}]%
        {mnih2016asynchronous}
\bibfield{author}{\bibinfo{person}{Volodymyr Mnih}, \bibinfo{person}{Adria~Puigdomenech Badia}, \bibinfo{person}{Mehdi Mirza}, \bibinfo{person}{Alex Graves}, \bibinfo{person}{Timothy Lillicrap}, \bibinfo{person}{Tim Harley}, \bibinfo{person}{David Silver}, {and} \bibinfo{person}{Koray Kavukcuoglu}.} \bibinfo{year}{2016}\natexlab{}.
\newblock \showarticletitle{Asynchronous methods for deep reinforcement learning}. In \bibinfo{booktitle}{\emph{International conference on machine learning}}. PMLR, \bibinfo{pages}{1928--1937}.
\newblock


\bibitem[\protect\citeauthoryear{Mnih, Kavukcuoglu, Silver, Graves, Antonoglou, Wierstra, and Riedmiller}{Mnih et~al\mbox{.}}{2013}]%
        {MnihKSGAWR13}
\bibfield{author}{\bibinfo{person}{Volodymyr Mnih}, \bibinfo{person}{Koray Kavukcuoglu}, \bibinfo{person}{David Silver}, \bibinfo{person}{Alex Graves}, \bibinfo{person}{Ioannis Antonoglou}, \bibinfo{person}{Daan Wierstra}, {and} \bibinfo{person}{Martin Riedmiller}.} \bibinfo{year}{2013}\natexlab{}.
\newblock \showarticletitle{Playing atari with deep reinforcement learning}.
\newblock \bibinfo{journal}{\emph{arXiv preprint arXiv:1312.5602}} (\bibinfo{year}{2013}).
\newblock


\bibitem[\protect\citeauthoryear{Nachum, Gu, Lee, and Levine}{Nachum et~al\mbox{.}}{2018}]%
        {nachum2018data}
\bibfield{author}{\bibinfo{person}{Ofir Nachum}, \bibinfo{person}{Shixiang~Shane Gu}, \bibinfo{person}{Honglak Lee}, {and} \bibinfo{person}{Sergey Levine}.} \bibinfo{year}{2018}\natexlab{}.
\newblock \showarticletitle{Data-efficient hierarchical reinforcement learning}.
\newblock \bibinfo{journal}{\emph{Advances in neural information processing systems}}  \bibinfo{volume}{31} (\bibinfo{year}{2018}).
\newblock


\bibitem[\protect\citeauthoryear{Nair, Gupta, Dalal, and Levine}{Nair et~al\mbox{.}}{2020}]%
        {nair2020awac}
\bibfield{author}{\bibinfo{person}{Ashvin Nair}, \bibinfo{person}{Abhishek Gupta}, \bibinfo{person}{Murtaza Dalal}, {and} \bibinfo{person}{Sergey Levine}.} \bibinfo{year}{2020}\natexlab{}.
\newblock \showarticletitle{Awac: Accelerating online reinforcement learning with offline datasets}.
\newblock \bibinfo{journal}{\emph{arXiv preprint arXiv:2006.09359}} (\bibinfo{year}{2020}).
\newblock


\bibitem[\protect\citeauthoryear{Nair, McGrew, Andrychowicz, Zaremba, and Abbeel}{Nair et~al\mbox{.}}{2018}]%
        {nair2018overcoming}
\bibfield{author}{\bibinfo{person}{Ashvin Nair}, \bibinfo{person}{Bob McGrew}, \bibinfo{person}{Marcin Andrychowicz}, \bibinfo{person}{Wojciech Zaremba}, {and} \bibinfo{person}{Pieter Abbeel}.} \bibinfo{year}{2018}\natexlab{}.
\newblock \showarticletitle{Overcoming exploration in reinforcement learning with demonstrations}. In \bibinfo{booktitle}{\emph{2018 IEEE international conference on robotics and automation (ICRA)}}. IEEE, \bibinfo{pages}{6292--6299}.
\newblock


\bibitem[\protect\citeauthoryear{Ng, Harada, and Russell}{Ng et~al\mbox{.}}{1999a}]%
        {ng1999policy}
\bibfield{author}{\bibinfo{person}{Andrew~Y Ng}, \bibinfo{person}{Daishi Harada}, {and} \bibinfo{person}{Stuart Russell}.} \bibinfo{year}{1999}\natexlab{a}.
\newblock \showarticletitle{Policy invariance under reward transformations: Theory and application to reward shaping}. In \bibinfo{booktitle}{\emph{International conference on machine learning}}, Vol.~\bibinfo{volume}{99}. \bibinfo{pages}{278--287}.
\newblock


\bibitem[\protect\citeauthoryear{Ng, Harada, and Russell}{Ng et~al\mbox{.}}{1999b}]%
        {NgHR99}
\bibfield{author}{\bibinfo{person}{Andrew~Y. Ng}, \bibinfo{person}{Daishi Harada}, {and} \bibinfo{person}{Stuart~J. Russell}.} \bibinfo{year}{1999}\natexlab{b}.
\newblock \showarticletitle{Policy Invariance Under Reward Transformations: Theory and Application to Reward Shaping}. In \bibinfo{booktitle}{\emph{{International conference on machine learning}}}. \bibinfo{publisher}{Morgan Kaufmann}, \bibinfo{pages}{278--287}.
\newblock


\bibitem[\protect\citeauthoryear{Peters and Schaal}{Peters and Schaal}{2008}]%
        {peters2008reinforcement}
\bibfield{author}{\bibinfo{person}{Jan Peters} {and} \bibinfo{person}{Stefan Schaal}.} \bibinfo{year}{2008}\natexlab{}.
\newblock \showarticletitle{Reinforcement learning of motor skills with policy gradients}.
\newblock \bibinfo{journal}{\emph{Neural networks}} \bibinfo{volume}{21}, \bibinfo{number}{4} (\bibinfo{year}{2008}), \bibinfo{pages}{682--697}.
\newblock


\bibitem[\protect\citeauthoryear{Pohlen, Piot, Hester, Azar, Horgan, Budden, Barth-Maron, Van~Hasselt, Quan, Ve{\v{c}}er{\'\i}k, et~al\mbox{.}}{Pohlen et~al\mbox{.}}{2018}]%
        {pohlen2018observe}
\bibfield{author}{\bibinfo{person}{Tobias Pohlen}, \bibinfo{person}{Bilal Piot}, \bibinfo{person}{Todd Hester}, \bibinfo{person}{Mohammad~Gheshlaghi Azar}, \bibinfo{person}{Dan Horgan}, \bibinfo{person}{David Budden}, \bibinfo{person}{Gabriel Barth-Maron}, \bibinfo{person}{Hado Van~Hasselt}, \bibinfo{person}{John Quan}, \bibinfo{person}{Mel Ve{\v{c}}er{\'\i}k}, {et~al\mbox{.}}} \bibinfo{year}{2018}\natexlab{}.
\newblock \showarticletitle{Observe and look further: Achieving consistent performance on atari}.
\newblock \bibinfo{journal}{\emph{arXiv preprint arXiv:1805.11593}} (\bibinfo{year}{2018}).
\newblock


\bibitem[\protect\citeauthoryear{Pomerleau}{Pomerleau}{1988}]%
        {pomerleau1988alvinn}
\bibfield{author}{\bibinfo{person}{Dean~A Pomerleau}.} \bibinfo{year}{1988}\natexlab{}.
\newblock \showarticletitle{Alvinn: An autonomous land vehicle in a neural network}.
\newblock \bibinfo{journal}{\emph{Advances in neural information processing systems}}  \bibinfo{volume}{1} (\bibinfo{year}{1988}).
\newblock


\bibitem[\protect\citeauthoryear{Randl{\o}v and Alstr{\o}m}{Randl{\o}v and Alstr{\o}m}{1998}]%
        {randlov1998learning}
\bibfield{author}{\bibinfo{person}{Jette Randl{\o}v} {and} \bibinfo{person}{Preben Alstr{\o}m}.} \bibinfo{year}{1998}\natexlab{}.
\newblock \showarticletitle{Learning to Drive a Bicycle Using Reinforcement Learning and Shaping.}. In \bibinfo{booktitle}{\emph{International conference on machine learning}}, Vol.~\bibinfo{volume}{98}. Citeseer, \bibinfo{pages}{463--471}.
\newblock


\bibitem[\protect\citeauthoryear{Ravichandar, Polydoros, Chernova, and Billard}{Ravichandar et~al\mbox{.}}{2020}]%
        {ravichandar2020recent}
\bibfield{author}{\bibinfo{person}{Harish Ravichandar}, \bibinfo{person}{Athanasios~S Polydoros}, \bibinfo{person}{Sonia Chernova}, {and} \bibinfo{person}{Aude Billard}.} \bibinfo{year}{2020}\natexlab{}.
\newblock \showarticletitle{Recent advances in robot learning from demonstration}.
\newblock \bibinfo{journal}{\emph{Annual Review of Control, Robotics, and Autonomous Systems}}  \bibinfo{volume}{3} (\bibinfo{year}{2020}), \bibinfo{pages}{297--330}.
\newblock


\bibitem[\protect\citeauthoryear{Rengarajan, Vaidya, Sarvesh, Kalathil, and Shakkottai}{Rengarajan et~al\mbox{.}}{2022}]%
        {rengarajan2022reinforcement}
\bibfield{author}{\bibinfo{person}{D Rengarajan}, \bibinfo{person}{G Vaidya}, \bibinfo{person}{A Sarvesh}, \bibinfo{person}{D Kalathil}, {and} \bibinfo{person}{S Shakkottai}.} \bibinfo{year}{2022}\natexlab{}.
\newblock \showarticletitle{Reinforcement Learning with Sparse Rewards using Guidance from Offline Demonstration}. In \bibinfo{booktitle}{\emph{International Conference on Learning Representations)}}.
\newblock


\bibitem[\protect\citeauthoryear{Rojanavasu, Srinil, and Pinngern}{Rojanavasu et~al\mbox{.}}{2005}]%
        {rojanavasu2005new}
\bibfield{author}{\bibinfo{person}{Pornthep Rojanavasu}, \bibinfo{person}{Phaitoon Srinil}, {and} \bibinfo{person}{Ouen Pinngern}.} \bibinfo{year}{2005}\natexlab{}.
\newblock \showarticletitle{New recommendation system using reinforcement learning}.
\newblock \bibinfo{journal}{\emph{Special Issue of the Intl. J. Computer, the Internet and Management}} \bibinfo{volume}{13}, \bibinfo{number}{SP 3} (\bibinfo{year}{2005}).
\newblock


\bibitem[\protect\citeauthoryear{Ross and Bagnell}{Ross and Bagnell}{2010}]%
        {ross2010efficient}
\bibfield{author}{\bibinfo{person}{St{\'e}phane Ross} {and} \bibinfo{person}{Drew Bagnell}.} \bibinfo{year}{2010}\natexlab{}.
\newblock \showarticletitle{Efficient reductions for imitation learning}. In \bibinfo{booktitle}{\emph{International conference on artificial intelligence and statistics}}. JMLR Workshop and Conference Proceedings, \bibinfo{pages}{661--668}.
\newblock


\bibitem[\protect\citeauthoryear{Ross, Gordon, and Bagnell}{Ross et~al\mbox{.}}{2011}]%
        {ross2011reduction}
\bibfield{author}{\bibinfo{person}{St{\'e}phane Ross}, \bibinfo{person}{Geoffrey Gordon}, {and} \bibinfo{person}{Drew Bagnell}.} \bibinfo{year}{2011}\natexlab{}.
\newblock \showarticletitle{A reduction of imitation learning and structured prediction to no-regret online learning}. In \bibinfo{booktitle}{\emph{International conference on artificial intelligence and statistics}}. JMLR Workshop and Conference Proceedings, \bibinfo{pages}{627--635}.
\newblock


\bibitem[\protect\citeauthoryear{Rugh and Shamma}{Rugh and Shamma}{2000}]%
        {rugh2000research}
\bibfield{author}{\bibinfo{person}{Wilson~J Rugh} {and} \bibinfo{person}{Jeff~S Shamma}.} \bibinfo{year}{2000}\natexlab{}.
\newblock \showarticletitle{Research on gain scheduling}.
\newblock \bibinfo{journal}{\emph{Automatica}} \bibinfo{volume}{36}, \bibinfo{number}{10} (\bibinfo{year}{2000}), \bibinfo{pages}{1401--1425}.
\newblock


\bibitem[\protect\citeauthoryear{Sammut, Hurst, Kedzier, and Michie}{Sammut et~al\mbox{.}}{1992}]%
        {SammutHKM92}
\bibfield{author}{\bibinfo{person}{Claude Sammut}, \bibinfo{person}{Scott Hurst}, \bibinfo{person}{Dana Kedzier}, {and} \bibinfo{person}{Donald Michie}.} \bibinfo{year}{1992}\natexlab{}.
\newblock \showarticletitle{Learning to Fly}. In \bibinfo{booktitle}{\emph{{ML}}}. \bibinfo{publisher}{Morgan Kaufmann}, \bibinfo{pages}{385--393}.
\newblock


\bibitem[\protect\citeauthoryear{Schaal}{Schaal}{1996}]%
        {schaal1996learning}
\bibfield{author}{\bibinfo{person}{Stefan Schaal}.} \bibinfo{year}{1996}\natexlab{}.
\newblock \showarticletitle{Learning from demonstration}.
\newblock \bibinfo{journal}{\emph{Advances in neural information processing systems}}  \bibinfo{volume}{9} (\bibinfo{year}{1996}).
\newblock


\bibitem[\protect\citeauthoryear{Schmitt, Hudson, Zidek, Osindero, Doersch, Czarnecki, Leibo, Kuttler, Zisserman, Simonyan, et~al\mbox{.}}{Schmitt et~al\mbox{.}}{2018}]%
        {schmitt2018kickstarting}
\bibfield{author}{\bibinfo{person}{Simon Schmitt}, \bibinfo{person}{Jonathan~J Hudson}, \bibinfo{person}{Augustin Zidek}, \bibinfo{person}{Simon Osindero}, \bibinfo{person}{Carl Doersch}, \bibinfo{person}{Wojciech~M Czarnecki}, \bibinfo{person}{Joel~Z Leibo}, \bibinfo{person}{Heinrich Kuttler}, \bibinfo{person}{Andrew Zisserman}, \bibinfo{person}{Karen Simonyan}, {et~al\mbox{.}}} \bibinfo{year}{2018}\natexlab{}.
\newblock \showarticletitle{Kickstarting deep reinforcement learning}.
\newblock \bibinfo{journal}{\emph{arXiv preprint arXiv:1803.03835}} (\bibinfo{year}{2018}).
\newblock


\bibitem[\protect\citeauthoryear{Schulman, Wolski, Dhariwal, Radford, and Klimov}{Schulman et~al\mbox{.}}{2017}]%
        {schulman2017proximal}
\bibfield{author}{\bibinfo{person}{John Schulman}, \bibinfo{person}{Filip Wolski}, \bibinfo{person}{Prafulla Dhariwal}, \bibinfo{person}{Alec Radford}, {and} \bibinfo{person}{Oleg Klimov}.} \bibinfo{year}{2017}\natexlab{}.
\newblock \showarticletitle{Proximal policy optimization algorithms}.
\newblock \bibinfo{journal}{\emph{arXiv preprint arXiv:1707.06347}} (\bibinfo{year}{2017}).
\newblock


\bibitem[\protect\citeauthoryear{Singh, Yu, Yang, Zhang, Kumar, and Levine}{Singh et~al\mbox{.}}{2020}]%
        {singh2020cog}
\bibfield{author}{\bibinfo{person}{Avi Singh}, \bibinfo{person}{Albert Yu}, \bibinfo{person}{Jonathan Yang}, \bibinfo{person}{Jesse Zhang}, \bibinfo{person}{Aviral Kumar}, {and} \bibinfo{person}{Sergey Levine}.} \bibinfo{year}{2020}\natexlab{}.
\newblock \showarticletitle{Cog: Connecting new skills to past experience with offline reinforcement learning}.
\newblock \bibinfo{journal}{\emph{arXiv preprint arXiv:2010.14500}} (\bibinfo{year}{2020}).
\newblock


\bibitem[\protect\citeauthoryear{Sutton and Barto}{Sutton and Barto}{1998}]%
        {SuttonB98}
\bibfield{author}{\bibinfo{person}{Richard~S. Sutton} {and} \bibinfo{person}{Andrew~G. Barto}.} \bibinfo{year}{1998}\natexlab{}.
\newblock \bibinfo{booktitle}{\emph{Reinforcement learning - an introduction}}.
\newblock \bibinfo{publisher}{{MIT} Press}.
\newblock


\bibitem[\protect\citeauthoryear{Tassa, Doron, Muldal, Erez, Li, Casas, Budden, Abdolmaleki, Merel, Lefrancq, et~al\mbox{.}}{Tassa et~al\mbox{.}}{2018}]%
        {tassa2018deepmind}
\bibfield{author}{\bibinfo{person}{Yuval Tassa}, \bibinfo{person}{Yotam Doron}, \bibinfo{person}{Alistair Muldal}, \bibinfo{person}{Tom Erez}, \bibinfo{person}{Yazhe Li}, \bibinfo{person}{Diego de~Las Casas}, \bibinfo{person}{David Budden}, \bibinfo{person}{Abbas Abdolmaleki}, \bibinfo{person}{Josh Merel}, \bibinfo{person}{Andrew Lefrancq}, {et~al\mbox{.}}} \bibinfo{year}{2018}\natexlab{}.
\newblock \showarticletitle{Deepmind control suite}.
\newblock \bibinfo{journal}{\emph{arXiv preprint arXiv:1801.00690}} (\bibinfo{year}{2018}).
\newblock


\bibitem[\protect\citeauthoryear{Taylor, Dorobantu, Dean, Recht, Yue, and Ames}{Taylor et~al\mbox{.}}{2021}]%
        {taylor2021towards}
\bibfield{author}{\bibinfo{person}{Andrew~J Taylor}, \bibinfo{person}{Victor~D Dorobantu}, \bibinfo{person}{Sarah Dean}, \bibinfo{person}{Benjamin Recht}, \bibinfo{person}{Yisong Yue}, {and} \bibinfo{person}{Aaron~D Ames}.} \bibinfo{year}{2021}\natexlab{}.
\newblock \showarticletitle{Towards robust data-driven control synthesis for nonlinear systems with actuation uncertainty}. In \bibinfo{booktitle}{\emph{2021 60th IEEE Conference on Decision and Control (CDC)}}. IEEE, \bibinfo{pages}{6469--6476}.
\newblock


\bibitem[\protect\citeauthoryear{Teh, Bapst, Czarnecki, Quan, Kirkpatrick, Hadsell, Heess, and Pascanu}{Teh et~al\mbox{.}}{2017}]%
        {teh2017distral}
\bibfield{author}{\bibinfo{person}{Yee Teh}, \bibinfo{person}{Victor Bapst}, \bibinfo{person}{Wojciech~M Czarnecki}, \bibinfo{person}{John Quan}, \bibinfo{person}{James Kirkpatrick}, \bibinfo{person}{Raia Hadsell}, \bibinfo{person}{Nicolas Heess}, {and} \bibinfo{person}{Razvan Pascanu}.} \bibinfo{year}{2017}\natexlab{}.
\newblock \showarticletitle{Distral: Robust multitask reinforcement learning}.
\newblock \bibinfo{journal}{\emph{Advances in neural information processing systems}}  \bibinfo{volume}{30} (\bibinfo{year}{2017}).
\newblock


\bibitem[\protect\citeauthoryear{Todorov}{Todorov}{2006}]%
        {todorov2006optimal}
\bibfield{author}{\bibinfo{person}{Emanuel Todorov}.} \bibinfo{year}{2006}\natexlab{}.
\newblock \showarticletitle{Optimal control theory}.
\newblock \bibinfo{journal}{\emph{Bayesian brain: probabilistic approaches to neural coding}} (\bibinfo{year}{2006}), \bibinfo{pages}{268--298}.
\newblock


\bibitem[\protect\citeauthoryear{Todorov, Erez, and Tassa}{Todorov et~al\mbox{.}}{2012}]%
        {todorov2012mujoco}
\bibfield{author}{\bibinfo{person}{Emanuel Todorov}, \bibinfo{person}{Tom Erez}, {and} \bibinfo{person}{Yuval Tassa}.} \bibinfo{year}{2012}\natexlab{}.
\newblock \showarticletitle{Mujoco: A physics engine for model-based control}. In \bibinfo{booktitle}{\emph{2012 IEEE/RSJ international conference on intelligent robots and systems}}. IEEE, \bibinfo{pages}{5026--5033}.
\newblock


\bibitem[\protect\citeauthoryear{Turchetta, Kolobov, Shah, Krause, and Agarwal}{Turchetta et~al\mbox{.}}{2020}]%
        {turchetta2020safe}
\bibfield{author}{\bibinfo{person}{Matteo Turchetta}, \bibinfo{person}{Andrey Kolobov}, \bibinfo{person}{Shital Shah}, \bibinfo{person}{Andreas Krause}, {and} \bibinfo{person}{Alekh Agarwal}.} \bibinfo{year}{2020}\natexlab{}.
\newblock \showarticletitle{Safe reinforcement learning via curriculum induction}.
\newblock \bibinfo{journal}{\emph{Advances in Neural Information Processing Systems}}  \bibinfo{volume}{33} (\bibinfo{year}{2020}), \bibinfo{pages}{12151--12162}.
\newblock


\bibitem[\protect\citeauthoryear{Uchendu, Xiao, Lu, Zhu, Yan, Simon, Bennice, Fu, Ma, Jiao, et~al\mbox{.}}{Uchendu et~al\mbox{.}}{2022}]%
        {uchendu2022jump}
\bibfield{author}{\bibinfo{person}{Ikechukwu Uchendu}, \bibinfo{person}{Ted Xiao}, \bibinfo{person}{Yao Lu}, \bibinfo{person}{Banghua Zhu}, \bibinfo{person}{Mengyuan Yan}, \bibinfo{person}{Jos{\'e}phine Simon}, \bibinfo{person}{Matthew Bennice}, \bibinfo{person}{Chuyuan Fu}, \bibinfo{person}{Cong Ma}, \bibinfo{person}{Jiantao Jiao}, {et~al\mbox{.}}} \bibinfo{year}{2022}\natexlab{}.
\newblock \showarticletitle{Jump-Start Reinforcement Learning}.
\newblock \bibinfo{journal}{\emph{arXiv preprint arXiv:2204.02372}} (\bibinfo{year}{2022}).
\newblock


\bibitem[\protect\citeauthoryear{Vecerik, Hester, Scholz, Wang, Pietquin, Piot, Heess, Roth{\"o}rl, Lampe, and Riedmiller}{Vecerik et~al\mbox{.}}{2017}]%
        {vecerik2017leveraging}
\bibfield{author}{\bibinfo{person}{Mel Vecerik}, \bibinfo{person}{Todd Hester}, \bibinfo{person}{Jonathan Scholz}, \bibinfo{person}{Fumin Wang}, \bibinfo{person}{Olivier Pietquin}, \bibinfo{person}{Bilal Piot}, \bibinfo{person}{Nicolas Heess}, \bibinfo{person}{Thomas Roth{\"o}rl}, \bibinfo{person}{Thomas Lampe}, {and} \bibinfo{person}{Martin Riedmiller}.} \bibinfo{year}{2017}\natexlab{}.
\newblock \showarticletitle{Leveraging demonstrations for deep reinforcement learning on robotics problems with sparse rewards}.
\newblock \bibinfo{journal}{\emph{arXiv preprint arXiv:1707.08817}} (\bibinfo{year}{2017}).
\newblock


\bibitem[\protect\citeauthoryear{Wagener, Boots, and Cheng}{Wagener et~al\mbox{.}}{2021}]%
        {wagener2021safe}
\bibfield{author}{\bibinfo{person}{Nolan~C Wagener}, \bibinfo{person}{Byron Boots}, {and} \bibinfo{person}{Ching-An Cheng}.} \bibinfo{year}{2021}\natexlab{}.
\newblock \showarticletitle{Safe reinforcement learning using advantage-based intervention}. In \bibinfo{booktitle}{\emph{International conference on machine learning}}. PMLR, \bibinfo{pages}{10630--10640}.
\newblock


\bibitem[\protect\citeauthoryear{Wang and Hong}{Wang and Hong}{2020}]%
        {wang2020reinforcement}
\bibfield{author}{\bibinfo{person}{Zhe Wang} {and} \bibinfo{person}{Tianzhen Hong}.} \bibinfo{year}{2020}\natexlab{}.
\newblock \showarticletitle{Reinforcement learning for building controls: The opportunities and challenges}.
\newblock \bibinfo{journal}{\emph{Applied Energy}}  \bibinfo{volume}{269} (\bibinfo{year}{2020}), \bibinfo{pages}{115036}.
\newblock


\bibitem[\protect\citeauthoryear{Wu, Tucker, and Nachum}{Wu et~al\mbox{.}}{2019}]%
        {wu2019behavior}
\bibfield{author}{\bibinfo{person}{Yifan Wu}, \bibinfo{person}{George Tucker}, {and} \bibinfo{person}{Ofir Nachum}.} \bibinfo{year}{2019}\natexlab{}.
\newblock \showarticletitle{Behavior regularized offline reinforcement learning}.
\newblock \bibinfo{journal}{\emph{arXiv preprint arXiv:1911.11361}} (\bibinfo{year}{2019}).
\newblock


\bibitem[\protect\citeauthoryear{Xiao, Herman, Wagner, Ziesche, Etesami, and Linh}{Xiao et~al\mbox{.}}{2019}]%
        {xiao2019wasserstein}
\bibfield{author}{\bibinfo{person}{Huang Xiao}, \bibinfo{person}{Michael Herman}, \bibinfo{person}{Joerg Wagner}, \bibinfo{person}{Sebastian Ziesche}, \bibinfo{person}{Jalal Etesami}, {and} \bibinfo{person}{Thai~Hong Linh}.} \bibinfo{year}{2019}\natexlab{}.
\newblock \showarticletitle{Wasserstein adversarial imitation learning}.
\newblock \bibinfo{journal}{\emph{arXiv preprint arXiv:1906.08113}} (\bibinfo{year}{2019}).
\newblock


\bibitem[\protect\citeauthoryear{Zheng, Zhang, Zheng, Xiang, Yuan, Xie, and Li}{Zheng et~al\mbox{.}}{2018}]%
        {ZhengZZXY0L18}
\bibfield{author}{\bibinfo{person}{Guanjie Zheng}, \bibinfo{person}{Fuzheng Zhang}, \bibinfo{person}{Zihan Zheng}, \bibinfo{person}{Yang Xiang}, \bibinfo{person}{Nicholas~Jing Yuan}, \bibinfo{person}{Xing Xie}, {and} \bibinfo{person}{Zhenhui Li}.} \bibinfo{year}{2018}\natexlab{}.
\newblock \showarticletitle{{DRN:} {A} Deep Reinforcement Learning Framework for News Recommendation}. In \bibinfo{booktitle}{\emph{{WWW}}}. \bibinfo{publisher}{{ACM}}, \bibinfo{pages}{167--176}.
\newblock


\bibitem[\protect\citeauthoryear{Zimmer, Viappiani, and Weng}{Zimmer et~al\mbox{.}}{2014}]%
        {zimmer2014teacher}
\bibfield{author}{\bibinfo{person}{Matthieu Zimmer}, \bibinfo{person}{Paolo Viappiani}, {and} \bibinfo{person}{Paul Weng}.} \bibinfo{year}{2014}\natexlab{}.
\newblock \showarticletitle{Teacher-student framework: a reinforcement learning approach}. In \bibinfo{booktitle}{\emph{AAMAS Workshop Autonomous Robots and Multirobot Systems}}.
\newblock


\bibitem[\protect\citeauthoryear{Zoboli, Andrieu, Astolfi, Casadei, Dibangoye, and Nadri}{Zoboli et~al\mbox{.}}{2021}]%
        {zoboli2021reinforcement}
\bibfield{author}{\bibinfo{person}{Samuele Zoboli}, \bibinfo{person}{Vincent Andrieu}, \bibinfo{person}{Daniele Astolfi}, \bibinfo{person}{Giacomo Casadei}, \bibinfo{person}{Jilles~S Dibangoye}, {and} \bibinfo{person}{Madiha Nadri}.} \bibinfo{year}{2021}\natexlab{}.
\newblock \showarticletitle{Reinforcement learning policies with local LQR guarantees for nonlinear discrete-time systems}. In \bibinfo{booktitle}{\emph{2021 60th IEEE Conference on Decision and Control (CDC)}}. IEEE, \bibinfo{pages}{2258--2263}.
\newblock


\end{thebibliography}


\newpage 

\appendix
\onecolumn

\section{Extension to discrete actions} \label{app:discrete}

One possible way to extend the different methods  described in this work to a discrete action space is the following.

For PIG, the behavioral metric could be changed with any opposite distance between categorical distributions. 

PAG policy could choose the action proposed by the guide with probability $1 - \beta_{\text{PAG}}$ or take the action following $\pi_\theta$. This is not a strict generalization as the expectations over $\pi_{\text{PAG}}$ and its discrete version do not exactly coincide. Besides, the perturbation $\xi_\phi$ is no longer relevant because the action cannot be centered around the action of the local guide. However, the idea remains the same: the guide plays an important role at the beginning and the agent slowly takes over thanks to $\beta_{\text{PAG}}$. The discrete policy $\pi^{k, \text{discrete}}_{\text{PAG}}$ is formalized as follow:

\begin{gather*}
    \pi^{k, \text{discrete}}_{\text{PAG}}(\cdot | s) = \left\{
    \begin{array}{ll}
        a_{\text{e}}^s & \mbox{if } \lambda(s) \geq \lambda^- \mbox{with probability } 1 - \beta^k_{\text{PAG}}, \\
        \pi^k_\theta(\cdot | s) & \mbox{otherwise.}
    \end{array}
\right.
\end{gather*}

The policy recovers the control sharing method of \citep{knox2011augmenting}, although the two methods differ because of our modification of the Approximate Policy Evaluation step.

\section{Additional information for the experimental reproduction} \label{app:details_env}

In this section, we detail all the information required to reproduce the different experiments we performed.

In the guided environments, the local policy was a policy extracted from a SAC guide stopped during mid-training to make it sub-optimal. In the safety-critical environments, the local policy was scripted.

\paragraph{Ball in Cup} A ball is attached to a bottom of an actuated cup with a rope, and the goal is to put the ball inside the cup. The action space controls the horizontal and vertical force applied to the cup. The reward is sparse and is equal to $1$ when the ball is inside the cup and $0$ otherwise. Besides, this environment was slightly modified to make it more difficult. In the original environment from Deepmind Control Suite, the ball may be initialized to be above the cup which would directly help the exploration process. We made sure the ball is always initialized below the cup in our experiments. The local policy is an SAC agent with medium performances that was trained in the entire environment. The confidence function $\lambda$ is set to $1$ when the ball is below the cup and $0$ otherwise. 

\paragraph{Point-Mass} An actuated point-mass in a $2$D-plan must reach a small target. To achieve that purpose, it controls a force towards the $x$ and $y$ axes. The reward is sparse and provides a signal only when the point mass reaches the target. The local policy is a SAC agent with medium performances that was trained to reach a bigger goal than in the original environment. More exactly, we trained the local policy to reach a ball of radius $0.1$ while the original environment is a ball of radius $0.015$. The confidence function $\lambda$ is set to $1$ when the agent does not reach the big goal of radius $0.1$ and $0$ otherwise.

\paragraph{Point-Fall} It is a typical task in the Hierarchical RL literature that was introduced in \citep{nachum2018data}, with an open source code proposed by \citep{khachaturov2021susceptibility}. An actuated point must reach a goal placed in front of it, but separated by a chasm. In order to be able to cross it and reach the target, it must first put a movable block in the chasm. The reward is equal to $1$ when the agent reaches the goal and $0$ otherwise. The local policy puts the movable block in the chasm, so $\lambda(s)$ is equal to $1$ when the movable block is not the chasm and $0$ otherwise. The local policy is a SAC agent with medium performances that was only trained to put the movable block in the chasm. Then, in order to further decrease its performance, we reduced the speed of the moving point by a multiplicative factor of $0.2$.

\paragraph{Safe Cartpole Swingup} The goal is to swing up and stabilize an unactuated pole on a moving cart. The cart is constrained to stay within a certain range defined by $x_{\max} = 1.9$ and is penalized with a reward of $-1000$ as soon as the agent enters the restricted area. This constraint directly perturbs the performance of the agent since it sometimes bounces on its limits to swing up. If the agent is in a safe state, it receives the traditional reward that pushes the pole to stay upright at the center of the environment. The action is continuous and moves the cart with a bounded force (left or right). The local policy applies the maximum force $a = \pm 1$ towards the center, as soon as $|x| > 0.2 $. This might seem like a big area, but increasing the number $0.2$ lead to safety violations for the SAG agent. The confidence function $\lambda$ is set to $1$ when $|x| \geq 0.2$ and $0$ otherwise.

\paragraph{Point-Circle} Inspired from \citep{achiam2017constrained}, an actuated point must move on a circle in a clockwise direction, while staying in a safe region defined by two vertical barriers located at the horizontal extremities of the circle ($\lvert x \rvert = 6$). Similar to the above environment, this constraint is an obstacle to finding the optimal policy, and we add a reward of $-1000$ when the agent goes beyond the allowed zone. The local policy is a repulsive policy that as soon as activated, positions the cart in the middle of the environment by applying the maximum $a = \pm [1, 0]$. The confidence function $\lambda$ is set to $1$ when $|x| \geq 2$ and $0$ otherwise.

\paragraph{Point-Reach} This environment considers a ball that must reach a target in a $2$D-plane while avoiding $5$ obstacles. Similarly, we add a $-1000$ reward when the agent collides with the obstacle and we keep the original reward of the environment otherwise. As soon as the agent enters the circle of radius $3$ centered around an obstacle, the local policy is activated to go away from the obstacle. Especially, the area around the circle has been divided into $4$ quadrants, and the emergency procedure pushes the agent towards the diagonal of the quadrant $a\in\{ [ 1, 1 ], [ 1, -1 ], [ -1, 1 ], [ -1, -1] \}$.

\section{Comparison of different Empirical Bellman Operators for the SAG agent}

As a reminder, at epoch $k$, the Approximate Policy Evaluation is classically done in the following way:

\begin{equation}
    Q_\omega^{{k+1}} \leftarrow \arg\min_{\omega\in\Omega} \,\hat{\mathbb{E}} \left[ \left( Q_\omega - \hat{\mathcal{B}}^{\pi_\theta^k}\left[ Q_{\bar{\omega}_k}^k \right] \right)^2 \right]
\end{equation}

However, when the agent gathers data with its switched policy $\pi_{\text{SAG}}^k$, this might raise the distribution shift problem present in the offline setting. Hence, we proposed to evaluate the policy of the switched policy instead with:

\begin{equation}
    Q_\omega^{{k+1}} \leftarrow \arg\min_{\omega\in\Omega} \,\hat{\mathbb{E}} \left[ \left( Q_\omega - \hat{\mathcal{B}}^{\pi_{SAG}^k}\left[ Q_{\bar{\omega}_k}^k \right] \right)^2 \right]
\end{equation}

We denote the first version the Naive SAG, and our version remains SAG. 

\begin{figure}[h!]
\centering
\begin{subfigure}{.32\textwidth}
  \centering
  \includegraphics[width=.99\linewidth]{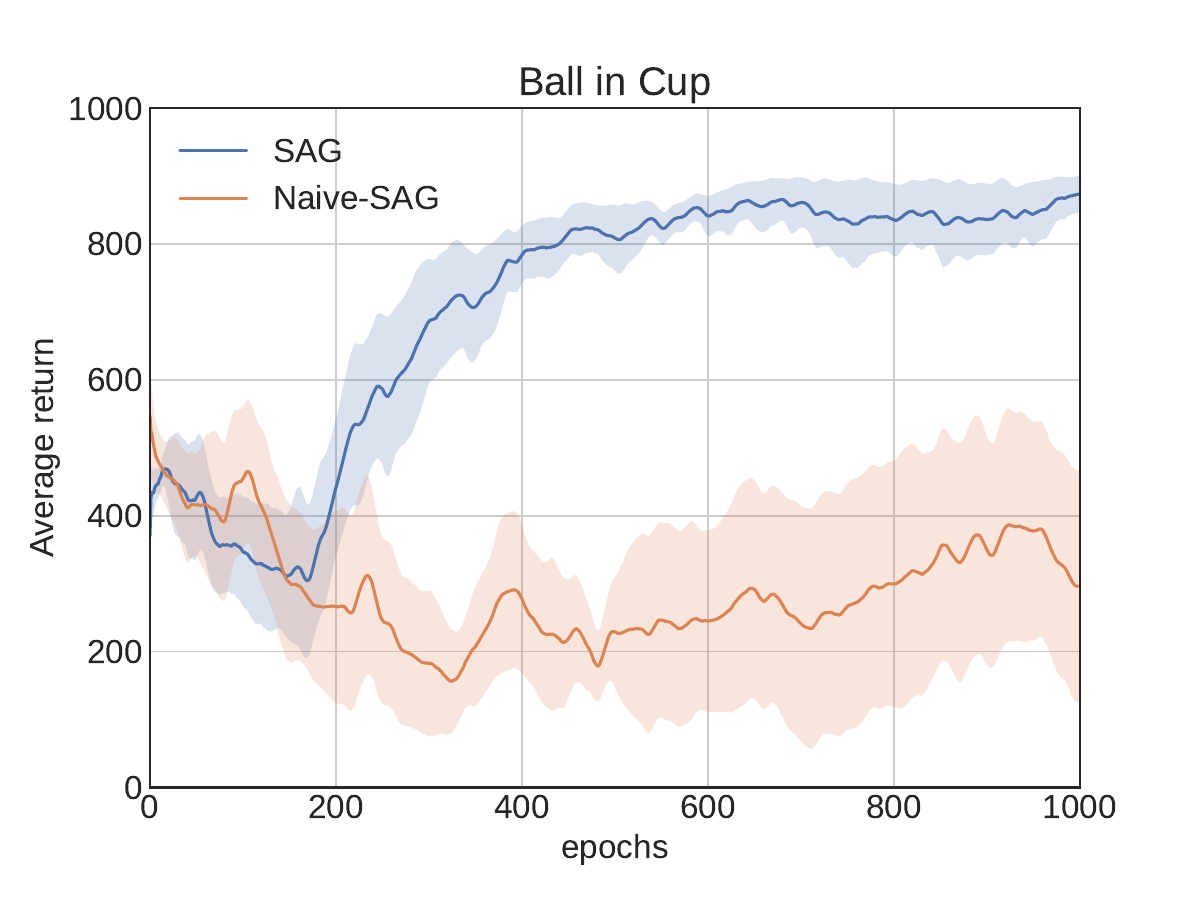}
  \caption{Ball-in-Cup}
  \label{fig:ball_in_cup_comparison_switched}
\end{subfigure}%
\begin{subfigure}{.32\textwidth}
  \centering
  \includegraphics[width=.99\linewidth]{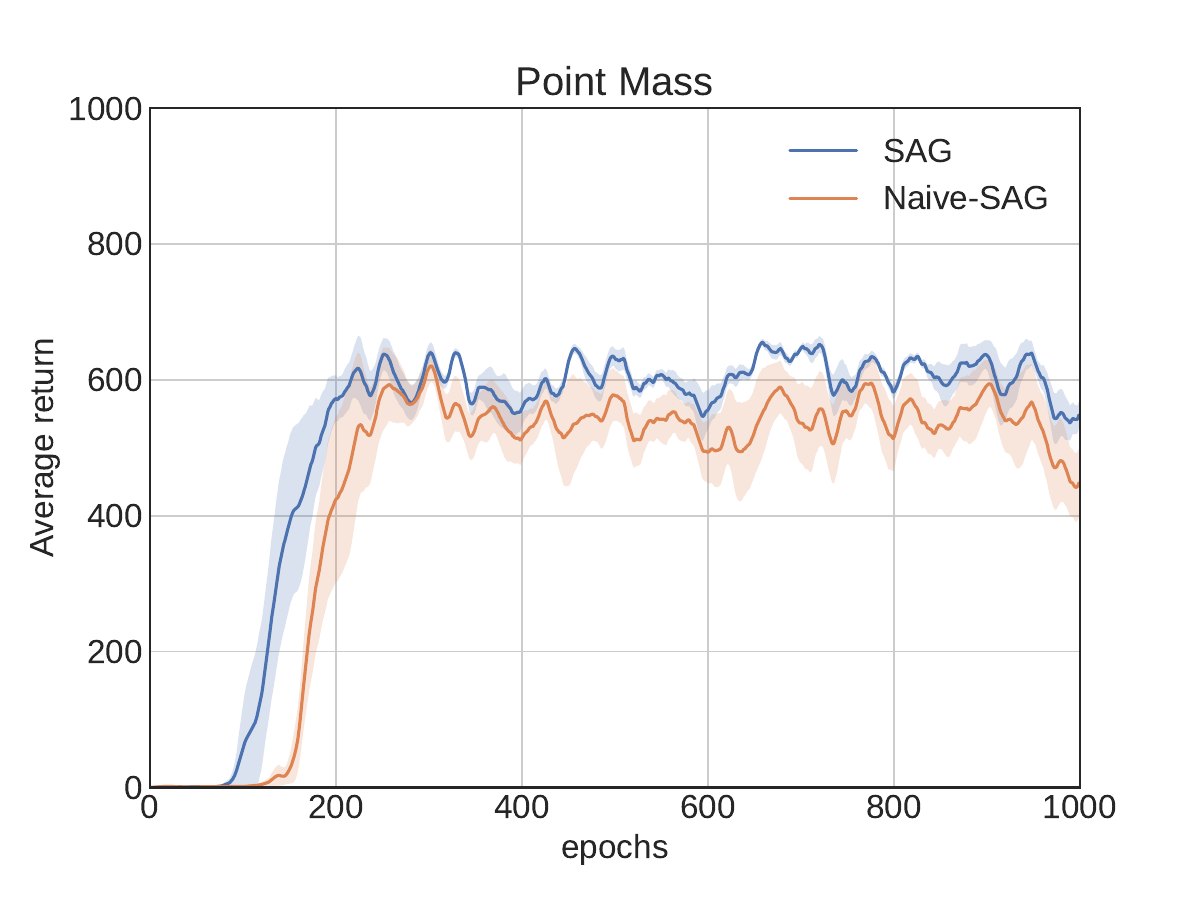}
  \caption{Point-Mass}
  \label{fig:point_mass_comparison_switched}
\end{subfigure}
\begin{subfigure}{.32\textwidth}
  \centering
  \includegraphics[width=.99\linewidth]{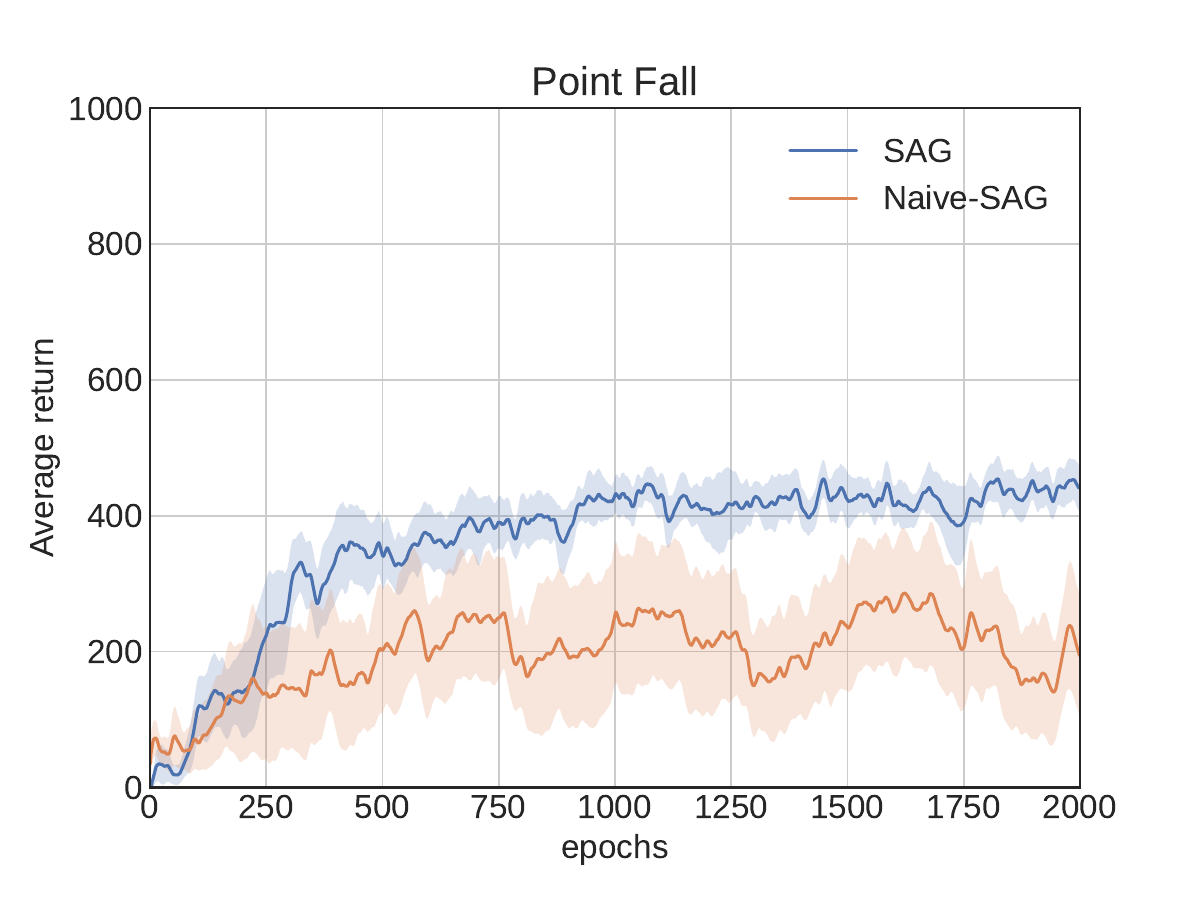}
  \caption{Point-Fall}
  \label{fig:point_fall_comparison_switched} \hspace{\fill}
\end{subfigure}
\begin{subfigure}{.32\textwidth}
  \centering
  \includegraphics[width=.99\linewidth]{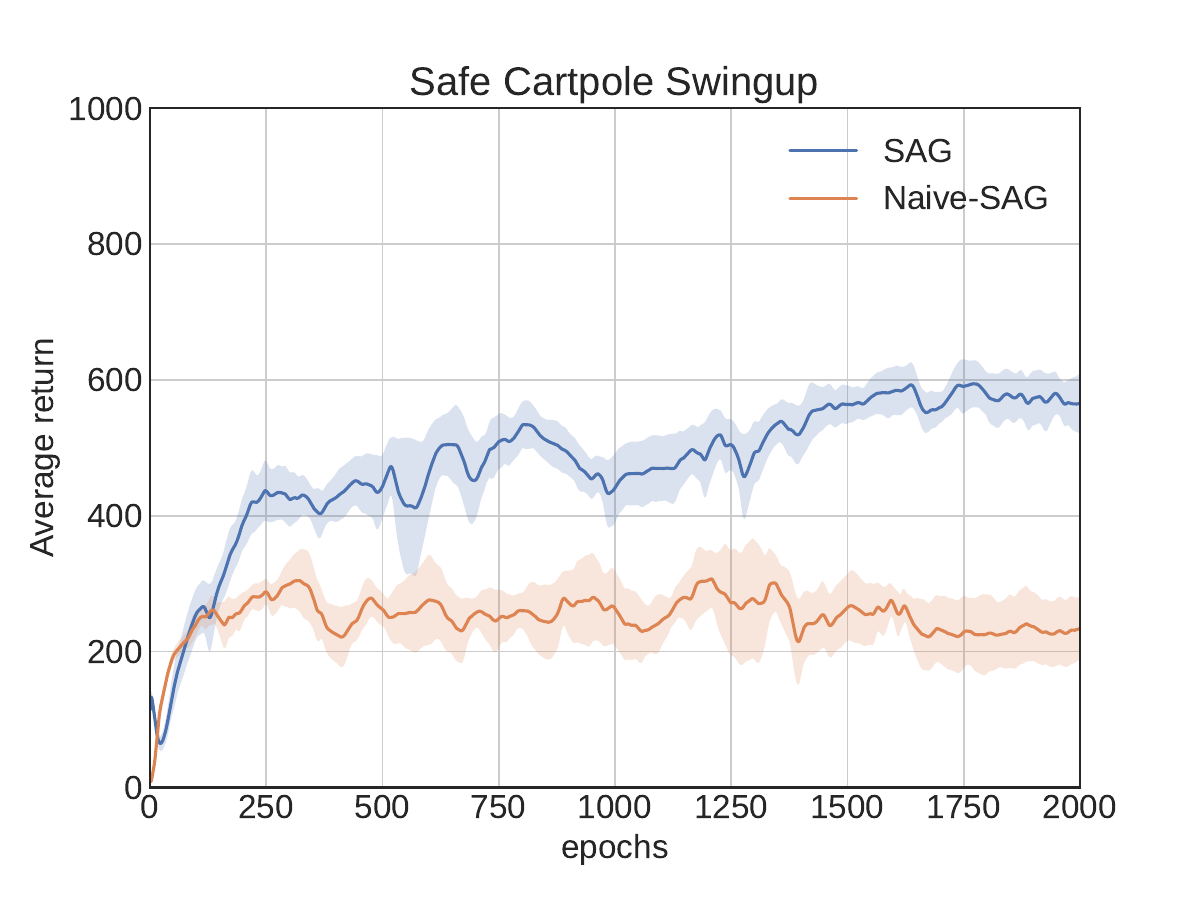}
  \caption{Safe-Cartpole-Swingup}
  \label{fig:safe_cartpole_comparison_switched}
\end{subfigure}
\begin{subfigure}{.32\textwidth}
  \centering
  \includegraphics[width=.99\linewidth]{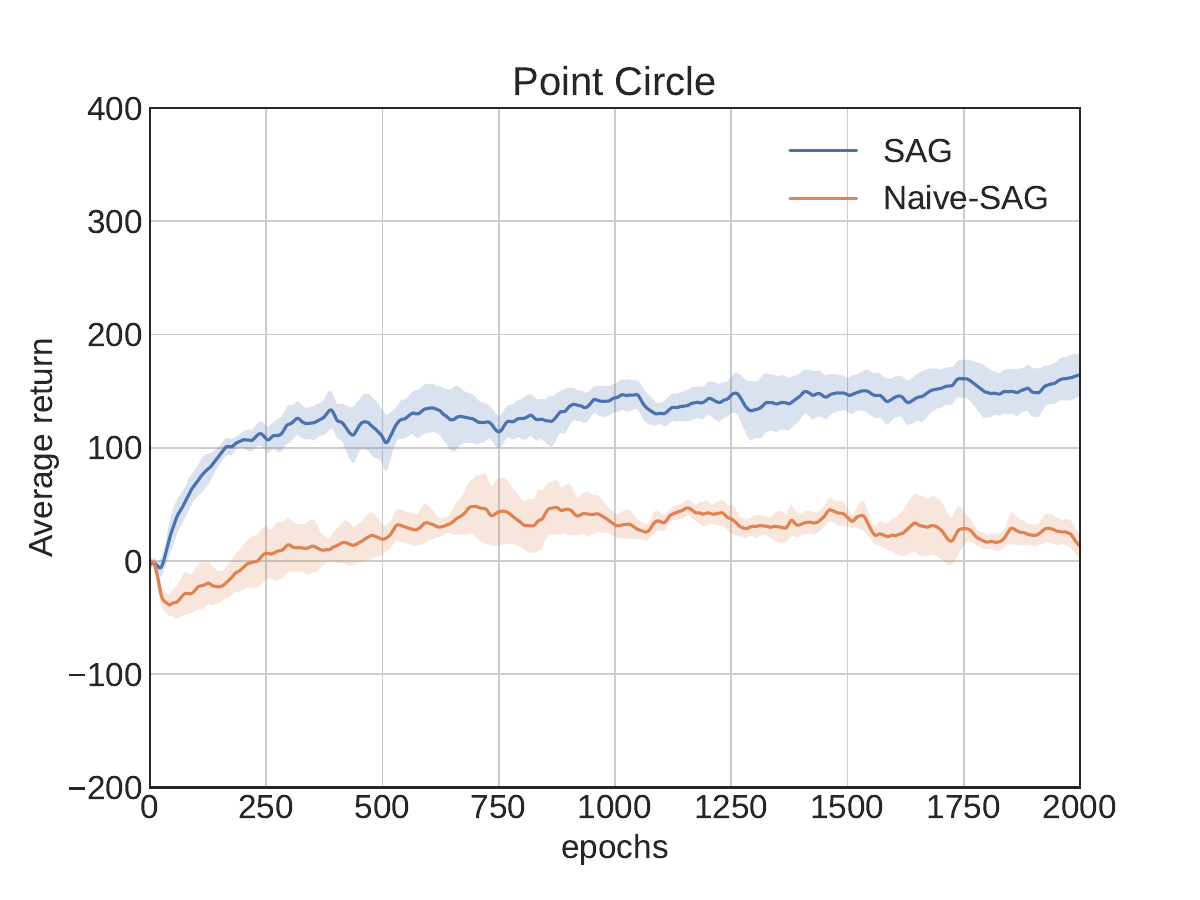}
  \caption{Point-Circle}
  \label{fig:point_circle_comparison_switched}
\end{subfigure}
\begin{subfigure}{.32\textwidth}
  \centering
  \includegraphics[width=.99\linewidth]{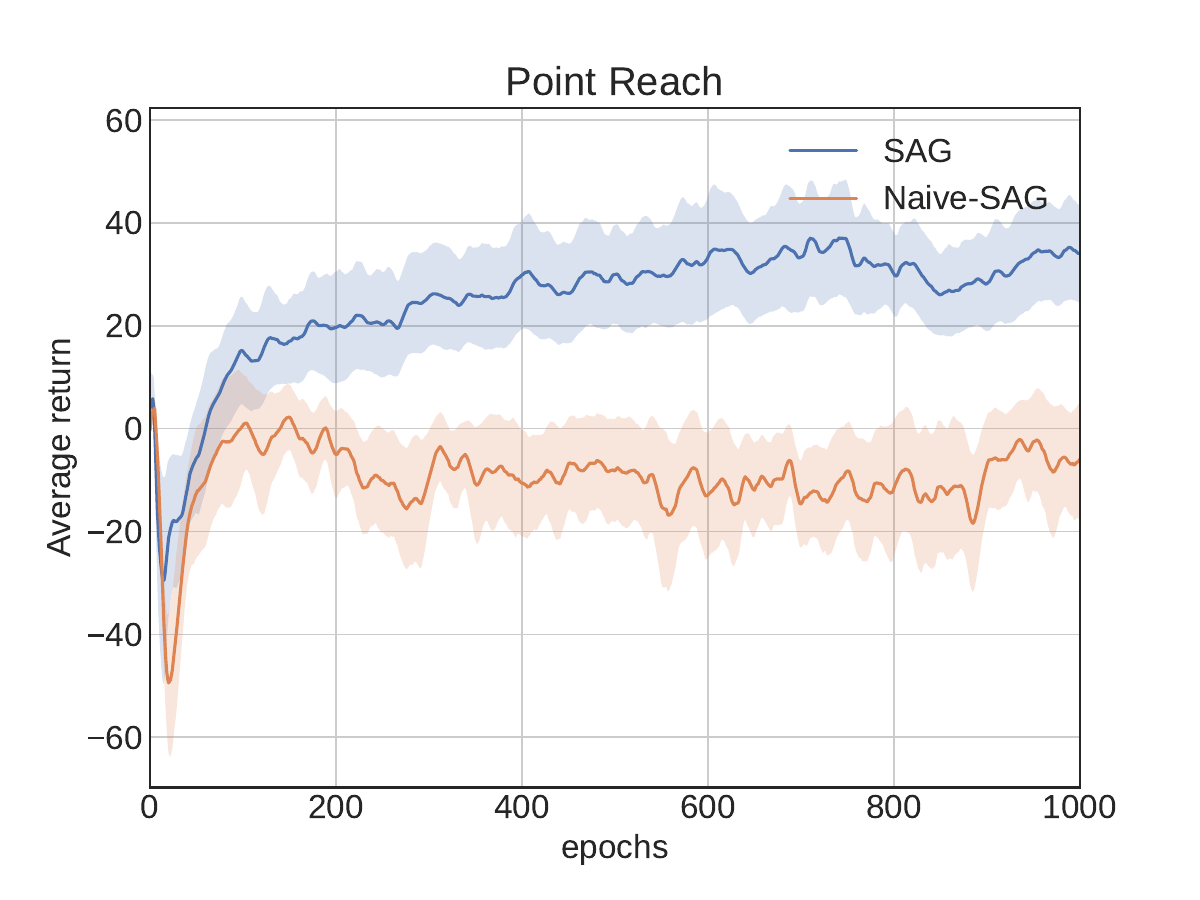}
  \caption{Point-Reach}
  \label{fig:point_reach_comparison_switched}
\end{subfigure}
\caption{Comparison between the Empirical Bellman Operators used on SAG on all environments.}
\label{fig:comparison_switched}
\end{figure}

We found that performing the Approximate Policy Evaluation with $\pi_{\text{SAG}}^k$ instead of $\pi_\theta^k$ leads to better results according to Figure~(\ref{fig:comparison_switched}) in all environments. Indeed, it seems that the data gathered with the different successive $\pi_{\text{SAG}}^k$ is not able to accurately estimate the $Q$-values associated to $\pi_\theta^k$. The distribution shift highlighted in the \textit{Offline} setting \citep{fujimoto2018addressing, kumar2020conservative, levine2020offline} also poses a problem here.

\section{Reward Guided agent} \label{app:reward_guided_agents}

\begin{figure}[ht!]
\centering
\begin{subfigure}{.32\textwidth}
  \centering
  \includegraphics[width=.99\linewidth]{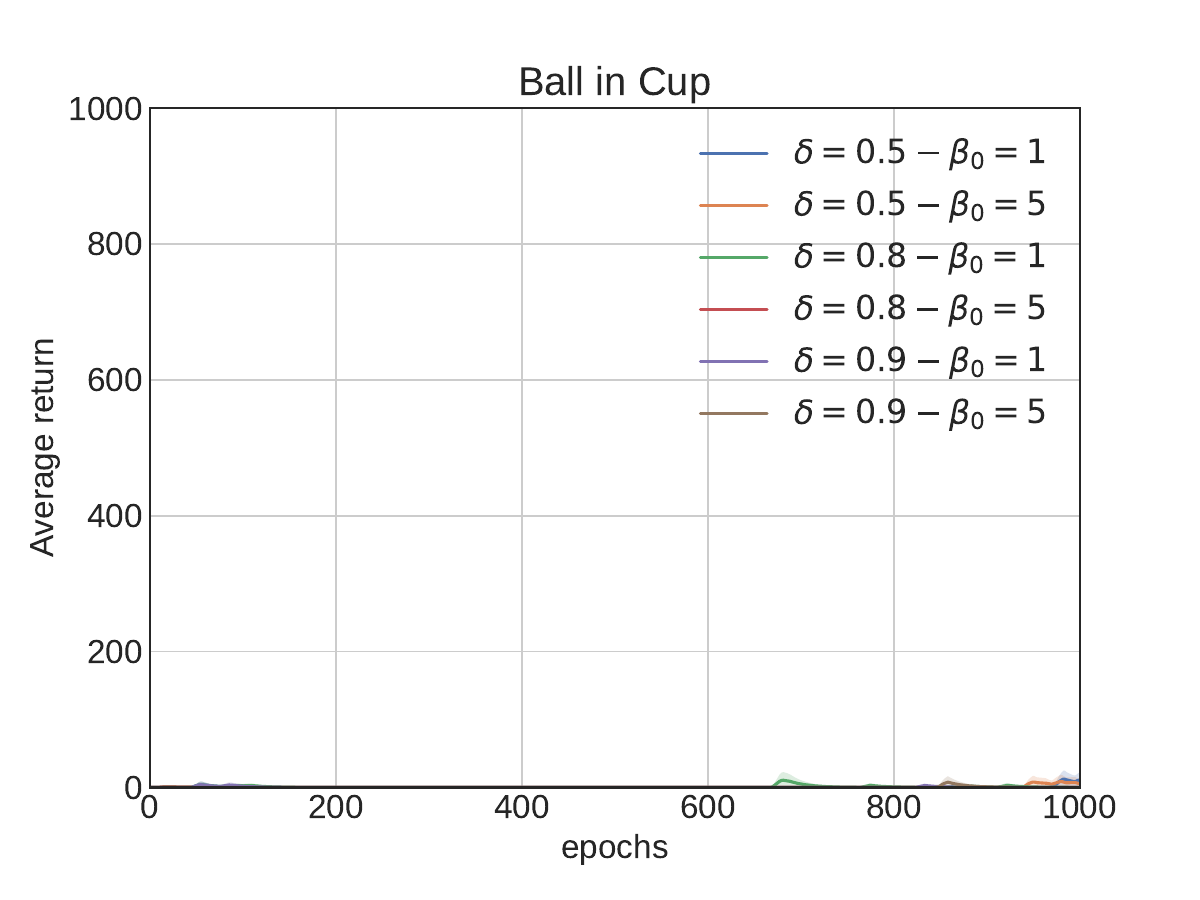}
  \caption{Ball-in-Cup}
  \label{fig:ball_in_cup_comparison_reward}
\end{subfigure}%
\begin{subfigure}{.32\textwidth}
  \centering
  \includegraphics[width=.99\linewidth]{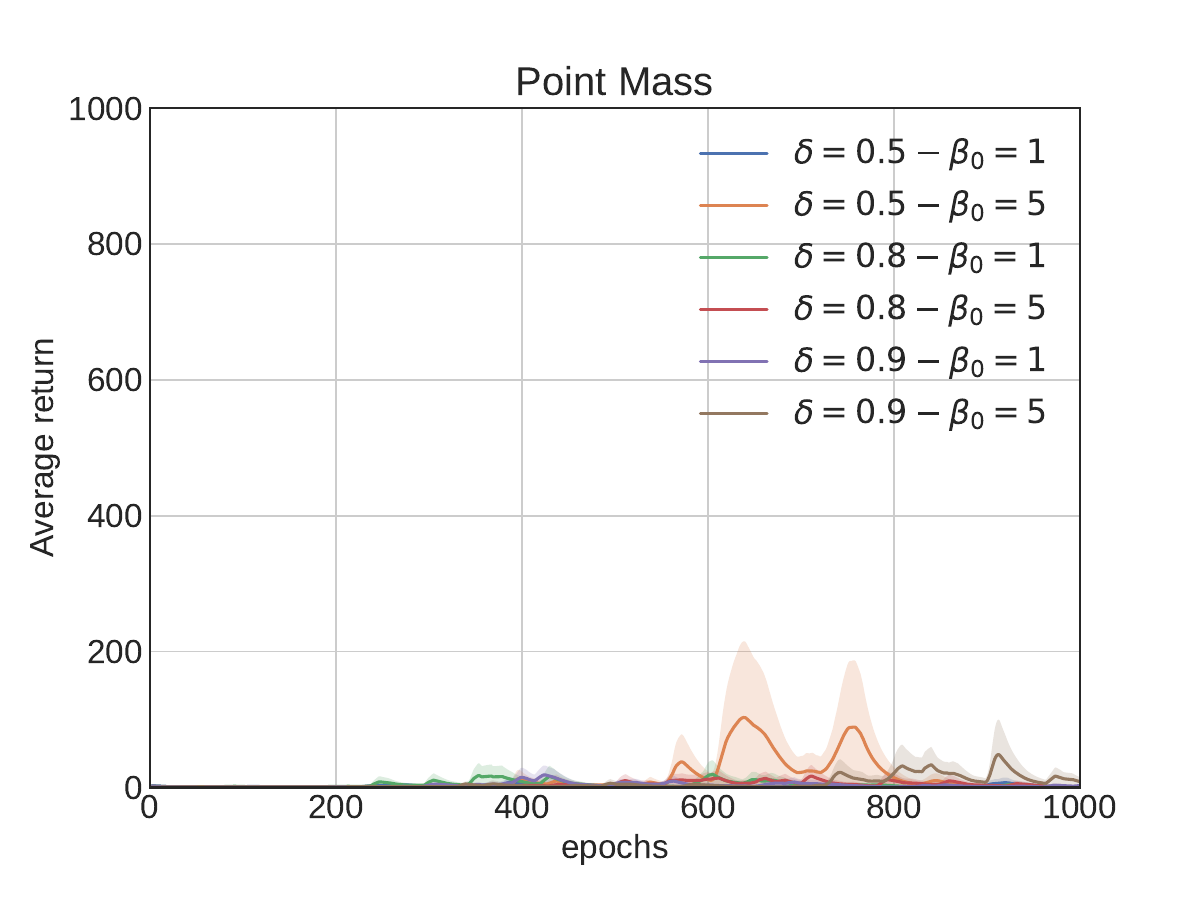}
  \caption{Point-Mass}
  \label{fig:point_mass_comparison_reward}
\end{subfigure}
\begin{subfigure}{.32\textwidth}
  \centering
  \includegraphics[width=.99\linewidth]{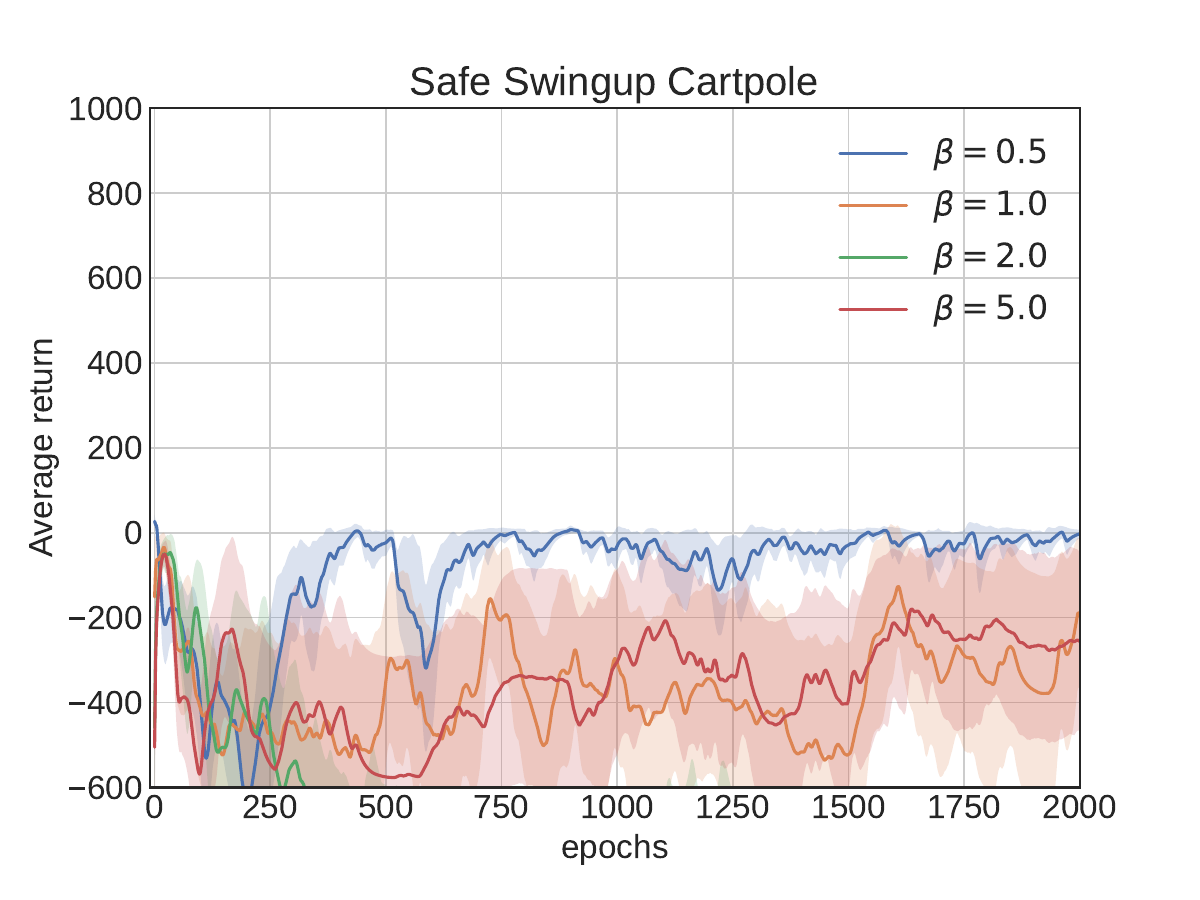}
  \caption{Cartpole Swingup}
  \label{fig:safe_cartpole_comparison_reward} 
\end{subfigure}
\caption{Reward Guided agent on Ball-in-Cup, Point-Mass and the Safe Cartpole Swingup environments.}
\label{fig:comparison_reward}
\end{figure}

Shaping the reward turned out to be quite complicated in the Reinforcement Learning with Local Guides setting. In fact, our experiments on \textit{Ball in Cup}, \textit{Point-Mass} and \textit{Safe Cartpole Swingup} in Figure~(\ref{fig:comparison_reward}) involving the Reward-Guided agents lead to poor results. We did not pursue these experiments on the other environments, as Figure~(\ref{fig:comparison_reward}) clearly demonstrates that the Reward-Guided agents are not suited for this setting. 

\end{document}